\documentclass[lettersize,journal]{IEEEtran}
\usepackage{amsmath,amsfonts}
\usepackage{algorithmic}
\usepackage{algorithm}
\usepackage{array}
\usepackage{textcomp}
\usepackage{stfloats}
\usepackage{url}
\usepackage{verbatim}
\usepackage{graphicx}
\usepackage{cite}
\usepackage{subcaption}
\usepackage{multirow}
\usepackage[T1]{fontenc}
\usepackage{adjustbox}
\usepackage{amssymb}
\usepackage{booktabs}
\usepackage{tikz}
\usepackage{tabularx}
\usepackage{mathrsfs}
\usepackage{etoolbox}

\newcommand{\disablecitations}{%
    \renewcommand{\cite}[1]{}%
}

\newcommand{\enablecitations}{%
    \let\cite\oldcite%
}

\let\oldcite\cite

\usepackage[colorlinks,bookmarksopen,bookmarksnumbered, linkcolor=red]{hyperref}
\definecolor{darkgreen}{RGB}{17,159,27} %
\begin{document}
\disablecitations
\enablecitations

\title{Polar R-CNN:\@ End-to-End Lane Detection with Fewer Anchors}

\author{Shengqi Wang, Junmin Liu, Xiangyong Cao, Zengjie Song, and Kai Sun\\

\thanks{This work was supported in part by the National Nature Science Foundation of China (Grant Nos. 62276208, 12326607) and in part by the Natural Science Basic Research Program of Shaanxi Province (Grant No. 2024JC-JCQN-02).}%
\thanks{S. Wang, J. Liu, Z. Song and K. Sun are with the School of Mathematics and Statistics, Xi'an Jiaotong University, Xi'an 710049, China.}
\thanks{X. Cao is with the School of Computer Science and Technology and the Ministry of Education Key Lab for Intelligent Networks and Network Security, Xi’an Jiaotong University, Xi’an 710049, China.}
}


\markboth{S. Wang \MakeLowercase{\textit{et al.}}: Polar R-CNN:\@ End-to-End Lane Detection with Fewer Anchors}%
{S. Wang \MakeLowercase{\textit{et al.}}: Polar R-CNN:\@ End-to-End Lane Detection with Fewer Anchors}

\maketitle

\begin{abstract}
    Lane detection is a critical and challenging task in autonomous driving, particularly in real-world scenarios where traffic lanes can be slender, lengthy, and often obscured by other vehicles, complicating detection efforts. Existing anchor-based methods typically rely on prior lane anchors to extract features and  subsequently refine the location and shape of lanes. While these methods achieve high performance, manually setting prior anchors is cumbersome, and ensuring sufficient  coverage across diverse datasets often requires a large amount of dense anchors. Furthermore,
    the use of \textit{Non-Maximum Suppression} (NMS) to eliminate redundant predictions complicates real-world deployment and may underperform in complex scenarios. In this paper, we propose \textit{Polar R-CNN}, an end-to-end anchor-based method for lane detection. By incorporating both local and global polar coordinate systems, Polar R-CNN facilitates flexible anchor proposals and significantly reduces the number of anchors required without compromising performance. Additionally, we introduce a triplet head with heuristic structure that supports NMS-free paradigm, enhancing deployment efficiency and performance in scenarios with dense lanes. Our method achieves competitive results on five popular lane detection benchmarks—\textit{Tusimple}, \textit{CULane}, \textit{LLAMAS}, \textit{CurveLanes}, and \textit{DL-Rail}—while maintaining a lightweight design and straightforward structure. Our source code is available at \href{https://github.com/ShqWW/PolarRCNN}{\textit{https://github.com/ShqWW/PolarRCNN}}.
\end{abstract}
\begin{IEEEkeywords}
Lane Detection, NMS-Free, Graph Neural Network, Polar Coordinate System.
\end{IEEEkeywords}

\section{Introduction}
\IEEEPARstart{L}{ane} detection is a critical task in computer vision and autonomous driving, aimed at identifying and tracking lane markings on the road \cite{adas}. While extensive research has been conducted in ideal environments, it is still challenging in adverse scenarios such as night driving, glare, crowd, and rainy conditions, where lanes may be occluded or damaged \cite{scnn}. Moreover, the slender shapes and complex topologies of lanes further complicate detection efforts \cite{polylanenet}.
\par
In the past few decades, a lot of methods primarily focus on handcrafted local feature extraction and lane shape modeling. Techniques such as the \textit{Canny edge detector}\cite{cannyedge},\textit{ Hough transform}\cite{houghtransform}, and \textit{deformable templates}\cite{kluge1995deformable} have been widely employed for lane fitting. However, these approaches often face limitations in real-world scenarios, especially when low-level and local features lack clarity and distinctiveness.
\par
In recent years, advancements in deep learning and the availability of large datasets have led to significant progress in lane detection, especially deep models such as \textit{Convolutional Neural Networks} (CNNs)\cite{scnn} and \textit{transformer-based} architectures \cite{lstr}. Based on this, earlier approaches typically framed lane detection as a \textit{segmentation task} \cite{lanenet}, which, despite its straightforward, required time-consuming computations. There are still some methods that rely on \textit{parameter-based} models, which directly output lane curve parameters rather than pixel locations \cite{polylanenet}\cite{lstr}\cite{bezierlanenet}. Although these segmentation-based and parameter-based methods provide end-to-end solutions, their sensitivity to lane shape compromises their robustness.
\begin{figure}[t]
        \centering
        \def\subwidth{0.24\textwidth}
        \def\imgwidth{\linewidth}
        \def\imgheight{0.5625\linewidth}

        \begin{subfigure}{\subwidth}
                \includegraphics[width=\imgwidth, height=\imgheight]{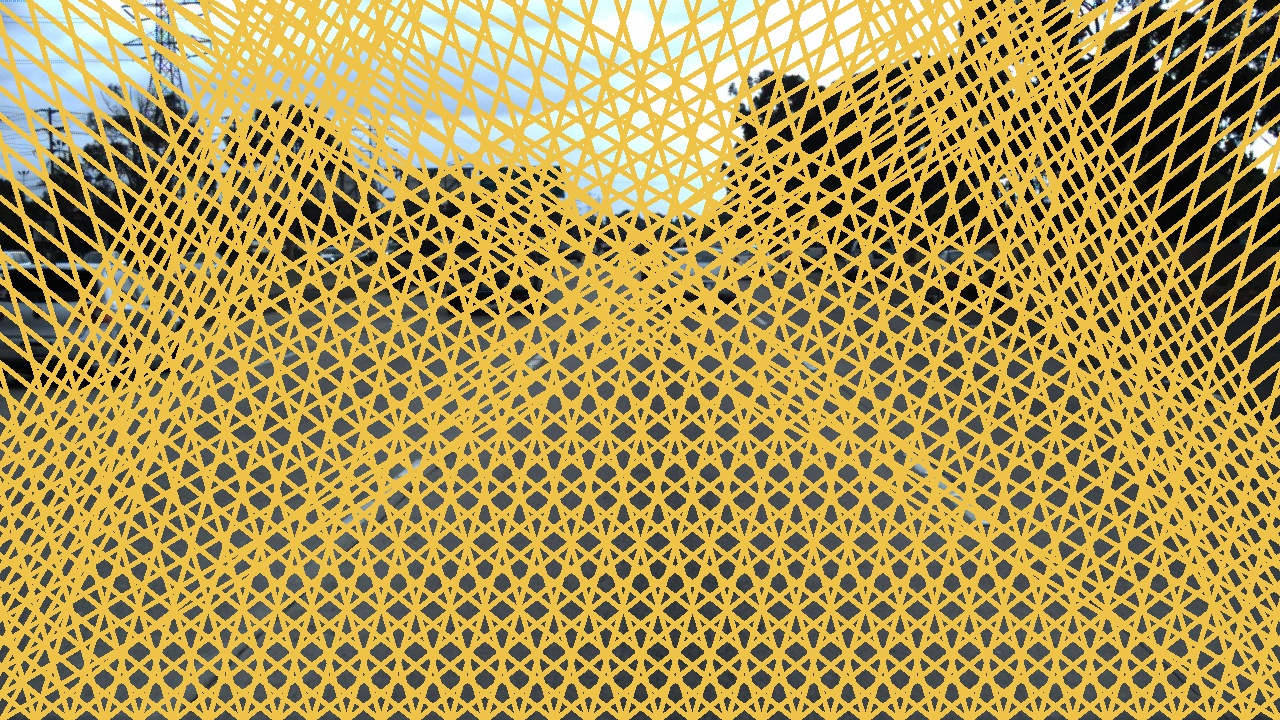}
                \caption{}
        \end{subfigure}
        \begin{subfigure}{\subwidth}
                \includegraphics[width=\imgwidth, height=\imgheight]{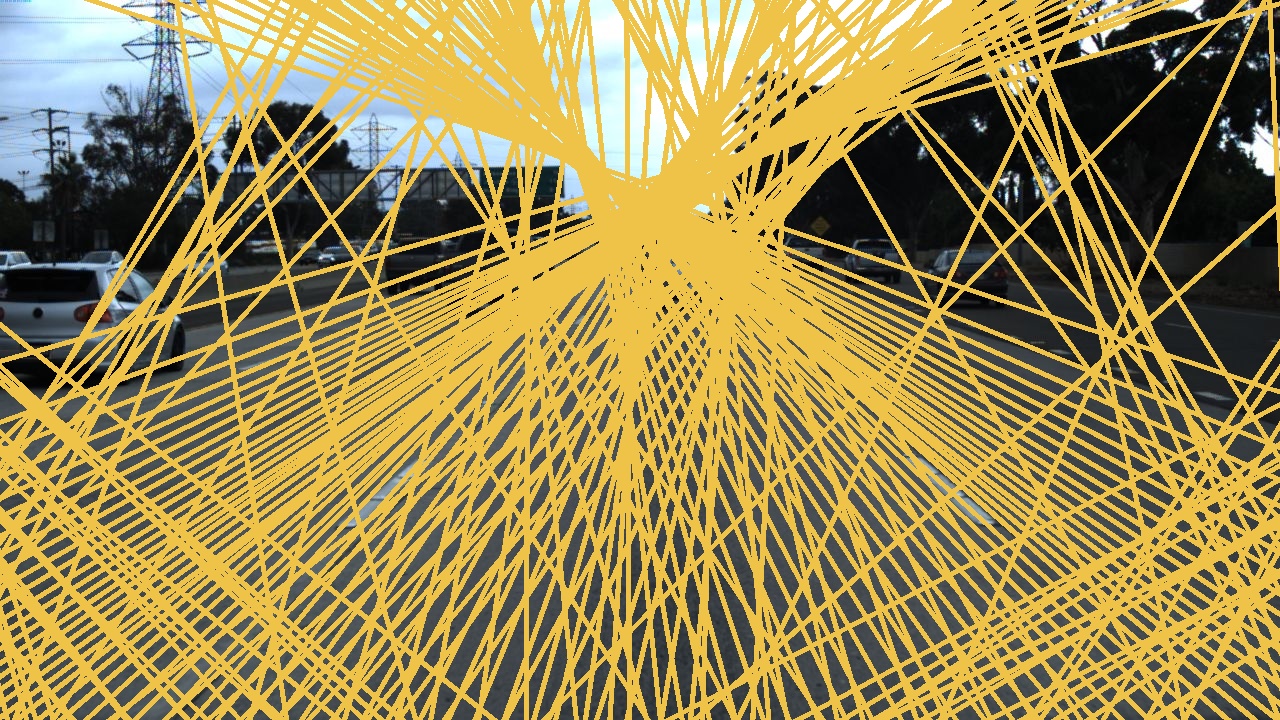}
                \caption{}
        \end{subfigure}

        \begin{subfigure}{\subwidth}
                \includegraphics[width=\imgwidth, height=\imgheight]{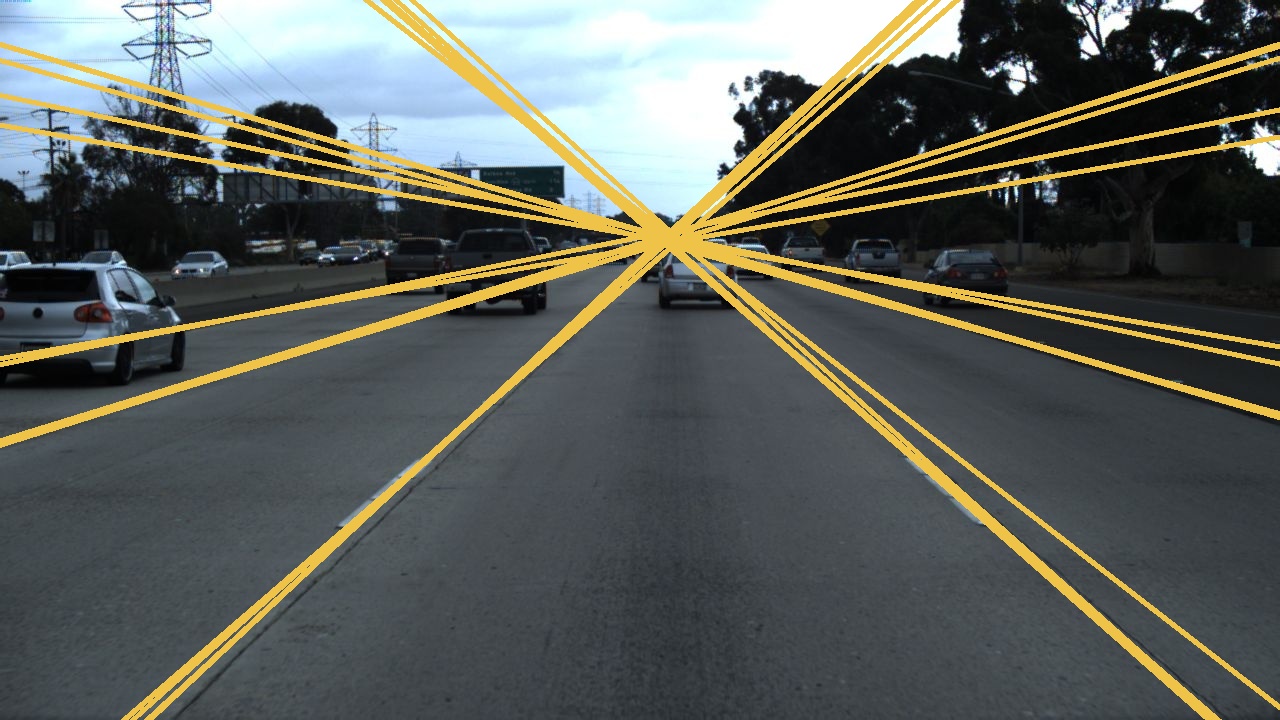}
                \caption{}
        \end{subfigure}
        \begin{subfigure}{\subwidth}
                \includegraphics[width=\imgwidth, height=\imgheight]{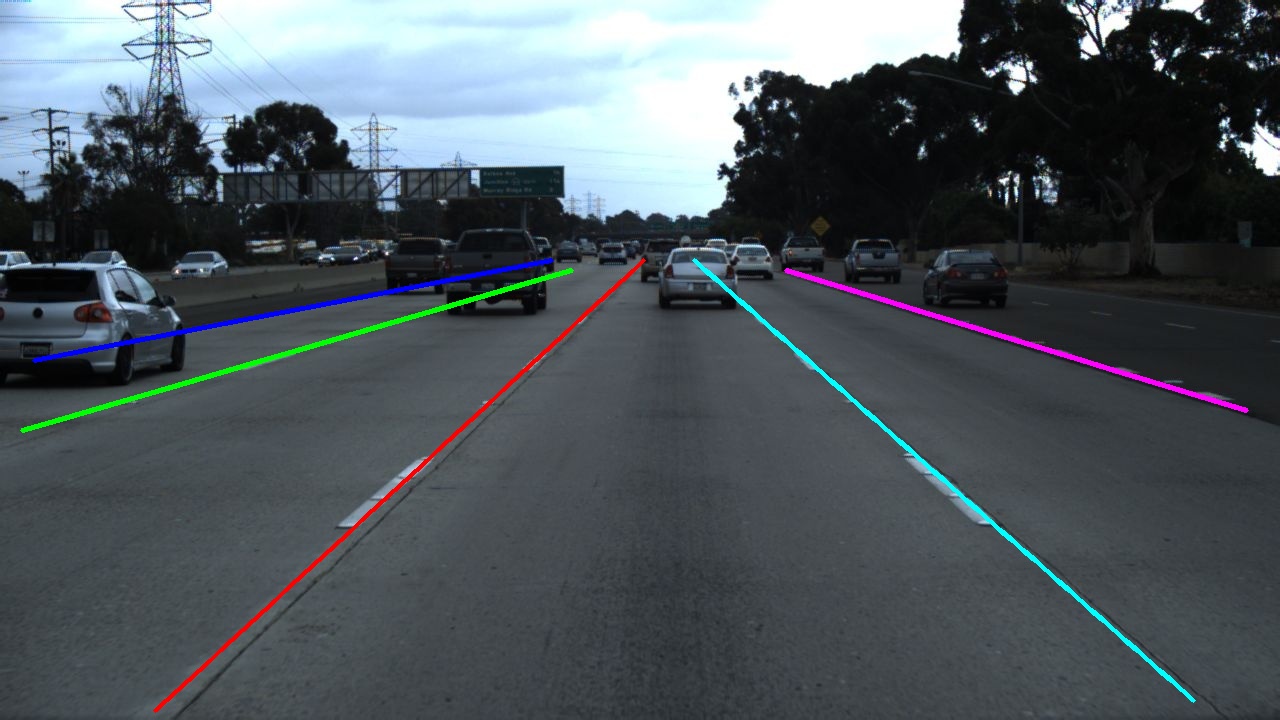}
                \caption{}
        \end{subfigure}
        \caption{Anchor (\textit{i.e.}, the yellow lines) settings of different methods and the ground truth lanes. (a) The initial anchor settings of CLRNet. (b) The learned anchor settings of CLRNet trained on CULane. (c) The flexible proposal anchors of our method. (d) The ground truth.}
        \label{anchor setting}
\end{figure}

\begin{figure}[t]
        \centering
        \def\subwidth{0.24\textwidth}
        \def\imgwidth{\linewidth}
        \def\imgheight{0.5625\linewidth}
        
        \begin{subfigure}{\subwidth}
                \includegraphics[width=\imgwidth, height=\imgheight]{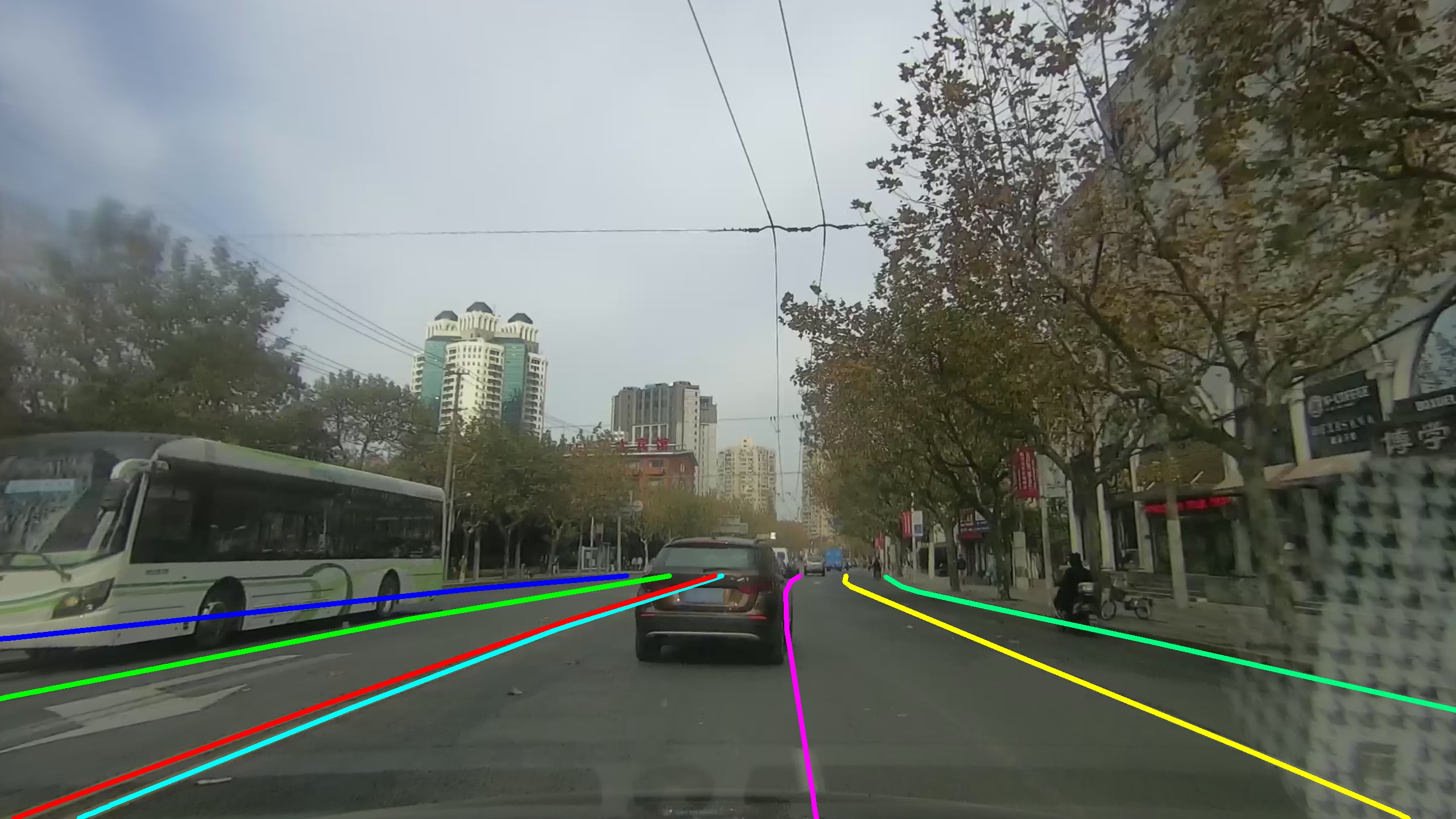}
                \caption{}
        \end{subfigure}
        \begin{subfigure}{\subwidth}
    	\includegraphics[width=\imgwidth, height=\imgheight]{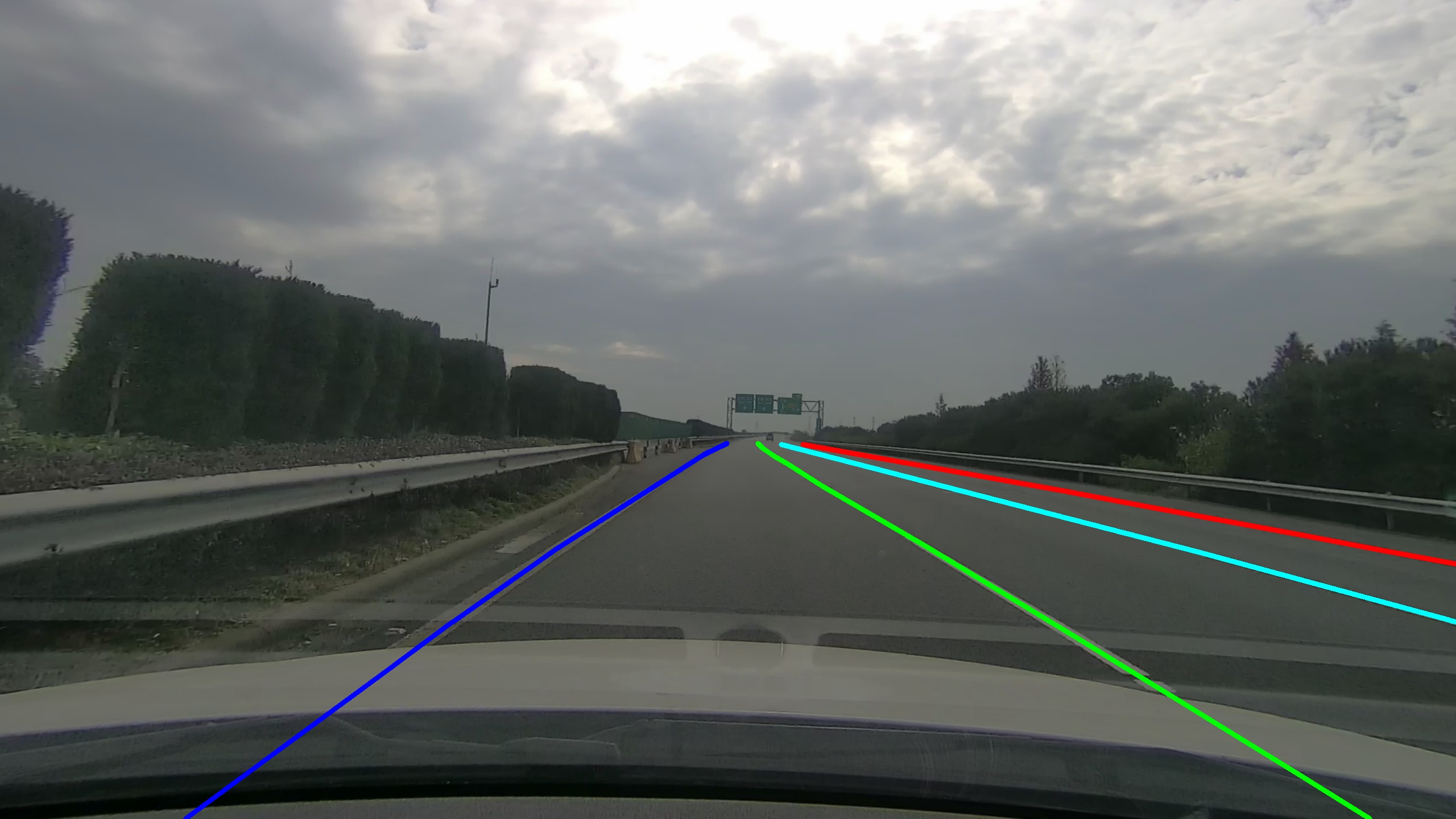}
    	\caption{}
        \end{subfigure}
        \begin{subfigure}{\subwidth}
                \includegraphics[width=\imgwidth, height=\imgheight]{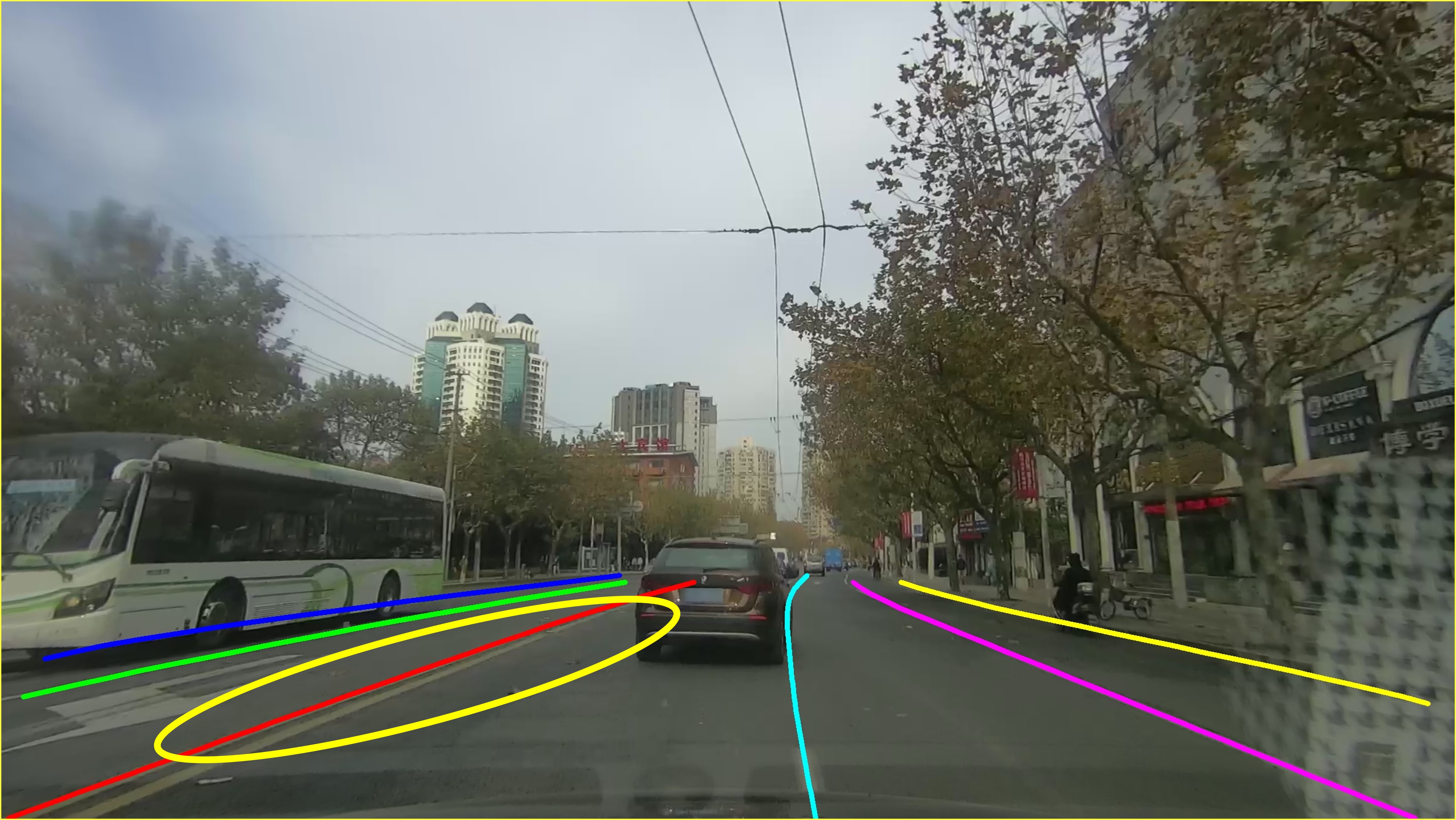}
                \caption{}
        \end{subfigure}
        \begin{subfigure}{\subwidth}
                \includegraphics[width=\imgwidth, height=\imgheight]{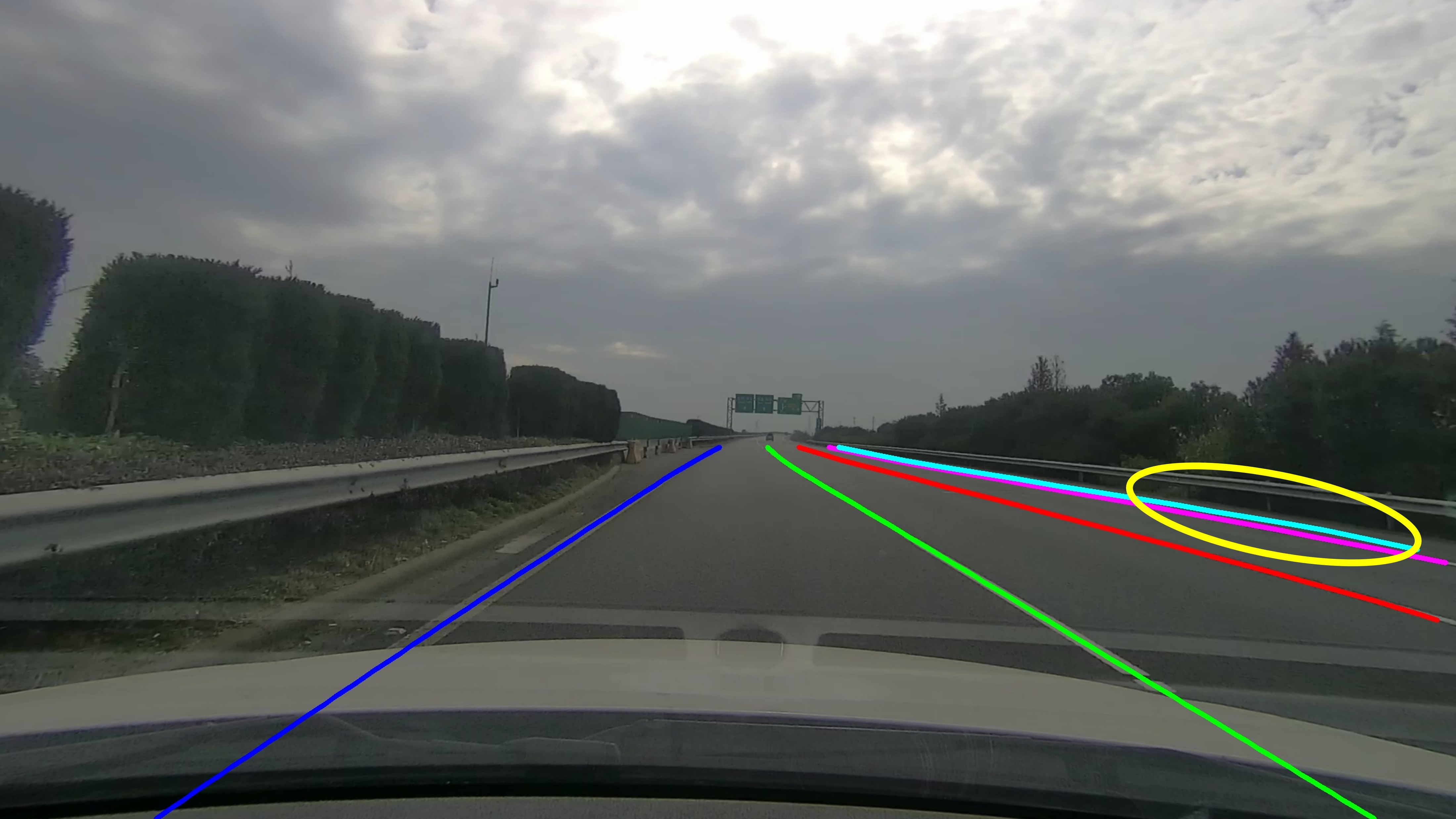}
                \caption{}
        \end{subfigure}

        \caption{Comparison of NMS thresholds in \textit{sparse} and \textit{dense} scenarios. (a) and (b) Ground truths in the dense and sparse scenarios, respectively. (c) Predictions with large NMS thresholds in a dense scenario, resulting in a lane prediction being mistakenly suppressed. (d) Predictions with small NMS thresholds in a sparse scenario, where redundant prediction results are not effectively removed.}
        \label{NMS setting}
\end{figure}
\par
Drawing inspiration from object detection methods such as \textit{YOLO} \cite{yolov10} and \textit{Faster R-CNN} \cite{fasterrcnn}, several anchor-based approaches have been introduced for lane detection, with representative works including \textit{LaneATT} \cite{laneatt} and \textit{CLRNet} \cite{clrnet}. These methods have shown superior performance by leveraging anchor \textit{priors} (as shown in Fig. \ref{anchor setting}) and enabling larger receptive fields for feature extraction. However, anchor-based methods encounter similar drawbacks to those in general object detection, including the following:
\begin{itemize}
\item As shown in Fig. \ref{anchor setting}(a), a large amount of lane anchors are predefined in the image, even in \textbf{\textit{sparse scenarios}}---the situations where lanes are distributed widely and located far apart from each other, as illustrated in the Fig. \ref{anchor setting}(d).
\item \textit{Non-Maximum Suppression} (NMS) \cite{nms} post-processing is required to eliminate redundant predictions but may struggle in \textbf{\textit{dense scenarios}} where lanes are close to each other, such as forked lanes and double lanes, as illustrated in the Fig. \ref{NMS setting}(a).
\end{itemize}
\par
Regrading the first issue, \cite{clrnet} introduced learned anchors that optimize the anchor parameters during training to better adapt to lane distributions, as shown in Fig. \ref{anchor setting}(b). However, the number of anchors remains excessive to adequately cover the diverse potential distributions of lanes. Furthermore, \cite{adnet} proposes flexible anchors for each image by generating start points with directions, rather than using a fixed set of anchors. Nevertheless, these start points of lanes are subjective and lack clear visual evidence due to the global nature of lanes. In contrast, \cite{srlane} uses a local angle map to propose sketch anchors according to the direction of ground truth. While this approach considers directional alignment, it neglects precise anchor positioning, resulting in suboptimal performance. Overall, the abundance of anchors is unnecessary in sparse scenarios.
\par
Regarding the second issue, nearly all anchor-based methods \cite{laneatt}\cite{clrnet}\cite{adnet}\cite{srlane} rely on direct or indirect NMS post-processing to eliminate redundant predictions. Although it is necessary to eliminate redundant predictions, NMS remains a suboptimal solution. On one hand, NMS is not deployment-friendly because it requires defining and calculating distances between lane pairs using metrics such as \textit{Intersection over Union} (IoU). This task is more challenging than in general object detection due to the intricate geometry of lanes. On the other hand, NMS can struggle in dense scenarios. Typically, a large distance threshold may lead to false negatives, as some true positive predictions could be mistakenly eliminated, as illustrated in Fig. \ref{NMS setting}(a)(c). Conversely, a small distance threshold may fail to eliminate redundant predictions effectively, resulting in false positives, as shown in Fig. \ref{NMS setting}(b)(d). Therefore, achieving an optimal trade-off across all scenarios by manually setting the distance threshold is challenging. 
\par
To address the above two issues, we propose Polar R-CNN, a novel anchor-based method for lane detection. For the first issue, we introduce \textit{Local Polar Module} based on the polar coordinate system to create anchors with more accurate locations, thereby reducing the number of proposed anchors in sparse scenarios, as illustrated in Fig. \ref{anchor setting}(c). In contrast to \textit{State-Of-The-Art} (SOTA) methods \cite{clrnet}\cite{clrernet}, which utilize 192 anchors, Polar R-CNN employs only 20 anchors to effectively cover potential lane ground truths. For the second issue, we have incorporated a triplet head with a new heuristic \textit{Graph Neural Network} (GNN) \cite{gnn} block. The GNN block offers an interpretable structure, achieving nearly equivalent performance in sparse scenarios and superior performance in dense scenarios. We conducted experiments on five major benchmarks: \textit{TuSimple} \cite{tusimple}, \textit{CULane} \cite{scnn}, \textit{LLAMAS} \cite{llamas}, \textit{CurveLanes} \cite{curvelanes}, and \textit{DL-Rail} \cite{dalnet}. Our proposed method demonstrates competitive performance compared to SOTA approaches. Our main contributions are summarized as follows:
\begin{itemize}
\item We design a strategy to simplify the anchor parameters by using local and global polar coordinate systems and applied these to the two-stage lane detection framework. Compared to other anchor-based methods, this strategy significantly reduces the number of proposed anchors while achieving better performance.
\item We propose a novel triplet detection head with a GNN block to implement a NMS-free paradigm. The block is inspired by Fast NMS, providing enhanced interpretability. Our model supports end-to-end training and testing while still allowing for traditional NMS post-processing as an option for a NMS version of our model.
\item By integrating the polar coordinate systems and NMS-free paradigm, we present a Polar R-CNN model for fast and efficient lane detection. And we conduct extensive experiments on five benchmark datasets to demonstrate the effectiveness of our model in high performance with fewer anchors and a NMS-free paradigm. 
\end{itemize}
\begin{figure*}[ht]
	\centering
	\includegraphics[width=0.99\linewidth]{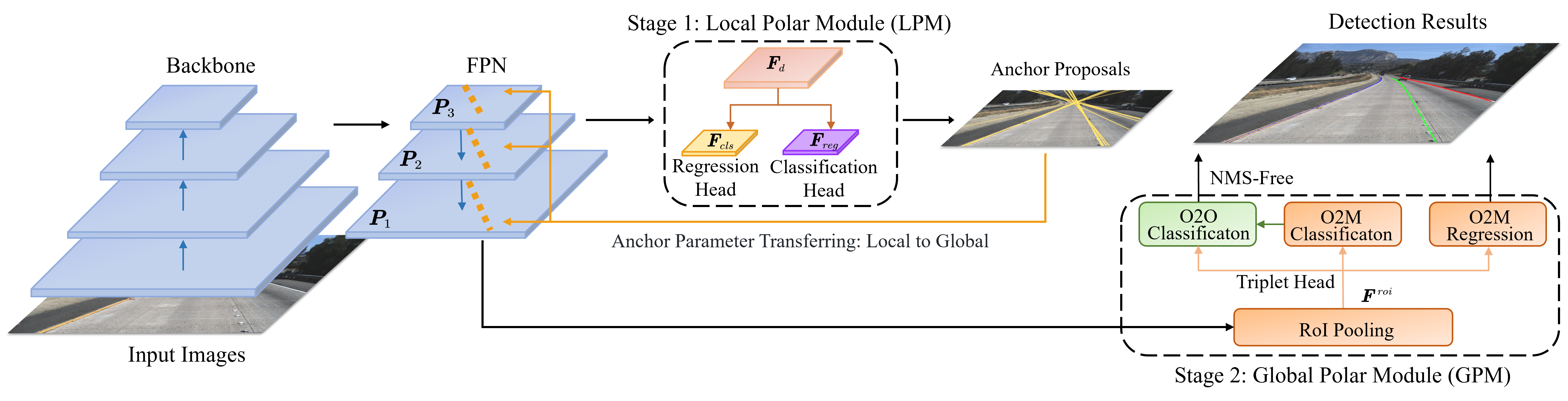} 
	\caption{An illustration of the Polar R-CNN architecture. It has a similar pipeline with the Faster R-CNN for the task of object detection, and consists of a backbone, a \textit{Feature Pyramid Network} with three levels of feature maps, respectively denote by $\boldsymbol{P}_1$, $\boldsymbol{P}_2$ and $\boldsymbol{P}_3$, followed by a \textit{Local Polar Module}, and a \textit{Global Polar Module} for lane detection. Based on the designed lane representation and lane anchor representation in polar coordinate system, the local polar module can propose sparse line anchors and the global polar module can produce the final accurate lane predictions. The global polar module includes a triplet head, which comprises the \textit{one-to-one} (O2O) classification subhead, the \textit{one-to-many} (O2M) classification subhead, and the \textit{one-to-many} (O2M) regression subhead.}
	\label{overall_architecture}
\end{figure*}
\section{Related Works}
Generally, deep learning-based lane detection methods can be categorized into three groups: segmentation-based, parameter-based, and anchor-based methods. Additionally, NMS-free is an important technique for anchor-based methods, and it will also be described in this section.
\par
\textbf{Segmentation-based Methods.} These methods focus on pixel-wise prediction. They predefined each pixel into different categories according to different lane instances and background\cite{lanenet} and predicted information pixel by pixel. However, they often overly emphasize low-level and local features, neglecting global semantic information and real-time detection. To address this issue, \textit{SCNN} \cite{scnn} uses a larger receptive field. There are some methods such as \textit{UFLDv1-v2} \cite{ufld}\cite{ufldv2} and \textit{CondLaneNet}\cite{CondLaneNet} by utilizing row-wise or column-wise classification instead of pixel classification to improve detection speed. Another issue with these methods is that the lane instance prior is learned by the model itself, leading to a lack of prior knowledge. For example, \textit{LaneNet}\cite{lanenet} uses post-clustering to distinguish each lane instance, while \textit{UFLDv1-v2}
categorizes lane instances by angles and locations, allowing it to detect only a fixed number of lanes. In contrast, \textit{CondLaneNet} employs different conditional dynamic kernels to predict different lane instances. Additionally, some methods such as \textit{FOLOLane}\cite{fololane} and \textit{GANet}\cite{ganet} adopt bottom-up strategies to detect a few key points and model their global relations to form lane instances.
\par
\textbf{Parameter-based Methods.} Instead of predicting a series of points locations or pixel classifications, the parameter-based methods directly generate the curve parameters of lane instances. For example, \textit{PolyLanenet}\cite{polylanenet} and \textit{LSTR}\cite{lstr} consider the lane instance as a polynomial curve, outputting the polynomial coefficients directly. \textit{BézierLaneNet}\cite{bezierlanenet} treats the lane instance as a Bézier curve, generating the locations of their control points, while \textit{BSLane}\cite{bsnet} uses B-Spline to describe the lane, with curve parameters that emphasize local lane shapes. These parameter-based methods are mostly end-to-end and do not require post-processing, resulting in faster inference speed. However, since the final visual lane shapes are sensitive to their shapes, the robustness and generalization of these methods may not be optimal.
\par
\textbf{Anchor-Based Methods.} These methods are inspired by general object detection models, such as YOLO \cite{yolov10} and Faster R-CNN \cite{fasterrcnn}, for lane detection. The earliest work is Line-CNN, which utilizes line anchors designed as rays emitted from the three edges (left, bottom, and right) of an image. However, the model’s receptive field is limited to the edges, rendering it suboptimal for capturing the entirety of the lane. LaneATT \cite{laneatt} improves upon this by employing anchor-based feature pooling to aggregate features along the entire line anchor, achieving faster speeds and better performance. Nevertheless, its grid sampling strategy and label assignment still pose limitations. A key advantage of the anchor-based methods is their flexibility, allowing the integration of strategies from anchor-based object detection. For example, \textit{CLRNet} \cite{clrnet} enhances the performance with \textit{cross-layer refinement strategies}, \textit{SimOTA label assignment} \cite{yolox}, and \textit{LIOU loss}, outperforming many previous methods. They also have some essential drawbacks, \textit{e.g.}, lane anchors are often handcrafted and numerous. Some approaches, such as \textit{ADNet} \cite{adnet}, \textit{SRLane} \cite{srlane}, and \textit{Sparse Laneformer} \cite{sparse}, attempt to reduce the number of anchors and provide more flexible proposals; however, this can slightly impact performance. Additionally, methods such as \cite{adnet}\cite{clrernet} still rely on NMS post-processing, complicating NMS threshold settings and model deployment. Although one-to-one label assignment during training, without NMS \cite{detr}\cite{o2o} during evaluation, alleviates this issue, its performance is still less satisfactory compared to NMS-based models.
\par
\textbf{NMS-free Methods.} Due to the threshold sensitivity and computational overhead of NMS, many studies attempt to NMF-free methods or models that do not use NMS during the detection process. For example, \textit{DETR} \cite{detr} employs one-to-one label assignment to avoid redundant predictions without using NMS. Other NMS-free methods \cite{yolov10}\cite{learnNMS}\cite{date} have also been proposed to addressing this issue from two aspects: \textit{model architecture} and \textit{label assignment}. For example, studies in \cite{yolov10}\cite{date} suggest that one-to-one assignments are crucial for NMS-free predictions, but maintaining one-to-many assignments is still necessary to ensure effective feature learning of the model. While some works in \cite{o3d} \cite{relationnet} consider the model’s expressive capacity to provide non-redundant predictions. However, compared to the extensive studies conducted in general object detection, there has been limited research analyzing the NMS-free paradigm. 
\par
In this work, we aim to address the above two issues in the framework of anchor-based lane detection to achieve NMF-free and non-redundant lane predictions.
\section{Polar R-CNN}
\begin{figure}[t]
	\centering
	\def\subwidth{0.24\textwidth}
	\def\imgwidth{\linewidth}
	\def\imgheight{0.4\linewidth}
	
	\begin{subfigure}{\subwidth}
		\includegraphics[width=\imgwidth]{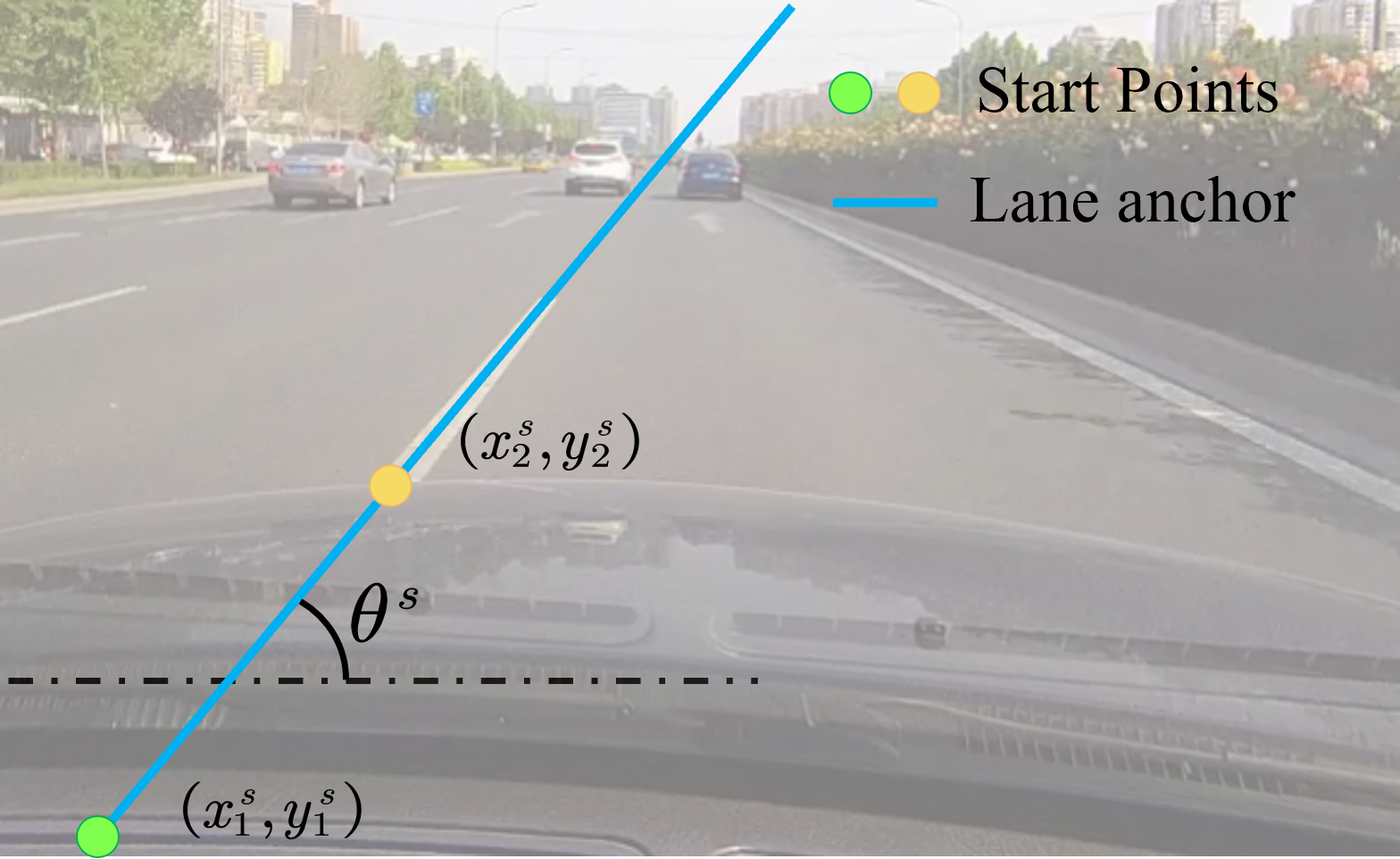}
		\caption{}
	\end{subfigure}
	\begin{subfigure}{\subwidth}
		\includegraphics[width=\imgwidth]{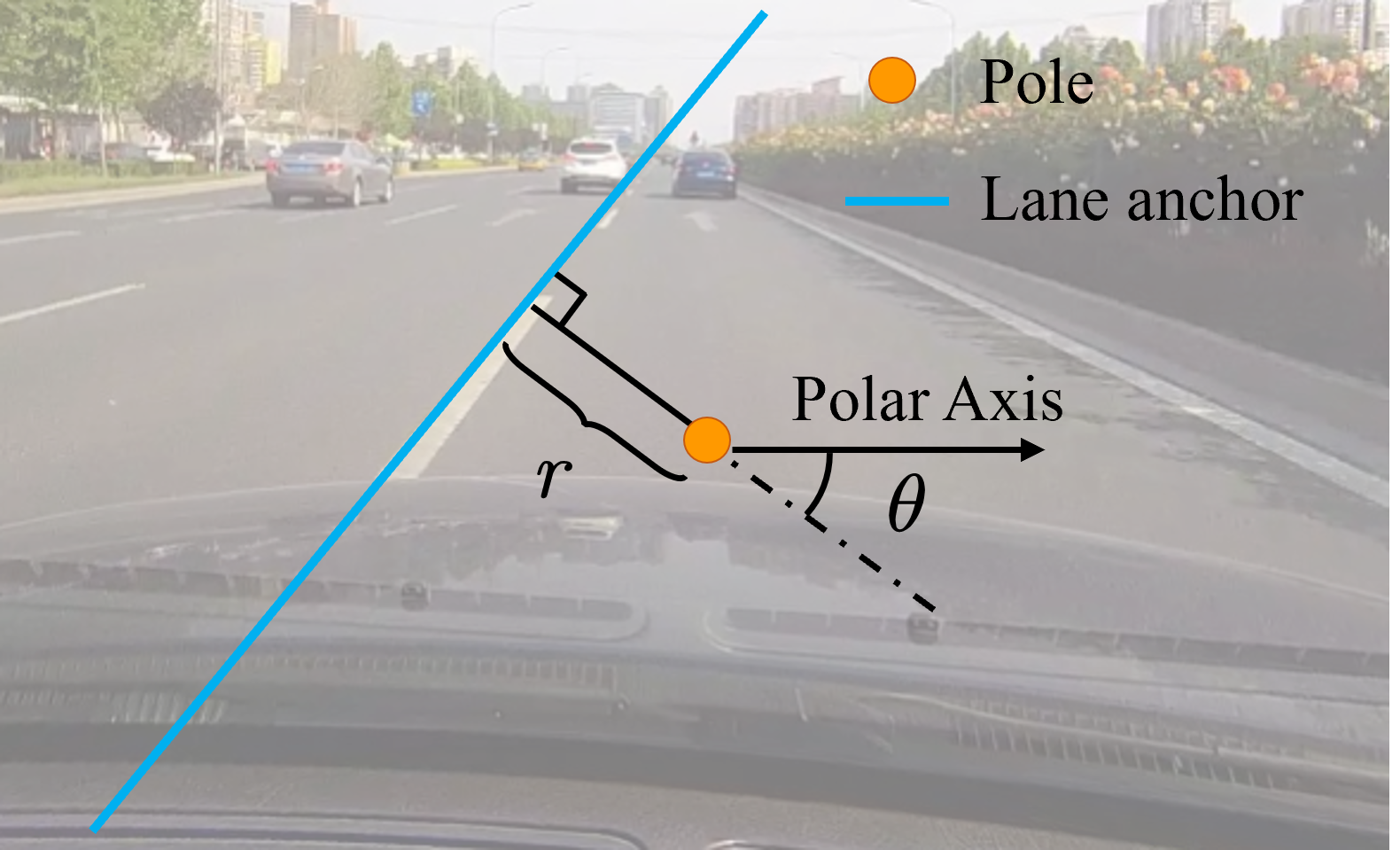}
		\caption{}
	\end{subfigure}
	\caption{Different descriptions for anchor parameters: (a) Ray: defined by its start point (\textit{e.g.}, the green point $\left( x_{1}^{s},y_{1}^{s} \right)$ or the yellow point $\left( x_{2}^{s},y_{2}^{s} \right) $) and direction $\theta^{s}$. (b) Polar: defined by its radius $r$ and angle $\theta$.} 
	\label{coord}
\end{figure}
The overall architecture of our Polar R-CNN is illustrated in Fig. \ref{overall_architecture}. As shown in this figure, our Polar R-CNN for lane detection has a parallel  pipeline with Faster R-CNN \cite{fasterrcnn}, which consists of a backbone\cite{resnet}, a \textit{Feature Pyramid Network} (FPN) \cite{fpn}, a \textit{Local Polar Module} (LPM) as the \textit{Region Proposal Network} (RPN) \cite{fasterrcnn}, and a \textit{Global Polar Module} (GPM) as the \textit{Region of Interest} (RoI) \cite{fasterrcnn} pooling module. In the following, we first introduce the polar coordinate representation of lane anchors, and then present the designed LPM and GPM in our Polar R-CNN.

\subsection{Representation of Lane and Lane Anchor}
Lanes are characterized by their thin, elongated, and curved shapes. A well-defined lane prior aids the model in feature extraction and location prediction. 
\par
\textbf{Lane and Anchor Representation as Ray.} Given an input image with dimensions of width $W$ and height $H$, a lane is represented by a set of 2D points $X=\{(x_1,y_1),(x_2,y_2),\cdots,(x_N,y_N)\}$ with equally spaced y-coordinates, \textit{i.e.}, $y_i=i\times\frac{H}{N}$, where $N$ is the number of data points. Since the y-coordinate is fixed, a lane can be uniquely defined by its x-coordinates. Previous studies \cite{linecnn}\cite{laneatt} have introduced \textit{lane priors}, also known as \textit{lane anchors}, which are represented as straight lines in the image plane and served as references. From a geometric perspective, a lane anchor can be viewed as a ray defined by a start point $(x^{s},y^{s})$ located at the edge of an image (left/bottom/right boundaries), along with a direction $\theta^s$. The primary task of a lane detection model is to estimate the x-coordinate offset from the lane anchor to the ground truth of the lane instance. 
\par
However, the representation of lane anchors as rays presents certain limitations. Notably, a lane anchor can have an infinite number of potential start points, which makes the definition of its start point ambiguous and subjective. As illustrated in Fig. \ref{coord}(a), the studies in \cite{dalnet}\cite{laneatt}\cite{linecnn} define the start points as being located at the boundaries of an image, such as the green point in Fig. \ref{coord}(a). In contrast, the research presented in \cite{adnet} defines the start points, exemplified by the purple point in Fig. \ref{coord}(a), based on their actual visual locations within the image. Moreover, occlusion and damage to the lane significantly affect the detection of these start points, highlighting the need for the model to have a large receptive field \cite{adnet}. Essentially, a straight lane has two degrees of freedom: the slope and the intercept, under a Cartesian coordinate system, implying that the lane anchor could be described using just two parameters instead of the three redundant parameters (\textit{i.e.}, two for the start point and one for the direction) employed in ray representation.
\begin{figure}[t]
	\centering
	\includegraphics[width=0.87\linewidth]{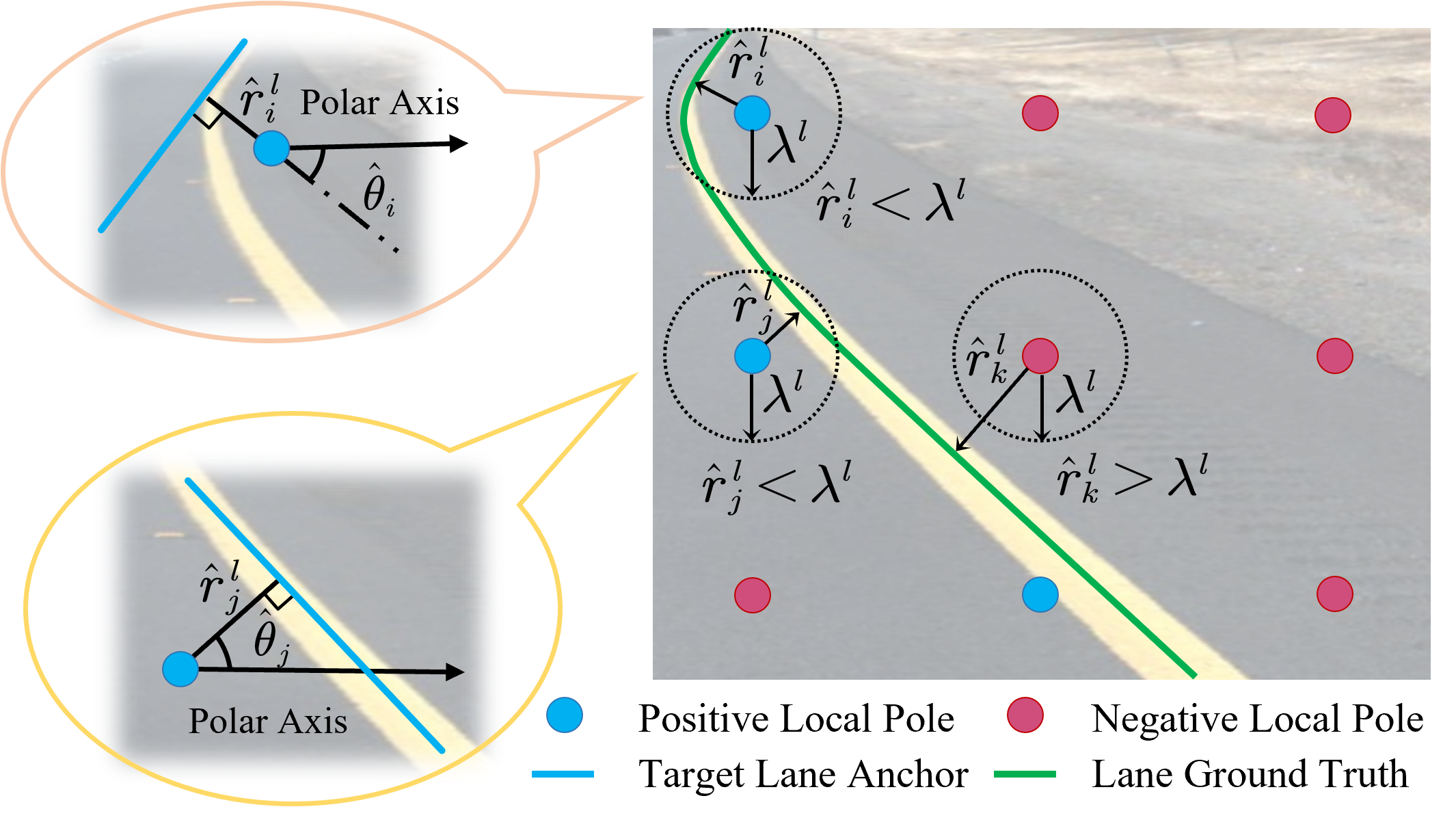}
	\caption{The local polar coordinate system. The ground truth of the radius $\hat{r}_{i}^{l}$ of the $i$-th local pole is defines as the minimum distance from the pole to the lane curve instance. A positive pole has a radius $\hat{r}_{i}^{l}$ that is below a threshold $\lambda^{l}$, and vice versa. Additionally, the ground truth angle $\hat{\theta}_i$ is determined by the angle formed between the radius vector (connecting the pole to the closest point on the lanes) and the polar axis.}
	\label{lpmlabel}
\end{figure}
\par
\textbf{Representation in Polar Coordinate.} As stated above, lane anchors represented by rays have some drawbacks. To address these issues, we introduce a polar coordinate representation of lane anchors. In mathematics, the polar coordinate is a two-dimensional coordinate system in which each point on a plane is determined by a distance from a reference point (\textit{i.e.}, pole) and an angle $\theta$ from a reference direction (\textit{i.e.}, polar axis). As shown in Fig. \ref{coord}(b), given a polar corresponding to the yellow point, a lane anchor for a straight line can be uniquely defined by two parameters: the radial distance from the pole (\textit{i.e.}, radius), $r$, and the counterclockwise angle from the polar axis to the perpendicular line of the lane anchor, $\theta$, with $r \in \mathbb{R}$ and $\theta\in\left(-\frac{\pi}{2}, \frac{\pi}{2}\right)$. 
\par
To better leverage the local inductive bias properties of CNNs, we define two types of polar coordinate systems: the local and global coordinate systems. The local polar coordinate system is to generate lane anchors, while the global coordinate system expresses these anchors in a form within the entire image and regresses them to the ground truth lane instances. Given the distinct roles of the local and global systems, we adopt a two-stage framework for our Polar R-CNN, similar to Faster R-CNN\cite{fasterrcnn}.  
\par
The local polar system is designed to predict lane anchors adaptable to both sparse and dense scenarios. In this system, there are many poles with each as the lattice point of the feature map, referred to as local poles. As illustrated on the left side of Fig. \ref{lpmlabel}, there are two types of local poles: positive and negative. Positive local poles (\textit{i.e.}, the blue points) have a radius $r_{i}^{l}$ below a threshold $\lambda^l$, otherwise, they are classified as negative local poles (\textit{i.e.}, the red points). Each local pole is responsible for predicting a single lane anchor. While a lane ground truth may generate multiple lane anchors, as shown in Fig. \ref{lpmlabel}, there are three positive poles around the lane instance (green lane), which are expected to generate three lane anchors. 

\par
In the local polar coordinate system, the parameters of each lane anchor are determined based on the location of its corresponding local pole. However, in practical terms, once a lane anchor is generated, its definitive position becomes immutable and independent of its original local pole. To simplify the representation of lane anchors in the second stage of Polar R-CNN, a global polar system has been designed, featuring a singular and unified pole that serves as a reference point for the entire image. The location of this global pole is manually set, and in this case, it is positioned near the static \textit{vanishing point} observed across the entire lane image dataset \cite{vanishing}. This approach ensures a consistent and unified polar coordinate for expressing lane anchors within the global context of the image, facilitating accurate regression to the ground truth lane instances.

\begin{figure}[t]
        \centering
        \includegraphics[width=0.45\textwidth]{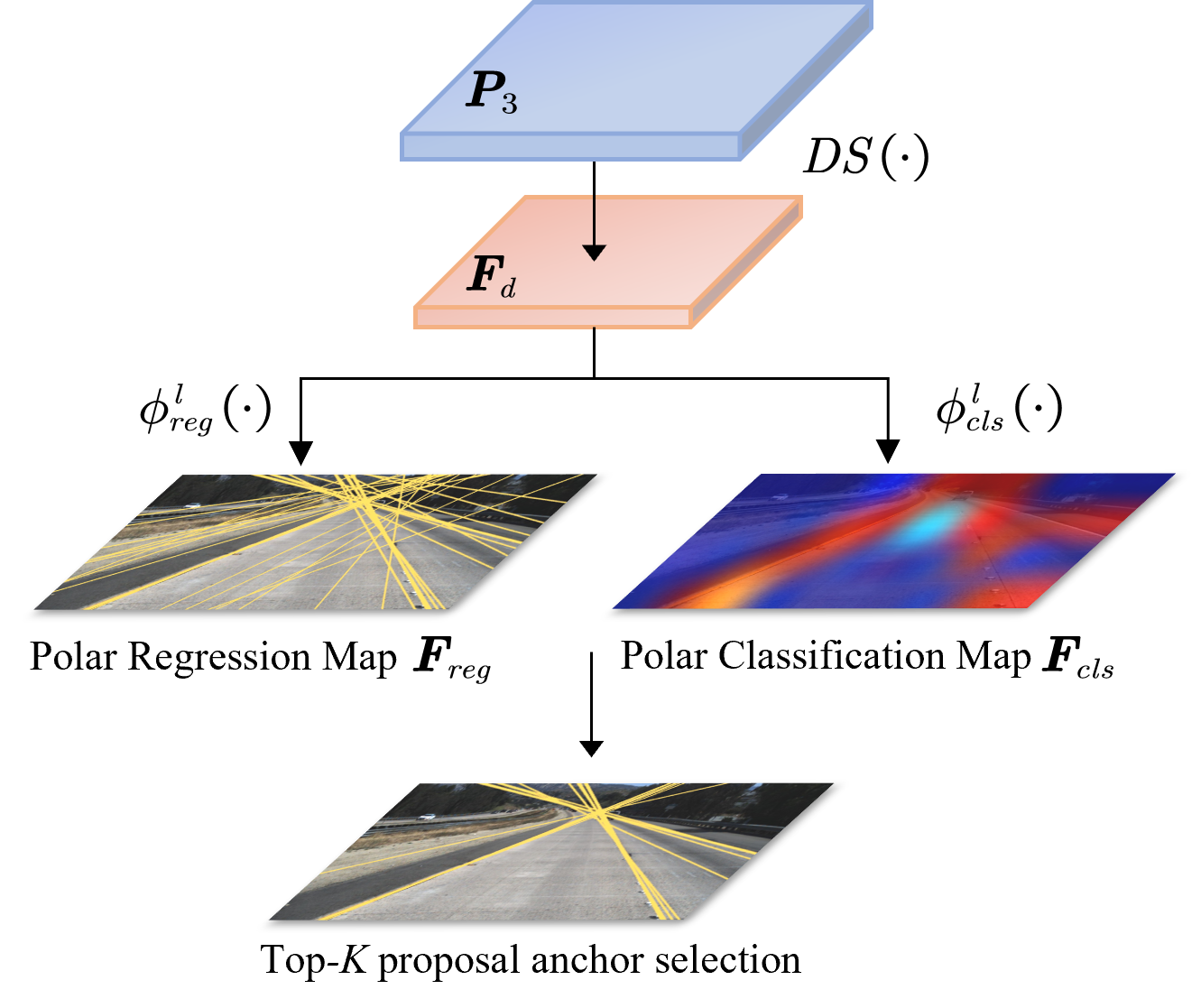}
        \caption{An illustration of the structure of LPM.}
        \label{lpm}
\end{figure}
\subsection{Local Polar Module}
As shown in Fig. \ref{overall_architecture}, three levels of feature maps, denoted as $\boldsymbol{P}_1, \boldsymbol{P}_2, \boldsymbol{P}_3$, are extracted using a \textit{Feature Pyramid Network} (FPN). To generate high-quality anchors around the lane ground truths within an image, we introduce the \textit{Local Polar Module} (LPM), which takes the highest feature map $\boldsymbol{P}_3\in\mathbb{R}^{C_{f} \times H_{f} \times W_{f}}$  as input and outputs a set of lane anchors along with their confidence scores. As demonstrated in Fig. \ref{lpm}, it undergoes a \textit{downsampling} operation $DS(\cdot)$ to produce a lower-dimensional feature map of a size $H^l\times W^l$:
\begin{equation}
		\boldsymbol{F}_d\gets DS\left( \boldsymbol{P}_{3} \right)\ \text{and}\ \boldsymbol{F}_d\in \mathbb{R} ^{C_f\times H^{l}\times W^{l}}.
\end{equation}
The downsampled feature map $\boldsymbol{F}_d$ is then fed into two branches: a \textit{regression} branch $\phi _{reg}^{l}\left(\cdot \right)$ and a \textit{classification} branch $\phi _{cls}^{l}\left(\cdot \right)$, \textit{i.e.},
\begin{align}
\boldsymbol{F}_{reg}&\gets \phi _{reg}^{l}\left( \boldsymbol{F}_d \right),\,\,\boldsymbol{F}_{reg\,\,}\in \mathbb{R} ^{2\times H^{l}\times W^{l}},\\
\boldsymbol{F}_{cls}&\gets \phi _{cls}^{l}\left( \boldsymbol{F}_d \right),\,\,\boldsymbol{F}_{cls}\in \mathbb{R} ^{H^{l}\times W^{l}}.	\label{lpm equ}
\end{align}
The regression branch consists of a single $1\times1$ convolutional layer and with the goal of generating lane anchors by outputting their angles $\theta_j$ and the radius $r^{l}_{j}$, \textit{i.e.}, $\boldsymbol{F}_{reg\,\,} \equiv \left\{\theta_{j}, r^{l}_{j}\right\}_{j=1}^{H^{l}\times W^{l}}$, in the defined local polar coordinate system previously introduced. Similarly, the classification branch $\phi _{cls}^{l}\left(\cdot \right)$ only consists of two $1\times1$ convolutional layers for simplicity. This branch is to predict the confidence heat map $\boldsymbol{F}_{cls\,\,}\equiv \left\{ s_j^l \right\} _{j=1}^{H^l\times W^l}$ for local poles, each associated with a feature point. By discarding local poles with lower confidence, the module increases the likelihood of selecting potential positive foreground lane anchors while effectively removing background lane anchors.
\par
\textbf{Loss Function for LPM.} To train the LPM, we define the ground truth labels for each local pole as follows: the ground truth radius, $\hat{r}^l_i$, is set to be the minimum distance from a local pole to the corresponding lane curve, while the ground truth angle, $\hat{\theta}_i$, is set to be the orientation of the vector extending from the local pole to the nearest point on the curve. Consequently, we have a label set of local poles $\hat{\boldsymbol{F}}_{cls}=\{\hat{s}_j^l\}_{j=1}^{H^l\times W^l}$, where $\hat{s}_j^l=1$ if the $j$-th local pole is positive and $\hat{s}_j^l=0$ if it is negative. Once the regression and classification labels are established, as shown in Fig. \ref{lpmlabel}, LPM can be trained using the $Smooth_{L1}$ loss $S_{L1}\left(\cdot \right)$ for regression branch and the \textit{Binary Cross-Entropy} loss $BCE\left( \cdot , \cdot \right)$ for classification branch. The loss functions for LPM are given as follows:
\begin{align}
\mathcal{L} ^{l}_{cls}&=BCE\left( \boldsymbol{F}_{cls},\hat{\boldsymbol{F}}_{cls} \right),\\
\mathcal{L} _{reg}^{l}&=\frac{1}{N_{pos}^{l}}\sum_{j\in \left\{ j|\hat{r}_{j}^{l}<\lambda^l \right\}}{\left( S_{L1}\left( \theta _{j}^{l}-\hat{\theta}_{j}^{l} \right) +S_{L1}\left( r_{j}^{l}-\hat{r}_{j}^{l} \right) \right)},
\label{loss_lph}
\end{align}
where $N^{l}_{pos}=\left|\{j|\hat{r}_j^l<\lambda^{l}\}\right|$ is the number of positive local poles in LPM.
\par
\textbf{Top-$K$ Anchor Selection.} As discussed above, all $H^{l}\times W^{l}$ anchors, each associated with a local pole in the feature map, are all considered as candidates during the training stage. However, some of these anchors serve as background anchors. We select $K$ anchors with the top-$K$ highest confidence scores as the foreground candidates to feed into the second stage (\textit{i.e.}, global polar module). During training, all anchors are chosen as candidates, where $K=H^{l}\times W^{l}$. It assists \textit{Global Polar Module} (the second stage) in learning from a diverse range of features, including various negative background anchor samples. Conversely, during the evaluation stage, some anchors with lower confidence can be excluded such that $K\leq H^{l}\times W^{l}$. This strategy effectively filters out potential negative anchors and reduces the computational complexity of the second stage. By doing so, it maintains the adaptability and flexibility of anchor distribution while decreasing the total number of anchors especially in the sparse scenarios. The following experiments will demonstrate the effectiveness of different top-$K$ anchor selection strategies.

\begin{figure}[t]
	\centering
	\includegraphics[width=\linewidth]{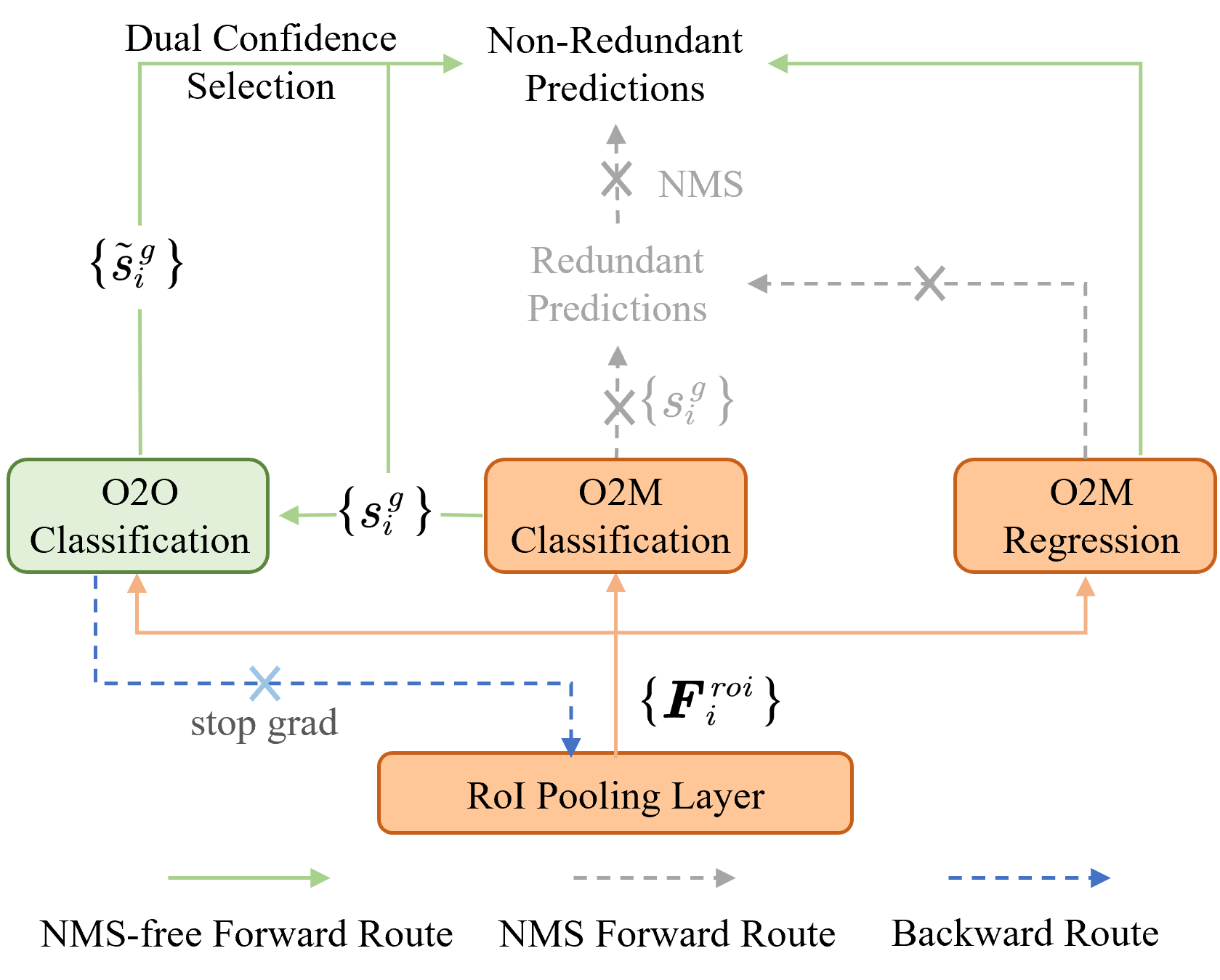} 
	\caption{The pipeline of GPM integrates a RoI Pooling Layer with a triplet head. The triplet head comprises three components: the O2O classification, the O2M classification, and the O2M regression. The dashed path with ``$\times$'' indicates that NMS is no longer necessary. Both sets of $\left\{s_i^g\right\}$ and $\left\{\tilde{s}_i^g\right\}$ participate in the process of selecting the ultimate non-redundant outcomes, a procedure referred to as dual confidence selection. During the backward training phase, the gradients from the O2O classification (the blue dashed route with  ``$\times$'') are stopped.}
        \label{gpm}
\end{figure}

\subsection{Global Polar Module}
 We introduce a novel \textit{Global Polar Module} (GPM) as the second stage to achieve final lane prediction. As illustrated in Fig. \ref{overall_architecture}, GPM takes features samples from anchors proposed by LPM and provides the precise location and confidence scores of final lane detection results. The overall architecture of GPM is illustrated in the Fig. \ref{gpm}. As we can see from this figure, there are two components: a \textit{ROI pooling layer} and a \textit{triplet head}, which will be detailed as follows.
\par
\textbf{RoI Pooling Layer.} It is designed to sample features for lane anchors from feature maps. For ease of the sampling operation, we first transform the radius of the positive lane anchors in a local polar coordinate, $r_j^l$, into the equivalent in a global polar coordinate system, $r_j^g$, by the following equation:
\begin{align}
	r_{j}^{g}&=r_{j}^{l}+\left[ \cos \theta _j;\sin \theta _j \right] ^T\left( \boldsymbol{c}_{j}^{l}-\boldsymbol{c}^g \right), \label{l2g}\\
	j &= 1, 2, \cdots, K, \notag
\end{align}
where $\boldsymbol{c}^{g} \in \mathbb{R}^{2}$ and $\boldsymbol{c}^{l}_{j} \in \mathbb{R}^{2}$ represent the Cartesian coordinates of the global pole and the $j$-th local pole, respectively. It is noteworthy that the angle $\theta_j$ remains unaltered, as the local and global polar coordinate systems share the same polar axis. And next, the feature points are sampled on each lane anchor as follows:
\begin{align}
        x_{i,j}^{s}&=-y_{i,j}^{s}\tan \theta _j+\frac{r_{j}^{g}+\left[ \cos \theta _j;\sin \theta _j \right] ^T\boldsymbol{c}^g}{\cos \theta _j},\label{positions}\\
        i&=1,2,\cdots,N;j=1,2,\cdots,K,\notag
\end{align}
where the y-coordinates $\boldsymbol{y}_{j}\equiv \{y_{1,j},y_{2,j},\cdots ,y_{N,j}\}$ of the $j$-th lane anchor are uniformly sampled vertically from the image, as previously mentioned. The proof of Eqs. (\ref{l2g})-(\ref{positions}) can be found in Appendix \textcolor{red}{A}. Then coordinates of the $j$-th lane anchor can be given by $\boldsymbol{\ell}_j=\{\boldsymbol{x}_{j},\boldsymbol{y}_j\}\equiv \left\{(x_{1,j},y_{1,j}),(x_{2,j},y_{2,j}),\cdots ,(x_{N,j}, y_{N,j})\right\}$.
\par
Given the different level feature maps $\boldsymbol{P}_1, \boldsymbol{P}_2, \boldsymbol{P}_3$ from FPN, we can extract the channel-wise features of each point corresponding to the positions of $\{(x_{1,j},y_{1,j}),(x_{2,j},y_{2,j}),\cdots,(x_{N,j},y_{N,j})\}_{j=1}^{K}$, respectively denoted as $\boldsymbol{F}_{1,j}, \boldsymbol{F}_{2,j}, \boldsymbol{F}_{3,j}\in \mathbb{R} ^{N\times C_f}$. To enhance representation, similar to \cite{srlane}, we employ a weighted sum strategy to combine features from the three levels by:
\begin{align}
\boldsymbol{F}^s_j=\sum_{k=1}^3{\frac{e^{\boldsymbol{w}_{k}}}{\sum_{k=1}^3{e^{\boldsymbol{w}_{k}}}}\circ \boldsymbol{F}_{k,j}},
\end{align}
where $\boldsymbol{w}_{k}\in \mathbb{R}^{N}$ represents trainable aggregate weight ascribed to $N$ sampled points, and the symbol ``$\circ$'' represents element-wise multiplication (\textit{i.e.}, Hadamard product). Instead of concatenating the three sampling features into $\boldsymbol{F}^s_j\in \mathbb{R} ^{N\times 3C_f}$ directly, the adaptive summation significantly reduces the feature dimensions to $\boldsymbol{F}^s_j\in \mathbb{R} ^{N\times C_f}$, which is one-third of the initial dimension. The weighted sum of the tensors is flattened into a vector $\widehat{\boldsymbol{F}}^s_j\in \mathbb{R} ^{NC_f}$, and then integrated through a linear transformation:
\begin{align}
        \boldsymbol{F}_{j}^{roi}\gets \boldsymbol{W}_{pool}\widehat{\boldsymbol{F}}_{j}^{s},\quad j=1,2,\cdots,K.
\end{align}
Here, $\boldsymbol{W}_{pool}\in \mathbb{R} ^{d_r\times NC_f}$ is employed to further reduce the dimension of integrated feature $\widehat{\boldsymbol{F}}_{j}^{s}$, thereby yielding the final RoI features $\{\boldsymbol{F}_{i}^{roi}\in \mathbb{R} ^{d_r}\}_{i=1}^K$, which are feed to the following triplet head.
\par
\textbf{Triplet Head.} The lane detection head is to classify and regress the lane anchors generated from the LPM based on the ROI pooling features $\{\boldsymbol{F}_{i}^{roi}\in \mathbb{R} ^{d_r}\}_{i=1}^K$. As we know, traditional lane detection head\cite{laneatt} is usually equipped with a \textit{One-to-Many} (O2M) classification subhead and a \textit{One-to-Many} (O2M) regression subhead. However, the one-to-many mechanism (\textit{i.e.}, \textit{many candidates for one ground truth}) will cause redundant predictions for each lane, thus need the NMS post-processing operator. While the NMS is non-differentiable and non-end-to-end, resulting in the challenges of manually setting of hyperparameters and suboptimal of performance. To eliminate NMS post-processing while achieving end-to-end learning, we introduce a triplet head module for lane detection. 
\par
As shown in Fig. \ref{gpm}, the triplet head consists of three components: the O2M classification, the O2M regression, and another \textit{One-to-One} (O2O) classification. The features of each lane anchor $\{\boldsymbol{F}_{j}^{roi}\}$ are fed into the aforementioned three subheads, respectively. To keep both simplicity and efficiency, both the O2M classification and O2M regression subheads apply two \textit{multi-layer perceptions} (MLPs) to $\{\boldsymbol{F}_{j}^{roi}\}$ and then generate the confidence scores $\left\{{s}_j^g\right\}$ by the O2M classification subhead and the x-coordinate offsets $\{\Delta\boldsymbol{x}_j\}$ by the O2M regression subhead for each lane anchor. More details about the O2M classification and O2M regression subheads can be referred to \cite{yolox}. The O2O classification subhead is introduced to generate non-redundant lane candidates within a NMS-free paradigm. However, the direct use of one-to-one strategy (\textit{i.e.}, \textit{assigning one positive anchor for one ground truth lane}) based on the extracted features will damage model's learning\cite{dualassign}\cite{yolov10}. Instead, the proposed O2O classification subhead considers both the \textit{confidence prior} $\left\{{s}_j^g\right\}$ of O2M classification subhead and the \textit{spatial geometric prior} of the polar parameters (\textit{i.e.}, the angle $\theta$ and the radius $r$), and apply these priors to adaptively refine the lane anchor features $\{\boldsymbol{F}_{j}^{roi}\}$, which generates the refined lane anchor features $\{\boldsymbol{D}_{j}^{roi}\}$ and the confidence scores $\left\{\tilde{s}_j^g\right\}$. The structural design draws inspiration from the Fast NMS \cite{yolact}, with further particulars accessible in Appendix \textcolor{red}{B}.
\par 
More specifically, the O2O classification subhead first calculates the \textit{semantic distance} between the $i$-th anchor with its x-coordinate $\boldsymbol{x}_{i}$ and the $j$-th anchor with its x-coordinate $\boldsymbol{x}_{j}$ as follows:
\begin{align}
\widehat{\boldsymbol{F}}_{i}^{roi}&\gets \mathrm{ReLU}\left( \boldsymbol{W}_{roi}\boldsymbol{F}_{i}^{roi}+\boldsymbol{b}_{roi} \right), i=1,\cdots,K,\label{edge_layer_1}\\
\boldsymbol{F}_{ij}^{edge}&\gets \boldsymbol{W}_{in}\widehat{\boldsymbol{F}}_{j}^{roi}-\boldsymbol{W}_{out}\widehat{\boldsymbol{F}}_{i}^{roi},\label{edge_layer_2}\\
\boldsymbol{D}_{ij}^{edge}&\gets \mathrm{MLP}_{edge}\left(\boldsymbol{F}_{ij}^{edge}+\boldsymbol{W}_s\left( \boldsymbol{x}_{j}-\boldsymbol{x}_{i} \right) +\boldsymbol{b}_s \right),\label{edge_layer_3}
\end{align}
where $\boldsymbol{D}_{ij}^{edge}\in \mathbb{R}^{d_n}$ denotes the implicit semantic distance between the $i$-th prediction and the $j$-th predictions. $\mathrm{ReLU}$ is the ReLU activation function, $\mathrm{MLP}_{edge}$ denotes a two-layer MLP operator, and $\{\boldsymbol{W}_{roi}, \boldsymbol{W}_{in}, \boldsymbol{W}_{out}, \boldsymbol{W}_s, \boldsymbol{b}_{roi}, \boldsymbol{b}_s\}$ are the model parameters to be trained. However, it is still difficult to make a one-to-one assignment for the ground truth instance based on above semantic distance, as some forked anchors may have similar distances. To increase the semantic distance gaps among anchors, we need to suppress the features of similar or overlapped anchors. Based on this, we have designed a adjacency matrix $\boldsymbol{A}\in\mathbb{R}^{K\times K}$ defined as 
\begin{equation}
\boldsymbol{A}=\boldsymbol{A}^C\odot\boldsymbol{A}^G,
\end{equation}
where $\odot$ is the element-wise multiplication, $\boldsymbol{A}^C\in\mathbb{R}^{K\times K}$ and $\boldsymbol{A}^G\in\mathbb{R}^{K\times K}$ are the confidence-prior adjacency matrix and the geometric-prior adjacency matrix, respectively. The confidence-prior adjacency matrix $\boldsymbol{A}^C=\left(A_{ij}^C\right)_{i,j=1}^K$ is defined as follows:
\begin{align}
A_{ij}^{C}=\begin{cases}
1,\, \mathrm{if}\,\,s_i^g>s_j^g\,\,or\,\,( s_i^g=s_j^g\,\,and\,\,i>j );\\
0,\,\mathrm{others}.
\end{cases}
\label{confidential matrix1}
\end{align}
Here, $s_i^g$ and $s_j^g$ are the confidence scores corresponding to the $i$-th and the $j$-th lane anchors and predicted by the O2M classification subhead. According to Eq. \eqref{confidential matrix1}, the role of $\boldsymbol{A}^C$ is to allow lane anchors with higher confidence scores to suppress those with lower scores. In order to leverage geometric priors and based on the representation in polar coordinate (\textit{i.e}, the global polar radius $r^g$ and angle $\theta$), we further introduce geometric-prior adjacency matrix $\boldsymbol{A}^G=\left(A_{ij}^G\right)_{i,j=1}^K$, defined by
\begin{align}
	A_{ij}^{G}=\begin{cases}
		1,\, \mathrm{if}\,\,\left| \theta _i-\theta _j \right|<\tau^{\theta}\,\,and\,\,\left| r_{i}^{g}-r_{j}^{g} \right|<\lambda^g;\\
		0,\,\mathrm{others},
	\end{cases}
	\label{geometric prior matrix1}
\end{align}
where $\tau^{\theta}$ and $\lambda^g$ are the thresholds to measure the geometric distances. Based on the definition of geometric-prior and confidence-prior adjacency matrices, the overall adjacency matrix $\boldsymbol{A}$ can be seen as a directed graph with each lane anchor as a node and the ROI features $\boldsymbol{F}_i^{roi}$ serving as their input features. Specifically, if an element $A_{ij}$ in $\boldsymbol{A}$ equals to 1, a directed edge exists from the $i$-th anchor and the $j$-th anchor, which implies that the $j$-th anchor may be suppressed by the $i$-th anchor when the confidence score of the $i$-th anchor exceeds that of the $j$-th anchor and their geometric distance is sufficiently small (\textit{i.e.}, less than a predefined threshold).
\par
And then, by considering the suppressive effect of the lane anchors induced by the overall adjacency matrix $\boldsymbol{A}$, the lane anchor features $\boldsymbol{F}_j^{roi}$ can be further refined from the semantic distance tensor $\mathcal{D}^{edge}=\{\boldsymbol{D}_{ij}^{edge}\}\in\mathbb{R}^{K\times K\times d_n}$ as follows:
\begin{align}
	\boldsymbol{D}_j^{roi}\in \mathbb{R}^{d_n}\gets\mathrm{MPool}_{col}\left(\mathcal{D}^{edge}(:,j,:)|\boldsymbol{A}(:,j)=1\right),
        \label{maxpooling}
\end{align}
where $j=1,2,\cdots,K$ and $\mathrm{MPool}_{col}(\cdot|\boldsymbol{A}(:,j)=1)$ is an element-wise max pooling operator along the $j$-th column of adjacency matrix $\boldsymbol{A}$ with the element $A_{:j}=1$. This is in inspired by the existing works\cite{o3d}\cite{pointnet}, which aims to extract the most distinctive features from the lane anchors that may potentially suppress the refined lane anchors. With the refined anchor features $\boldsymbol{D}_j^{roi}$, the final confidence scores of the O2O classification subhead are generated by a three-layer MLPs:
\begin{align}
	\tilde{s}_{j}^{g}\gets \mathrm{MLP}_{roi}\left( \boldsymbol{D}_{j}^{roi} \right), j=1,\cdots,K.	\label{node_layer}
\end{align}
As stated above, the O2O classification subhead is formed from Eqs. (\ref{edge_layer_1})-(\ref{node_layer}), which can be seen as a directed graph driven by neural networks, which is referred to as the \textit{graph neural network} (GNN) block.
\par
\textbf{Dual Confidence Selection with NMF-free.} With the help of adjacency matrix $\boldsymbol{A}$, the variability among semantic features $\{\boldsymbol{D}_j^{roi}\}$ has been enlarged, resulting in a significant gap in confidence scores $\{\tilde{s}_{j}^{g}\}$ generated by O2O classification subhead, which makes them easier to distinguish. Therefore, unlike conventional methods that feed the confidence scores $\{\tilde{s}_{j}^{g}\}$ obtained by O2M classification subhead into the NMS post-processing stage to remove redundant candidates, we have implemented the following dual confidence selection criterion for selecting positive anchors:
\begin{align}
	\Omega^{pos}=\left\{i|\tilde{s}_{i}^{g}>\tau_{o2o} \right\} \cap \left\{ i|s_{i}^{g}>\tau_{o2m} \right\},
\end{align}
where $\tau_{o2o}$ and $\tau_{o2m}$ are two confidence thresholds. The $\Omega^{pos}$ can allow for non-redundant positive predictions without NMS post-processing as the O2O classification subhead enhances the confidence score variability among similar anchors, making it less sensitive to the two confidence thresholds.
\par
\textbf{Loss function for GPM.} After obtaining the positive candidate set $\Omega^{pos}$ for the O2O classification subhead, the Hungarian algorithm \cite{detr} is applied to perform label assignment, \textit{i.e.}, a one-to-one assignment between the positive anchors and the ground truth instances. As for the O2M classification and O2M regression subheads, we use the same approach as in SimOTA \cite{yolox} for label assignment. More details about label assignment and cost function can be found in Appendix \textcolor{red}{C} and \textcolor{red}{D}. In the training, the Focal loss \cite{focal} is applied for both O2O classification subhead and the O2M classification subhead, respectively denoted as $\mathcal{L}^{o2o}_{cls}$ and $\mathcal{L}^{o2m}_{cls}$. Furthermore, we adopt the rank loss $\mathcal{L}_{rank}$ \cite{pss} to amplify the disparity between the positive and negative confidences of the O2O classification subhead. Note that, similar to \cite{pss}, we stop the gradient flow from the O2O classification subhead during the training stage to preserve the quality of RoI feature learning.

\begin{figure}[t]
	\centering
	\includegraphics[width=0.85\linewidth]{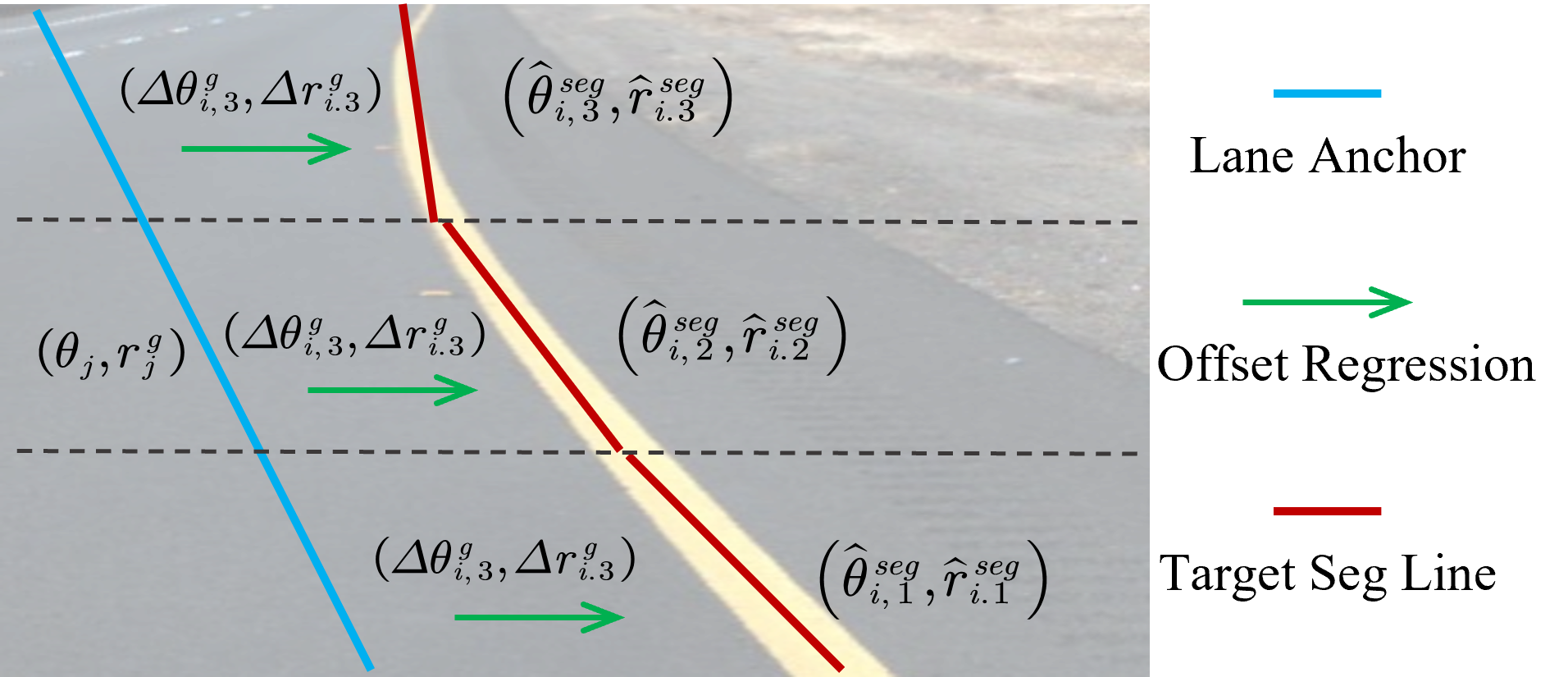} %
	\caption{Auxiliary loss for segment parameter regression. The ground truth lane curve is partitioned into several segments, with the parameters of each segment denoted as $\left( \hat{\theta}_{i,\cdot}^{seg},\hat{r}_{i,\cdot}^{seg} \right)$. The model output the parameter offsets $\left( \varDelta \theta _{j,\cdot},\varDelta r_{j,\cdot}^{g} \right)$ to regress from the original anchor to each target line segments.}
	\label{auxloss}
\end{figure}

To train the O2M regression subhead, we have redefined the GIoU concept (refer to Appendix \textcolor{red}{C} for more details) and adopt the GIoU loss $\mathcal{L}_{GIoU}^{o2m}$ to regress the x-coordinate offsets $\{\Delta\boldsymbol{x}_j\}$ for each positive lane anchor. The end points of lanes are trained with a $Smooth_{L1}$ loss $\mathcal{L}_{end}^{o2m}$. In addition, we propose an auxiliary loss $\mathcal{L}_{aux}$ to facilitate the learning of global features. As illustrated in Fig. \ref{auxloss}, the anchors and ground truth are divided into several segments, with each anchor segment being regressed to the primary components of the corresponding segment of the ground truth. The auxiliary loss $\mathcal{L}_{aux}$ helps the detection head gain a deeper understanding of the global geometric structure and the auxiliary regression branch is dropped during the evaluation stage. Finally, the classification loss $\mathcal{L} _{cls}^{g}$ and the regression loss $\mathcal{L} _{reg}^{g}$ for GPM are given as follows:
\begin{align}
	\mathcal{L} _{cls}^{g}&=w^{o2m}_{cls}\mathcal{L}^{o2m}_{cls}+w^{o2o}_{cls}\mathcal{L}^{o2o}_{cls}+w_{rank}\mathcal{L}_{rank},
	\\
	\mathcal{L} _{reg}^{g}&=w_{GIoU}^{o2m}\mathcal{L}_{GIoU}^{o2m}+w_{end}^{o2m}\mathcal{L}_{end}^{o2m}+w_{aux}\mathcal{L} _{aux},
\end{align}
where $w^{o2m}_{cls}, w^{o2o}_{cls}, w_{rank}, w_{GIoU}^{o2m}, w_{end}^{o2m}, w_{aux}$ are constant weights used for adjusting the effect of the different loss terms.
\subsection{The Overall Loss Function} 
By taking the two stage losses, the overall loss function is given as follows:
\begin{align}
\mathcal{L} =\mathcal{L} _{cls}^{l}+\mathcal{L} _{reg}^{l}+\mathcal{L} _{cls}^{g}+\mathcal{L} _{reg}^{g},
\end{align}
where $\mathcal{L} _{cls}^{l}, \mathcal{L} _{reg}^{l}$ are used to train parameters of classification and regression of LPM in the first stage and $\mathcal{L} _{cls}^{g},\mathcal{L} _{reg}^{g}$ are used to train that of GPM in the second stage. 
\section{Experiment}
\subsection{Dataset and Evaluation Metric}
We conducted experiments on four widely used lane detection benchmarks and one rail detection dataset: CULane\cite{scnn}, TuSimple\cite{tusimple}, LLAMAS\cite{llamas}, CurveLanes\cite{curvelanes}, and DL-Rail\cite{dalnet}. Among these datasets, CULane and CurveLanes are particularly challenging. The CULane dataset consists various scenarios but has sparse lane distributions, whereas CurveLanes includes a large number of curved and dense lane types, such as forked and double lanes. The DL-Rail dataset, focused on rail detection across different scenarios, is chosen to evaluate our model’s performance beyond traditional lane detection. 

We use the F1-score to evaluate our model on the CULane, LLAMAS, DL-Rail, and Curvelanes datasets, maintaining consistency with previous works. The F1-score is defined as follows:
\begin{align}
Pre\,\,&=\,\,\frac{TP}{TP+FP},
\\
Rec\,\,&=\,\,\frac{TP}{TP+FN},
\\
F1&=\frac{2\times Pre\times Rec}{Pre\,\,+\,\,Rec},
\end{align}
where $TP$, $FP$ and $FN$ represent the true positives, false positives, and false negatives of the entire dataset, respectively. In our experiment, we use different IoU thresholds to calculate the F1-score for different datasets: $F1@50$ and $F1@75$ for CULane \cite{clrnet}, $F1@50$ for LLAMAS \cite{clrnet} and Curvelanes \cite{CondLaneNet}, and $F1@50$, $F1@75$, and $mF1$ for DL-Rail \cite{dalnet}. The $mF1$ is defined as:
        \begin{align}
                mF1=\left( F1@50+F1@55+\cdots+F1@95 \right) /10,
        \end{align}
where $F1@50, F1@55, \cdots, F1@95$ are F1 metrics when IoU thresholds are $0.5, 0.55, \cdots, 0.95$,  respectively.
For Tusimple, the evaluation is formulated as follows:
        \begin{align}
                Accuracy=\frac{\sum{C_{clip}}}{\sum{S_{clip}}},
        \end{align}
where $C_{clip}$ and $S_{clip}$ represent the number of correct points (predicted points within 20 pixels of the ground truth) and the ground truth points, respectively. If the accuracy exceeds 85\%, the prediction is considered correct. TuSimple also reports the \textit{False Positive Rate} ($\mathrm{FPR}=1-\mathrm{Precision}$) and \textit{False Negative Rate} ($\mathrm{FNR}=1-\mathrm{Recall}$) metrics. 

\subsection{Implement Detail}
All input images are cropped and resized to $800\times320$. Similar to \cite{clrnet}, we apply random affine transformations and random horizontal flips. For the optimization process, we use the AdamW \cite{adam} optimizer with a learning rate warm-up and a cosine decay strategy. The initial learning rate is set to 0.006. The number of sampled points and regression points for each lane anchor are set to 36 and 72, respectively. The power coefficient of cost function $\beta$ is set to 6. The training processing of the whole model (including LPM and GPM) is end-to-end just like \cite{adnet}\cite{srlane}.  All the experiments are conducted on a single NVIDIA A100-40G GPU. To make our model simple, we only use CNN-based backbone, namely ResNet\cite{resnet} and DLA34\cite{dla}. Other details can be seen in Appendix \textcolor{red}{E}.

\begin{table*}[htbp]
        \centering
        \caption{Comparison results on the CULane test set with other methods.}
        \normalsize
        \begin{adjustbox}{width=\linewidth}
        \begin{tabular}{lrlllllllllll}
        \toprule
        \textbf{Method}& \textbf{Backbone}&\textbf{F1@50}$\uparrow$& \textbf{F1@75}$\uparrow$& \textbf{Normal}$\uparrow$&\textbf{Crowded}$\uparrow$&\textbf{Dazzle}$\uparrow$&\textbf{Shadow}$\uparrow$&\textbf{No line}$\uparrow$& \textbf{Arrow}$\uparrow$& \textbf{Curve}$\uparrow$& \textbf{Cross}$\downarrow$ & \textbf{Night}$\uparrow$  \\
        \hline
        \textbf{Seg \& Grid} \\
        \cline{1-1}
        SCNN\cite{scnn}        &VGG-16   &71.60&39.84&90.60&69.70&58.50&66.90&43.40&84.10&64.40&1990&66.10\\
        RESA\cite{resa}        &ResNet50 &75.30&53.39&92.10&73.10&69.20&72.80&47.70&83.30&70.30&1503&69.90\\
        LaneAF\cite{laneaf}     &DLA34    &77.41&-    &91.80&75.61&71.78&79.12&51.38&86.88&72.70&1360&73.03\\
        UFLDv2\cite{ufldv2}      &ResNet34 &76.0 &-    &92.5 &74.8 &65.5 &75.5 &49.2 &88.8 &70.1 &1910&70.8 \\
        CondLaneNet\cite{CondLaneNet} &ResNet101&79.48&61.23&93.47&77.44&70.93&80.91&54.13&90.16&75.21&1201&74.80\\
        \cline{1-1}
        \textbf{Parameter} \\
        \cline{1-1}
        BézierLaneNet\cite{bezierlanenet} &ResNet34&75.57&-&91.59&73.20&69.20&76.74&48.05&87.16 &62.45&\textbf{888}&69.90\\
        BSNet\cite{bsnet}         &DLA34   &80.28&-&93.87&78.92&75.02&82.52&54.84&90.73&74.71&1485&75.59\\
        Eigenlanes\cite{eigenlanes}    &ResNet50&77.20&-&91.7 &76.0 &69.8 &74.1 &52.2 &87.7 &62.9 &1509&71.8 \\
        \cline{1-1}
        \textbf{Keypoint} \\
        \cline{1-1}
        CurveLanes-NAS-L\cite{curvelanes} &-  &74.80&-&90.70&72.30&67.70&70.10&49.40&85.80&68.40&1746&68.90\\
        FOLOLane\cite{fololane}         &ResNet18 &78.80&-&92.70&77.80&75.20&79.30&52.10&89.00&69.40&1569&74.50\\
        GANet-L\cite{ganet}          &ResNet101&79.63&-&93.67&78.66&71.82&78.32&53.38&89.86&77.37&1352&73.85\\
        \cline{1-1}
        \textbf{Dense Anchor} \\
        \cline{1-1}
        LaneATT\cite{laneatt}  &ResNet18 &75.13&51.29&91.17&72.71&65.82&68.03&49.13&87.82&63.75&1020&68.58\\
        LaneATT\cite{laneatt}  &ResNet122&77.02&57.50&91.74&76.16&69.47&76.31&50.46&86.29&64.05&1264&70.81\\
        CLRNet\cite{clrnet}   &Resnet18 &79.58&62.21&93.30&78.33&73.71&79.66&53.14&90.25&71.56&1321&75.11\\
        CLRNet\cite{clrnet}   &DLA34    &80.47&62.78&93.73&79.59&75.30&82.51&54.58&90.62&74.13&1155&75.37\\
        CLRerNet\cite{clrernet} &DLA34    &81.12&64.07&94.02&80.20&74.41&\textbf{83.71}&56.27&90.39&74.67&1161&\textbf{76.53}\\
        \cline{1-1}
        \textbf{Sparse Anchor} \\
        \cline{1-1}
        ADNet \cite{adnet}            &ResNet34&78.94&-&92.90&77.45&71.71&79.11&52.89&89.90&70.64&1499&74.78\\
        SRLane \cite{srlane}         &ResNet18&79.73&-&93.52&78.58&74.13&81.90&55.65&89.50&75.27&1412&74.58\\
        Sparse Laneformer\cite{sparse} &Resnet50&77.83&-&-    &-    &-    &-    &-    &-    &-    &-   &-    \\
        \hline
        \textbf{Proposed Method} \\
        \cline{1-1}
        Polar R-CNN-NMS   &ResNet18&80.81&63.97&94.12&79.57&76.53&83.33&55.10&90.70&79.50&1088&75.25\\
        Polar R-CNN       &ResNet18&80.81&63.96&94.12&79.57&76.53&83.33&55.06&90.62&79.50&1088&75.25\\
        Polar R-CNN       &ResNet34&80.92&63.97&94.24&79.76&76.70&81.93&55.40&\textbf{91.12}&79.85&1158&75.71\\
        Polar R-CNN       &ResNet50&81.34&64.77&94.45&\textbf{80.42}&75.82&83.61&56.62&91.10&80.05&1356&75.94\\
        Polar R-CNN-NMS   &DLA34   &\textbf{81.49}&64.96&\textbf{94.44}&80.36&\textbf{76.79}&83.68&56.52&90.85&\textbf{80.09}&1133&76.32\\
        Polar R-CNN       &DLA34   &\textbf{81.49}&\textbf{64.97}&\textbf{94.44}&80.36&\textbf{76.79}&83.68&\textbf{56.55}&90.81&\textbf{79.80}&1133&76.33\\
        \bottomrule
        \end{tabular}  
      \end{adjustbox}
        \label{culane result}
\end{table*}

\begin{table}[h]
        \centering
        \caption{Comparison results on the TuSimple test set with other methods.}
        \begin{adjustbox}{width=\linewidth}
        \begin{tabular}{lrcccc}
        \toprule
        \textbf{Method}& \textbf{Backbone}& \textbf{Acc(\%)}&\textbf{F1(\%)}&\textbf{FPR(\%)}&\textbf{FNR(\%)} \\
        \midrule
        SCNN\cite{scnn}      &VGG16         &96.53&95.97&6.17&\textbf{1.80}\\
        PolyLanenet\cite{polylanenet}&EfficientNetB0&93.36&90.62&9.42&9.33\\
        UFLDv2\cite{ufld}     &ResNet34      &88.08&95.73&18.84&3.70\\
        LaneATT\cite{laneatt}    &ResNet34      &95.63&96.77&3.53&2.92\\
        FOLOLane\cite{laneatt}   &ERFNet        &\textbf{96.92}&96.59&4.47&2.28\\
        CondLaneNet\cite{CondLaneNet}&ResNet101     &96.54&97.24&2.01&3.50\\
        CLRNet\cite{clrnet}     &ResNet18      &96.84&97.89&2.28&1.92\\
        \midrule
        Polar R-CNN-NMS &ResNet18&96.21&\textbf{97.98}&2.17&1.86\\
        Polar R-CNN       &ResNet18&96.20&97.94&2.25&1.87\\ 
        \bottomrule
        \end{tabular}
        \end{adjustbox}
        \label{tusimple result}
\end{table}

\begin{table}[h]
        \centering
        \caption{Comparison results on the LLAMAS test set with other methods.}
        \begin{adjustbox}{width=\linewidth}
        \begin{tabular}{lrcccc}
        \toprule
        \textbf{Method}& \textbf{Backbone}&\textbf{F1@50(\%)}&\textbf{Precision(\%)}&\textbf{Recall(\%)} \\
        \midrule
        SCNN\cite{scnn}          &ResNet34&94.25&94.11&94.39\\
        BézierLaneNet\cite{bezierlanenet} &ResNet34&95.17&95.89&94.46\\
        LaneATT\cite{laneatt}       &ResNet34&93.74&96.79&90.88\\
        LaneAF\cite{laneaf}        &DLA34   &96.07&\textbf{96.91}&95.26\\
        DALNet\cite{dalnet}        &ResNet18&96.12&96.83&95.42\\
        CLRNet\cite{clrnet}       &DLA34   &96.12&-    &-    \\
        \midrule
        
        Polar R-CNN-NMS          &ResNet18&96.05&96.80&95.32\\
        Polar R-CNN                &ResNet18&96.06&96.81&95.32\\
        Polar R-CNN-NMS          &DLA34&96.13&96.80&\textbf{95.47}\\
        Polar R-CNN                &DLA34&\textbf{96.14}&96.82&\textbf{95.47}\\

        \bottomrule
        \end{tabular}
        \end{adjustbox}
        \label{llamas result}
\end{table}

    \begin{table}[h]
        \centering
        \caption{Comparison results on the DL-Rail test set with other methods.}
        \begin{adjustbox}{width=\linewidth}
        \begin{tabular}{lrccc}
        \toprule
        \textbf{Method}& \textbf{Backbone}&\textbf{mF1(\%)}&\textbf{F1@50(\%)}&\textbf{F1@75(\%)} \\
        \midrule
        BézierLaneNet\cite{bezierlanenet} &ResNet18&42.81&85.13&38.62\\
        GANet-S\cite{ganet}       &Resnet18&57.64&95.68&62.01\\
        CondLaneNet\cite{CondLaneNet}   &Resnet18&52.37&95.10&53.10\\
        UFLDv1\cite{ufld}       &ResNet34&53.76&94.78&57.15\\
        LaneATT(with RPN)\cite{dalnet} &ResNet18&55.57&93.82&58.97\\
        DALNet\cite{dalnet}      &ResNet18&59.79&96.43&65.48\\
        \midrule
        Polar R-CNN-NMS &ResNet18&\textbf{61.53}&\textbf{97.01}&\textbf{67.86}\\
        Polar R-CNN       &ResNet18&61.52&96.99&67.85\\
        \bottomrule
        \end{tabular}
        \end{adjustbox}
        \label{dlrail result}
    \end{table}

\begin{table}[h]
        \centering
        \caption{Comparison results on the CurveLanes validation set with other methods.}
        \begin{adjustbox}{width=\linewidth}
        \begin{tabular}{lrcccc}
        \toprule
        \textbf{Method}& \textbf{Backbone}&\textbf{F1@50 (\%)}&\textbf{Precision (\%)}&\textbf{Recall (\%)} \\
        \midrule
        SCNN\cite{scnn}          &VGG16    &65.02&76.13&56.74\\
        Enet-SAD\cite{enetsad}      &-        &50.31&63.60&41.60\\
        PointLanenet\cite{pointlanenet}  &ResNet101&78.47&86.33&72.91\\
        CurveLane-S\cite{curvelanes}   &-        &81.12&93.58&71.59\\
        CurveLane-M\cite{curvelanes}   &-        &81.80&93.49&72.71\\
        CurveLane-L\cite{curvelanes}   &-        &82.29&91.11&75.03\\
        UFLDv2\cite{ufldv2}        &ResNet34 &81.34&81.93&80.76\\
        CondLaneNet-M\cite{CondLaneNet} &ResNet34 &85.92&88.29&83.68\\
        CondLaneNet-L\cite{CondLaneNet} &ResNet101&86.10&88.98&83.41\\
        CLRNet\cite{clrnet}        &DLA34    &86.10&91.40&81.39\\
        CLRerNet\cite{clrernet}      &DLA34    &86.47&91.66&81.83\\
        \hline
        Polar R-CNN &DLA34&\textbf{87.29}&90.50&\textbf{84.31}\\
        \hline
        \end{tabular}
        \end{adjustbox}
        \label{curvelanes result}
\end{table}

\subsection{Comparison with the state-of-the-art method}
The comparison results of our proposed model with other methods are shown in Tables \ref{culane result}, \ref{tusimple result}, \ref{llamas result}, \ref{dlrail result}, and \ref{curvelanes result}. We present results for two versions of our model: the NMS-based version, denoted as \textit{Polar R-CNN-NMS}, and the NMS-free version, denoted as \textit{Polar R-CNN}. The NMS-based version utilizes predictions $\left\{s_i^g\right\}$ obtained from the O2M head followed by NMS post-processing, while the NMS-free version derives predictions via dual confidence selection.

To ensure a fair comparison, we also include results for CLRerNet \cite{clrernet} on the CULane and CurveLanes datasets, as we use a similar training strategy and dataset splits. As illustrated in the comparison results, our model demonstrates competitive performance across five datasets. Specifically, on the CULane, TuSimple, LLAMAS, and DL-Rail datasets of sparse scenarios, our model outperforms other anchor-based methods. Additionally, the performance of the NMS-free version is nearly identical to that of the NMS-based version, highlighting the effectiveness of the O2O classification subhead in eliminating redundant predictions in sparse scenarios. On the CurveLanes dataset, the NMS-free version achieves superior F1-measure and Recall compared to other methods.

\begin{figure}[t]
        \centering
        \includegraphics[width=\linewidth]{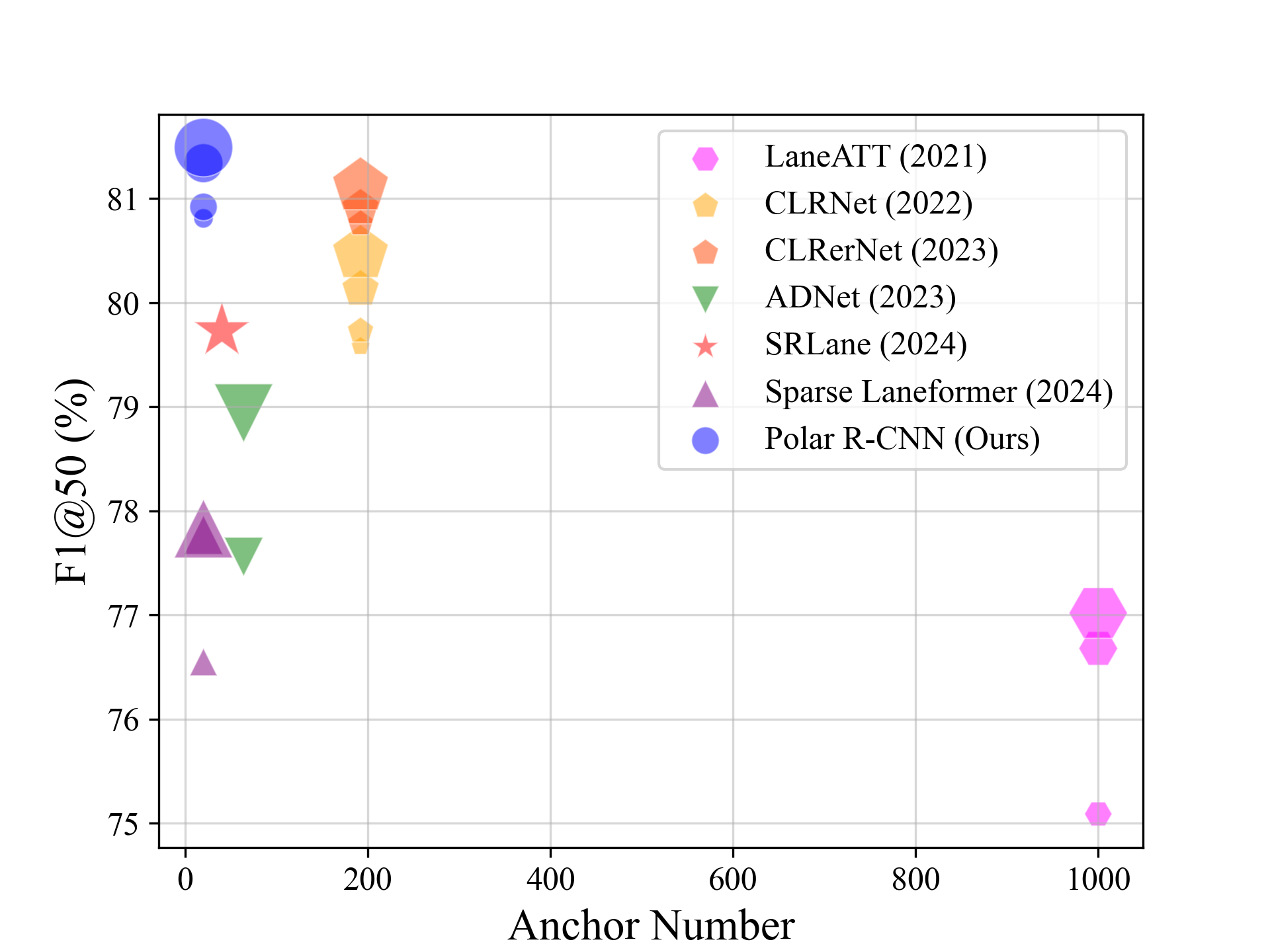}
        \caption{Anchor numbers vs. F1@50 of different methods on CULane lane detection benchmark.}
        \label{anchor_num_method}
\end{figure}

\begin{figure}[t]
        \centering
        \includegraphics[width=\linewidth]{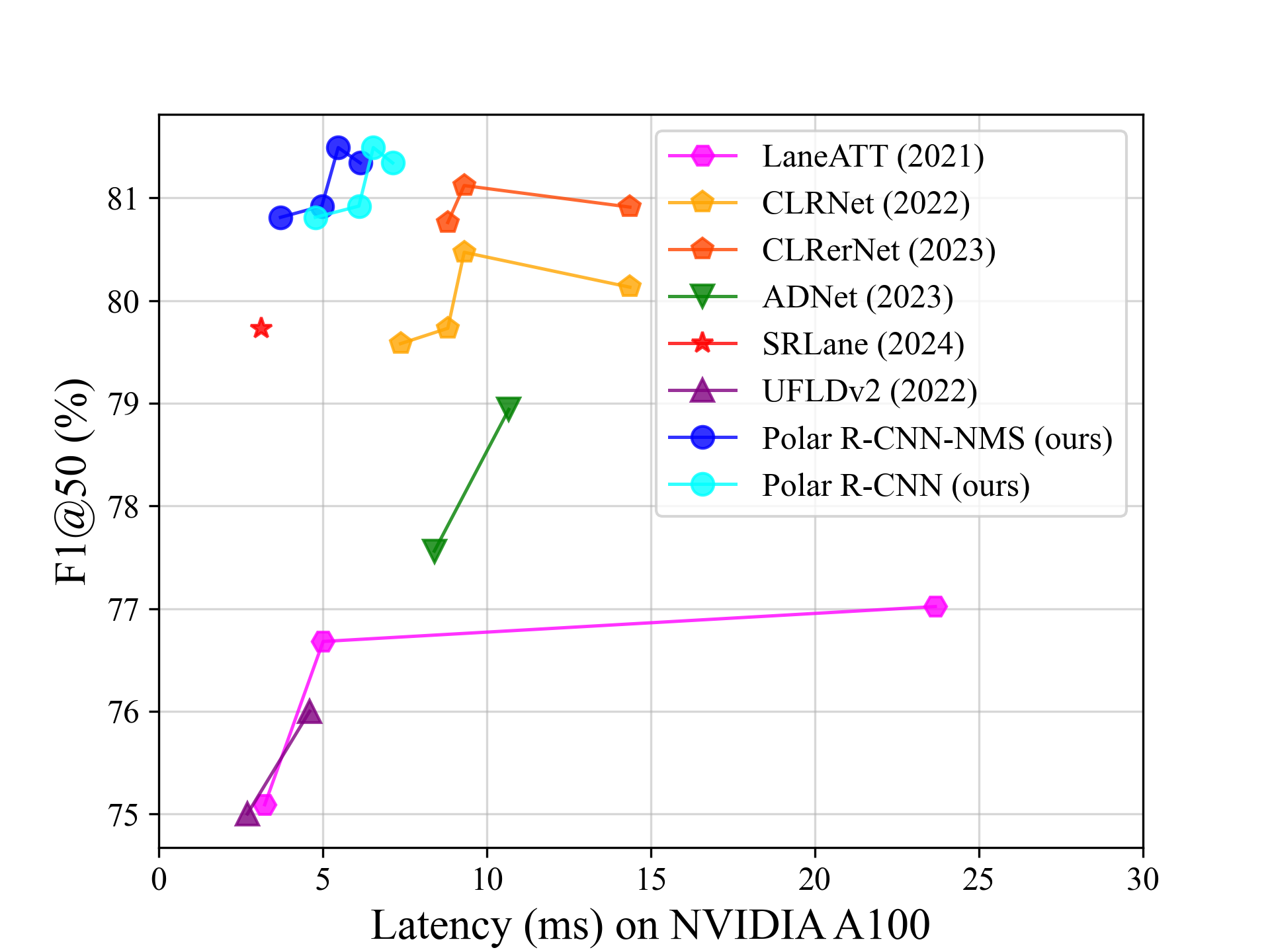}
        \caption{Latency vs. F1@50 of different methods on CULane lane detection benchmark.}
        \label{speed_method}
\end{figure}

We also compare the number of anchors and processing speed with other methods. Fig. \ref{anchor_num_method} illustrates the number of anchors used by several anchor-based methods on CULane dataset. Our proposed model utilizes the fewest proposal anchors (20 anchors) while achieving the highest F1-score on CULane. It remains competitive with state-of-the-art methods like CLRerNet, which uses 192 anchors and a cross-layer refinement. Conversely, the sparse Laneformer, which also uses 20 anchors, does not achieve optimal performance. It is important to note that our model is designed with a simpler structure without complicated components such as cross-layer refinement, indicating the pivotal role of  flexible anchors under polar coordinates in enhaning performance in sparse scenarios. Furthermore, due to its simple structure and fewer anchors, our model exhibits lower latency compared to most methods, as shown in Fig. \ref{speed_method}.

\begin{table}[t]
        \centering
        \caption{Ablation study of anchor proposal strategies}
        \begin{adjustbox}{width=\linewidth}
        \begin{tabular}{c|ccc|cc}
        \toprule
        \textbf{Anchor strategy}&\textbf{Local R}& \textbf{Local Angle}&\textbf{Auxloss}&\textbf{F1@50 (\%)}&\textbf{F1@75 (\%)}\\
        \midrule
        \multirow{2}*{Fixed}
                &-         &-         &          &79.90         &60.98\\
                &-         &-         &\checkmark&80.38         &62.35\\
        \midrule
        \multirow{5}*{Porposal}
                &          &          &          &75.85         &58.97\\
                &\checkmark&          &          &78.46         &60.32\\
                &          &\checkmark&          &80.31         &62.13\\
                &\checkmark&\checkmark&          &80.51         &63.38\\
                &\checkmark&\checkmark&\checkmark&\textbf{80.81}&\textbf{63.97}\\
        \bottomrule
        \end{tabular}
        \end{adjustbox}
        \label{aba_lph}
\end{table}

\begin{figure}[t]
        \centering
        \def\subwidth{0.24\textwidth}
        \def\imgwidth{\linewidth}
        \def\imgheight{0.4\linewidth}
        
        \begin{subfigure}{\subwidth}
                \includegraphics[width=\imgwidth, height=\imgheight]{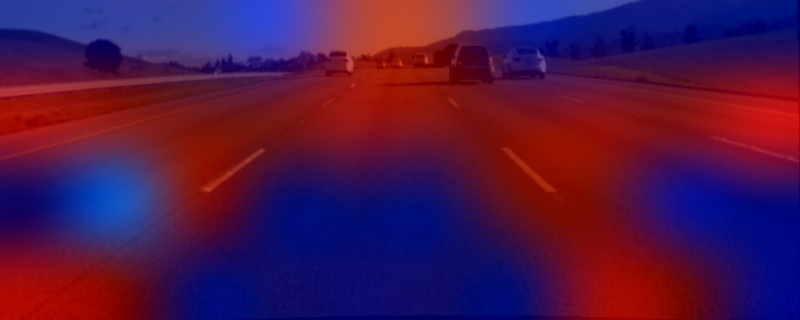}
                \caption{}
        \end{subfigure}
        \begin{subfigure}{\subwidth}
                \includegraphics[width=\imgwidth, height=\imgheight]{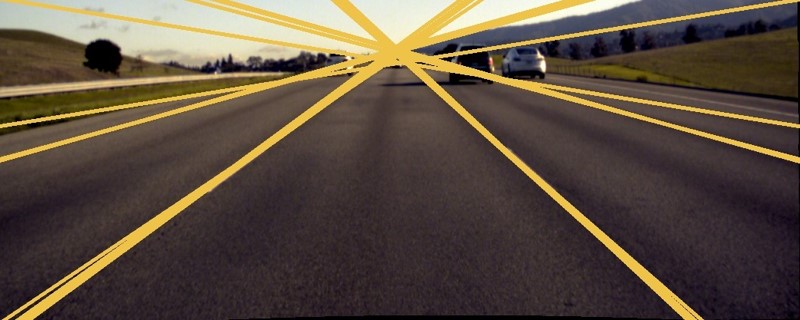}
                \caption{}
        \end{subfigure}

        \begin{subfigure}{\subwidth}
                \includegraphics[width=\imgwidth, height=\imgheight]{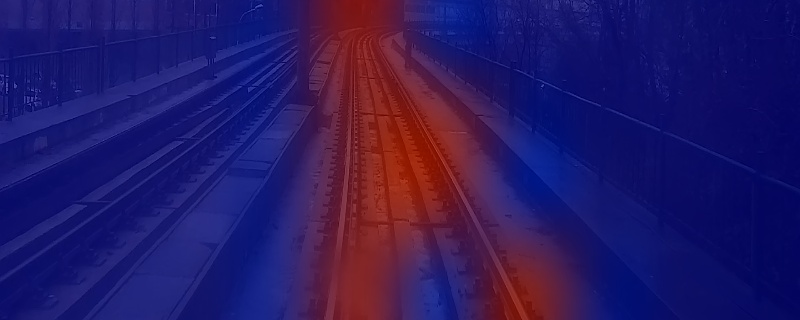}
                \caption{}
        \end{subfigure}
        \begin{subfigure}{\subwidth}
                \includegraphics[width=\imgwidth, height=\imgheight]{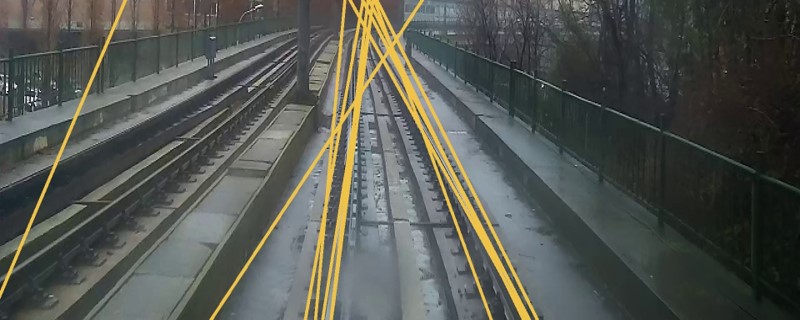}
                \caption{}
        \end{subfigure}
        \caption{(a) and (c) Heap maps of the local polar map; (b) and (d) Anchor proposals during the evaluation stage.}
        \label{cam}
\end{figure}

\begin{table}[t]
        \centering
        \caption{Ablation study on the O2O classification subhead.}
        \begin{adjustbox}{width=\linewidth}
        \begin{tabular}{cccc|ccc}
        \toprule
        \textbf{GNN}&$\boldsymbol{A}^{C}$&$\boldsymbol{A}^{G}$&\textbf{Rank Loss}&\textbf{F1@50 (\%)}&\textbf{Precision (\%)} & \textbf{Recall (\%)} \\
        \midrule
                  &          &          &          &16.19&69.05&9.17\\
        \checkmark&\checkmark&          &          &79.42&88.46&72.06\\
        \checkmark&          &\checkmark&          &71.97&73.13&70.84\\
        \checkmark&\checkmark&\checkmark&          &80.74&88.49&74.23\\
        \checkmark&\checkmark&\checkmark&\checkmark&\textbf{80.78}&\textbf{88.49}&\textbf{74.30}\\
        \bottomrule
        \end{tabular}\
        \end{adjustbox}
        \label{aba_NMSfree_block}
\end{table}

\subsection{Ablation Study}
To validate and analyze the effectiveness and influence of different component of Polar R-CNN, we conduct serveral ablation studies on CULane and CurveLanes datasets.

\textbf{Ablation study on polar coordinate system and anchor number.} To assess the importance of local polar coordinates of anchors, we examine the contribution of each component (\textit{i.e.}, angle and radius) to model performance. As shown in Table \ref{aba_lph}, both angle and radius parameters contribute to performance to varying degrees. Additionally, we conduct experiments with auxiliary loss using two anchor proposal strategies: ``fixed'' and ``proposal''. The term ``fixed'' denotes the fixed anchor configurations (192 anchors) trained by CLRNet, as illustrated in Fig. \ref{anchor setting}(b). Conversely,  ``proposals'' signifies anchors proposed by LPM (20 anchors). Model performance improves by 0.48\% and 0.3\% under the fixed anchor paradigm and proposal anchor paradigm, respectively. Moreover, the flexible anchors proposed by the LPM surpass fixed anchor configurations with auxiliary loss, utilizing fewer anchors to achieve superior performance.

\begin{figure*}[t]
        \centering
        \def\subwidth{0.325\textwidth}
        \def\imgwidth{\linewidth}
        
        \begin{subfigure}{\subwidth}
                \includegraphics[width=\imgwidth]{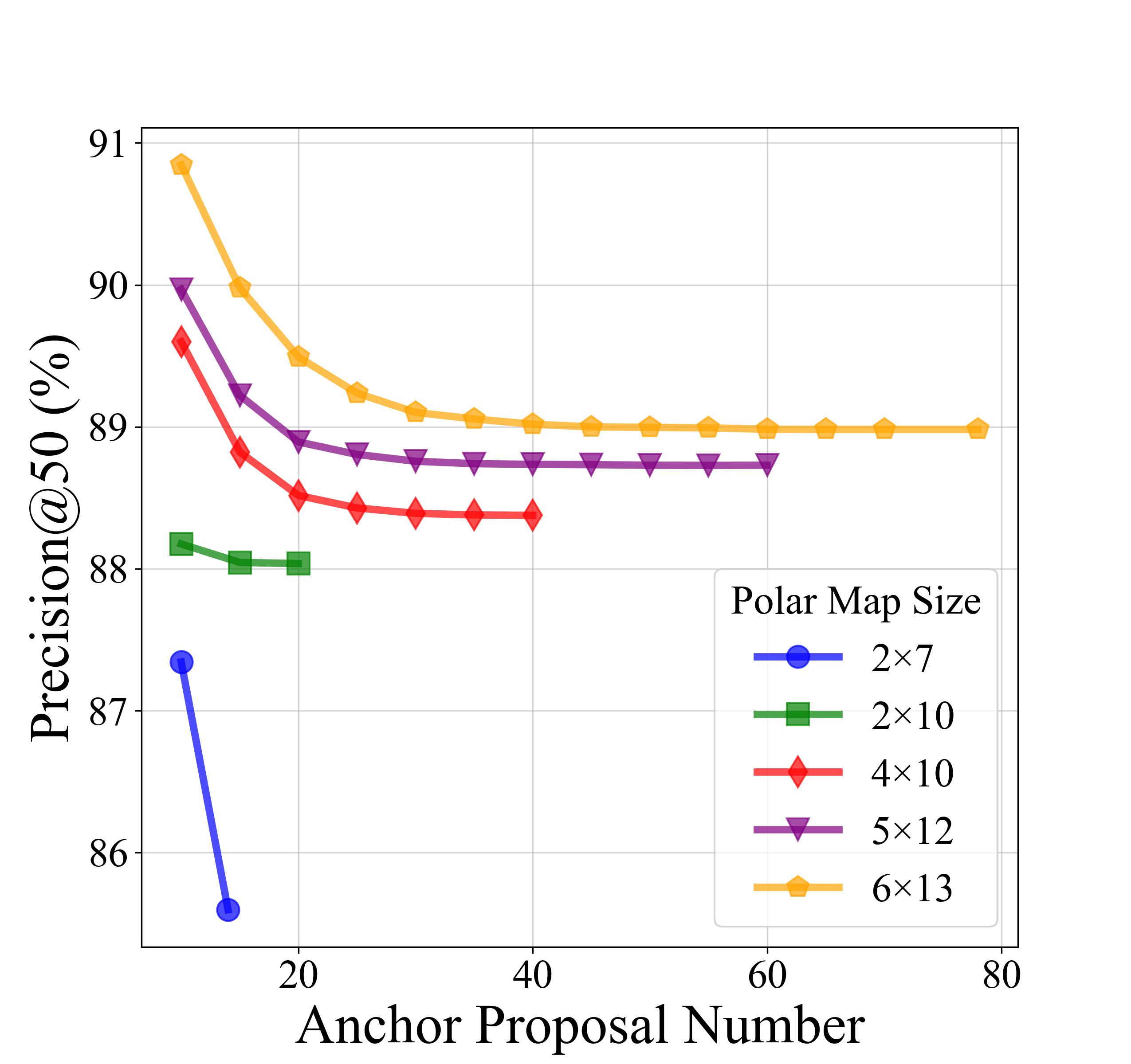}
        \end{subfigure}
        \begin{subfigure}{\subwidth}
                \includegraphics[width=\imgwidth]{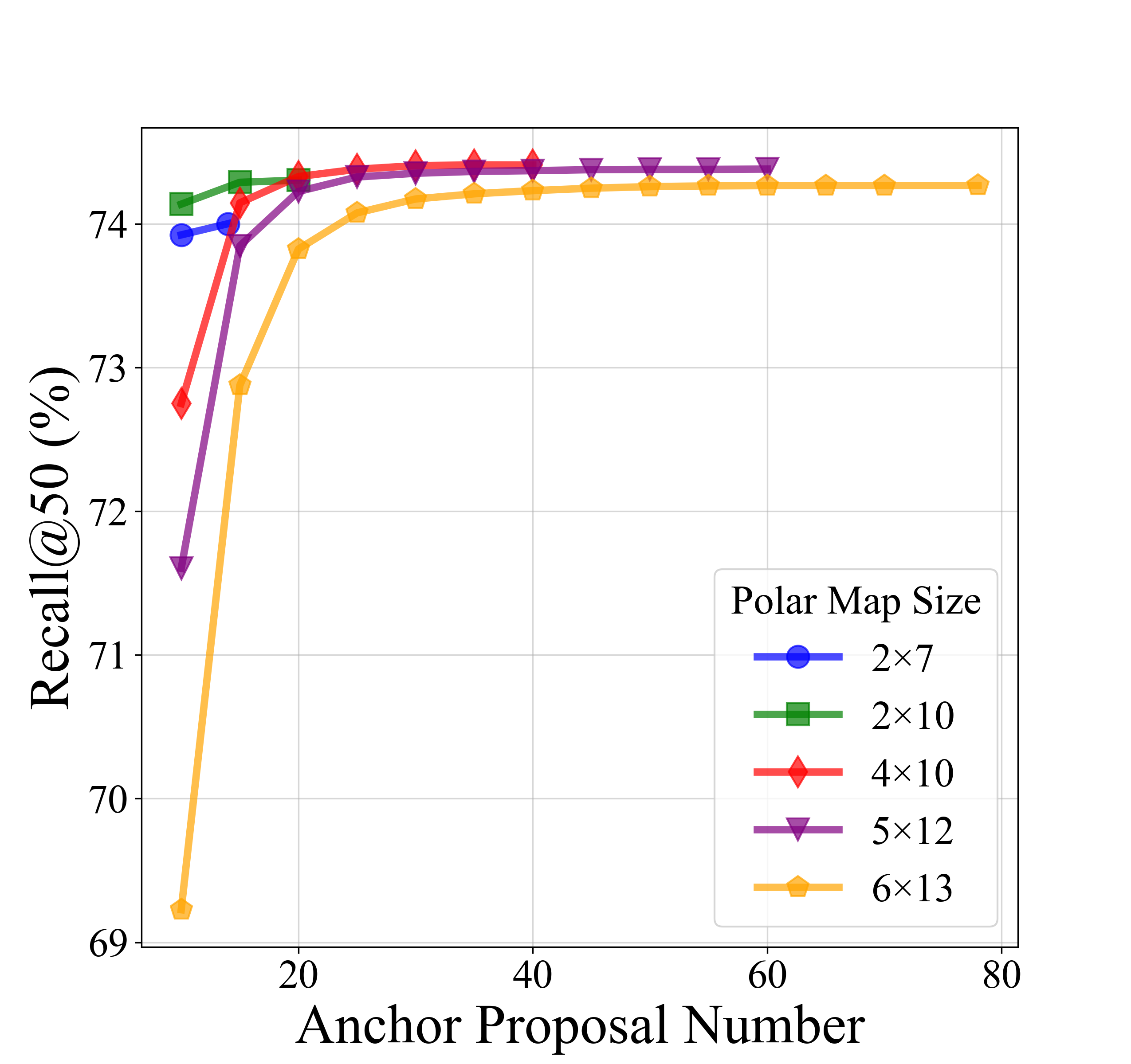}
        \end{subfigure}
        \begin{subfigure}{\subwidth}
                \includegraphics[width=\imgwidth]{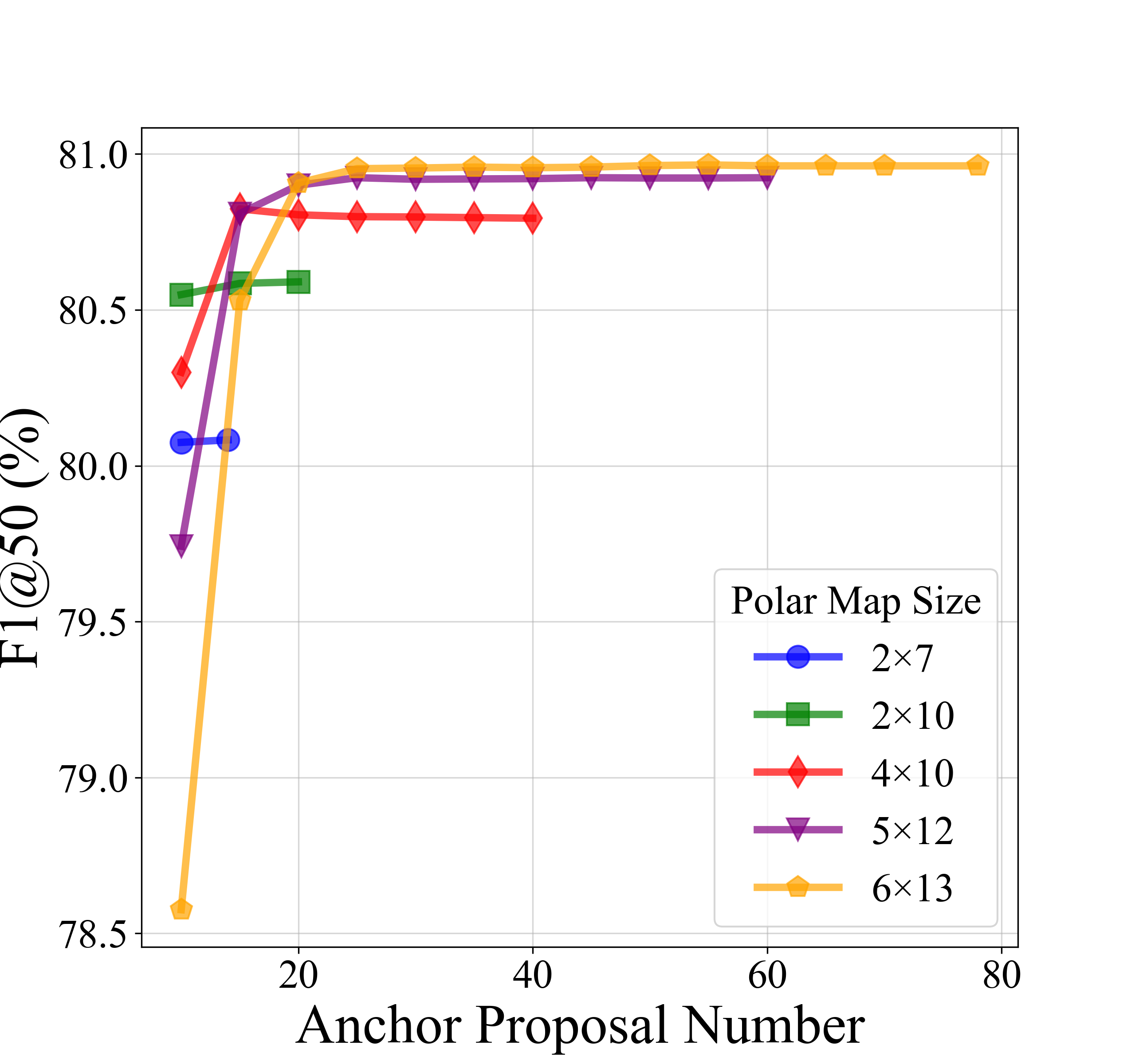}
        \end{subfigure}
        \caption{F1@50 preformance of different polar map sizes and different top-$K$ anchor selections on the CULane test set.}
        \label{anchor_num_testing}
\end{figure*}

Fig. \ref{cam} displays the heat map and top-$K$ selected anchors’ distribution in sparse scenarios. Brighter colors indicate a higher likelihood of anchors being foreground. It is evident that most of the proposed anchors are clustered around the lane ground truth. We also explore the effect of different local polar map sizes on our model, as illustrated in Fig. \ref{anchor_num_testing}. The overall F1 measure improves with the increasing in size of the local polar map and tends to stabilize when the size is sufficiently large. Specifically, precision improves, while recall decreases. A larger polar map size includes more background anchors in the second stage (since we choose $k_{dynamic}=4$ for SimOTA, with no more than 4 positive samples for each ground truth). Consequently, the model learns more negative samples, enhancing precision but reducing recall. Regarding the number of anchors chosen during the evaluation stage, recall and F1 measure show a significant improvement in the early stages of anchor number increasing but stabilize in later stages. This suggests that eliminating some anchors does not significantly affect performance. 

\begin{table}[t]
        \centering
        \caption{The ablation study for NMS and NMS-free on the CULane test set.}
        \begin{adjustbox}{width=\linewidth}
        \begin{tabular}{c|l|lll}
        \toprule
        \multicolumn{2}{c|}{\textbf{Anchor strategy~/~assign}} & \textbf{F1@50 (\%)} & \textbf{Precision (\%)} & \textbf{Recall (\%)} \\
        \midrule
        \multirow{6}*{Fixed}     
                                &O2M w/~ NMS &80.38&87.44&74.38\\
                                &O2M w/o NMS &44.03\textcolor{darkgreen}{~(36.35$\downarrow$)}&31.12\textcolor{darkgreen}{~(56.32$\downarrow$)}&75.23\textcolor{red}{~(0.85$\uparrow$)}\\
                                \cline{2-5}
                                &O2O-B w/~ NMS &78.72&87.58&71.50\\
                                &O2O-B w/o NMS &78.23\textcolor{darkgreen}{~(0.49$\downarrow$)}&86.26\textcolor{darkgreen}{~(1.32$\downarrow$)}&71.57\textcolor{red}{~(0.07$\uparrow$)}\\
                                \cline{2-5}
                                &O2O-G w/~ NMS &80.37&87.44&74.37\\
                                &O2O-G w/o NMS &80.27\textcolor{darkgreen}{~(0.10$\downarrow$)}&87.14\textcolor{darkgreen}{~(0.30$\downarrow$)}&74.40\textcolor{red}{~(0.03$\uparrow$)}\\
        \midrule
        \multirow{6}*{Proposal}     
                                &O2M w/~ NMS &80.81&88.53&74.33\\
                                &O2M w/o NMS &36.46\textcolor{darkgreen}{~(44.35$\downarrow$)}&24.09\textcolor{darkgreen}{~(64.44$\downarrow$)}&74.93\textcolor{red}{~(0.6$\uparrow$)}\\
                                \cline{2-5}
                                &O2O-B w/~ NMS &77.27&92.64&66.28\\
                                &O2O-B w/o NMS &47.11\textcolor{darkgreen}{~(30.16$\downarrow$)}&36.48\textcolor{darkgreen}{~(56.16$\downarrow$)}&66.48\textcolor{red}{~(0.20$\uparrow$)}\\
                                \cline{2-5}
                                &O2O-G w/~ NMS &80.81&88.53&74.32\\
                                &O2O-G w/o NMS &80.81\textcolor{red}{~(0.00$\uparrow$)}&88.52\textcolor{darkgreen}{~(0.01$\downarrow$)}&74.33\textcolor{red}{~(0.01$\uparrow$)}\\
        \bottomrule
        \end{tabular}
        \end{adjustbox}
        \label{NMS vs NMS-free}
\end{table}

\textbf{Ablation study on NMS-free block in sparse scenarios.} We conduct several experiments on the CULane dataset to evaluate the performance of the NMS-free paradigm in sparse scenarios. As shown in Table \ref{aba_NMSfree_block}, without using the GNN to establish relationships between anchors, Polar R-CNN fails to achieve a NMS-free paradigm, even with one-to-one assignment. Furthermore, confidence-prior adjacency matrix $\boldsymbol{A}^{C}$ proves crucial, indicating that the O2M confidence score is still essential in the NMS-free paradigm. Other components, such as the geometric-prior adjacency matrix $\boldsymbol{A}^{G}$ and rank loss, also contribute to the performance of the NMS-free block.

To compare the NMS-free paradigm with the traditional NMS paradigm, we perform experiments with the NMS-free block under both ``proposal'' and ``fixed'' anchor strategies. Table \ref{NMS vs NMS-free} presents the results of these experiments. ``O2M'' and ``O2O'' refer to the NMS (the gray dashed route in Fig. \ref{gpm}) and NMS-free paradigms (the green route in Fig. \ref{gpm}) respectively. The suffix ``-B'' signifies that the head consists solely of MLPs, whereas ``-G'' indicates that the head is equipped with the GNN block. In the fixed anchor paradigm, although the O2O classification subhead without GNN effectively eliminates redundant predictions, the performance still improved by incorporating GNN structure. In the proposal anchor paradigm, the O2O classification subhead without the GNN block fails to eliminate redundant predictions due to high anchor overlaps. In both the fixed and proposed anchor paradigms, the O2O classification subhead with the GNN block successfully eliminates redundant predictions, indicating that both label assignments and the architectural design of the head are pivotal in achieving end-to-end detection with non-redundant predictions.

We also explore the stop-gradient strategy for the O2O classification subhead. As shown in Table \ref{stop}, the gradient of the O2O classification subhead negatively impacts both the O2M classification subhead (with NMS post-processing) and the O2O classification subhead. This observation indicates that the one-to-one assignment induces significant bias into feature learning, thereby underscoring the necessity of the stop-gradient strategy to preserve optimal performance.

\begin{table}[t]
        \centering
        \caption{The ablation study for the stop gradient strategy on the CULane test set.}
        \begin{adjustbox}{width=\linewidth}
        \begin{tabular}{c|c|lll}
        \toprule
        \multicolumn{2}{c|}{\textbf{Paradigm}} & \textbf{F1@50 (\%)} & \textbf{Precision (\%)} & \textbf{Recall (\%)} \\
        \midrule
        \multirow{2}*{Baseline}  
                                 &O2M w/~ NMS &78.83&88.99&70.75\\
                                 &O2O-G w/o NMS &71.68\textcolor{darkgreen}{~(7.15$\downarrow$)}&72.56\textcolor{darkgreen}{~(16.43$\downarrow$)}&70.81\textcolor{red}{~(0.06$\uparrow$)}\\
        \midrule
        \multirow{2}*{Stop Grad} 
                                 &O2M w/~ NMS &80.81&88.53&74.33\\
                                 &O2O-G w/o NMS &80.81\textcolor{red}{~(0.00$\uparrow$)}&88.52\textcolor{darkgreen}{~(0.01$\downarrow$)}&74.33\textcolor{red}{~(0.00$\uparrow$)} \\
        \bottomrule
        \end{tabular}
        \end{adjustbox}
        \label{stop}
\end{table}

\begin{table}[t]
        \centering
        \caption{NMS vs NMS-free on CurveLanes validation set.}
        \begin{adjustbox}{width=\linewidth}
        \begin{tabular}{l|l|ccc}
        \toprule
        \textbf{Paradigm} & \textbf{NMS thres (pixel)} & \textbf{F1@50(\%)} & \textbf{Precision(\%)} & \textbf{Recall(\%)} \\
        \midrule
        \multirow{7}*{Polar R-CNN-NMS} 
                                & 50 (default) &85.38&\textbf{91.01}&80.40\\
                                & 40           &85.97&90.72&81.68\\
                                & 30           &86.26&90.44&82.45\\
                                & 25           &86.38&90.27&82.83\\
                                & 20           &86.57&90.05&83.37\\
                                & 15 (optimal) &\textbf{86.81}&89.64&84.16\\
                                & 10           &86.58&88.62&\textbf{84.64}\\
        \midrule
        Polar R-CNN & - &\textbf{87.29}&90.50&84.31\\
        \bottomrule
        \end{tabular}
        \end{adjustbox}
        \label{aba_NMS_dense}
\end{table}

\textbf{Ablation study on NMS-free block in dense scenarios.} Despite demonstrating the feasibility of replacing NMS with the O2O classification subhead in sparse scenarios, the shortcomings of NMS in dense scenarios remain. To investigate the performance of the NMS-free block in dense scenarios, we conduct experiments on the CurveLanes dataset, as detailed in Table \ref{aba_NMS_dense}.

In the traditional NMS post-processing \cite{clrernet}, the default IoU threshold is set to 50 pixels. However, this default setting may not always be optimal, especially in dense scenarios where some lane predictions might be erroneously eliminated. Lowering the IoU threshold increases recall but decreases precision. To find the most effective IoU threshold, we experimented with various values and found that a threshold of 15 pixels achieves the best trade-off, resulting in an F1-score of 86.81\%. In contrast, the NMS-free paradigm with the O2O classification subhead achieves an overall F1-score of 87.29\%, which is 0.48\% higher than the optimal threshold setting in the NMS paradigm. Additionally, both precision and recall are improved under the NMS-free approach. This indicates the O2O classification subhead with proposed GNN block is capable of learning both explicit geometric distance and implicit semantic distances between anchors, thus providing a more effective solution for dense scenarios compared to traditional NMS post-processing. More visualization outcomes can be seen in Appendix \textcolor{red}{E}.

\section{Conclusion and Future Work}
In this paper, we propose Polar R-CNN to address two key issues in anchor-based lane detection methods. By incorporating a local and global polar coordinate system, our Polar R-CNN achieves improved performance with fewer anchors. Additionally, the introduction of the O2O classification subhead with GNN block allows us to replace the traditional NMS post-processing, and the NMS-free paradigm demonstrates superior performance in dense scenarios. Our model is highly flexible and the number of anchors can be adjusted based on the specific scenario. Polar R-CNN is also deployment-friendly due to its simple structure, making it a potential new baseline for lane detection. Future work could explore new label assignment, anchor sampling strategies and complicated model structures, such as large kernels and attention mechanisms. We also plan to extend Polar R-CNN to video instance and 3D lane detection tasks, utilizing advanced geometric modeling techniques.
\bibliographystyle{IEEEtran}
\bibliography{reference} 
\clearpage
\enablecitations

\begin{appendices}
\setcounter{table}{0}   
\setcounter{figure}{0}
\setcounter{section}{0}
\setcounter{equation}{0}
\renewcommand{\thetable}{A\arabic{table}}
\renewcommand{\thefigure}{A\arabic{figure}}
\renewcommand{\thesection}{A\arabic{section}}
\renewcommand{\theequation}{A\arabic{equation}}
\section{Details about the Coordinate Systems}
In this section, we introduce the details about the coordinate systems employed in our model and coordinate transformations between them.
For convenience, we adopted Cartesian coordinate system instead of the image coordinate system, wherein the y-axis is oriented from bottom to top and the x-axis from left to right. The coordinates of of the local poles $\left\{\boldsymbol{c}^l_i\right\}$, the global pole $\boldsymbol{c}^g$, and the sampled points $\{(x_{1,j}^s,y_{1,j}^s),(x_{2,j}^s,y_{2,j}^s),\cdots,(x_{N,j}^s,y_{N,j}^s)\}_{j=1}^{K}$ of anchors are all within this coordinate by default. 

\begin{figure}[t]
        \centering
        \includegraphics[width=\linewidth]{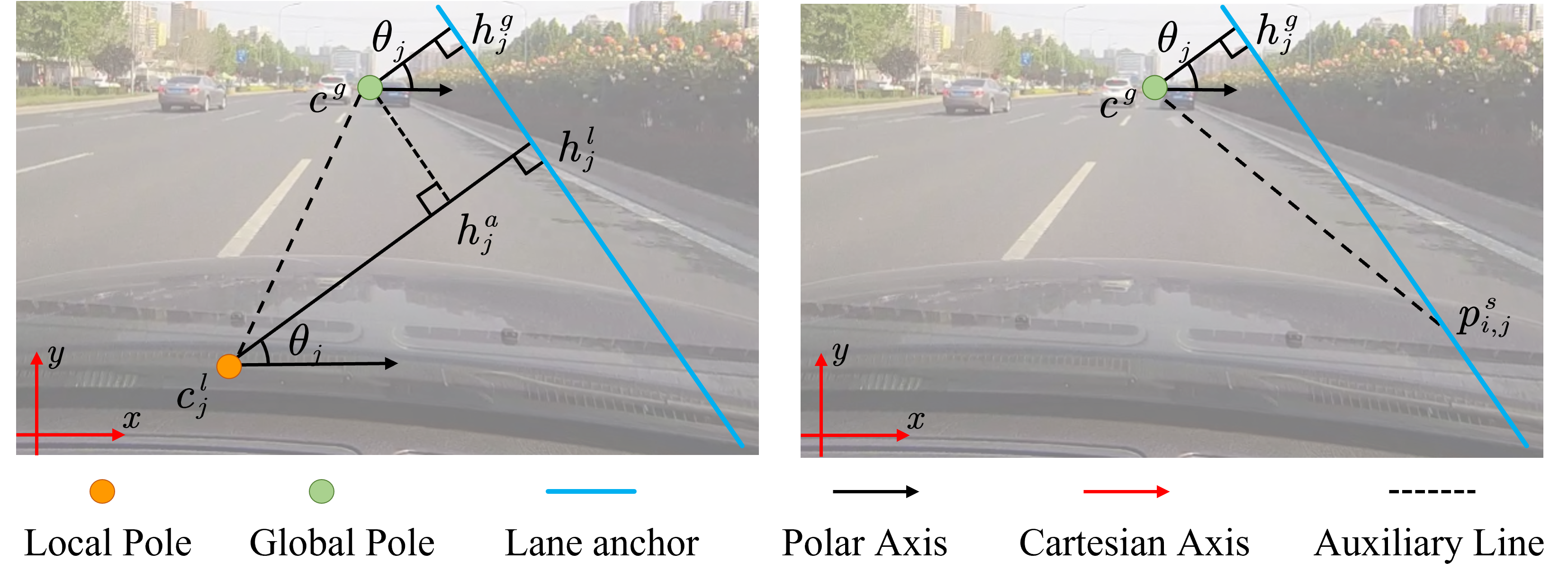}
        \caption{The symbols employed in the derivation of coordinate transformations across different coordinate systems.}
        \label{elu_proof}
\end{figure}

We now furnish the derivation of the Eq. (\ref{l2g}) and Eq. (\ref{positions}), with the crucial symbols elucidated in Fig. \ref{elu_proof}. These geometric transformations can be demonstrated with Analytic geometry theory in Euclidean space. The derivation of Eq. (\ref{l2g}) is presented as follows:
\begin{align}
        r_{j}^{g}&=\left\| \overrightarrow{c^gh_{j}^{g}} \right\| =\left\| \overrightarrow{h_{j}^{a}h_{j}^{l}} \right\| =\left\| \overrightarrow{h_{j}^{a}h_{j}^{l}} \right\| \notag\\
	&=\left\| \overrightarrow{c_{j}^{l}h_{j}^{l}}-\overrightarrow{h_{j}^{a}c_{j}^{l}} \right\| =\left\| \overrightarrow{c_{j}^{l}h_{j}^{l}} \right\| -\left\| \overrightarrow{c_{j}^{l}h_{j}^{a}} \right\| \notag\\
	&=\left\| \overrightarrow{c_{j}^{l}h_{j}^{l}} \right\| - \frac{\overrightarrow{c_{j}^{l}h_{j}^{a}}}{\left\| \overrightarrow{c_{j}^{l}h_{j}^{a}} \right\|}\cdot \overrightarrow{c_{j}^{l}h_{j}^{a}} =\left\| \overrightarrow{c_{j}^{l}h_{j}^{l}} \right\| +\frac{\overrightarrow{c_{j}^{l}h_{j}^{a}}}{\left\| \overrightarrow{c_{j}^{l}h_{j}^{a}} \right\|}\cdot \overrightarrow{c^gc_{j}^{l}} \notag\\
	&=r_{j}^{l}+\left[ \cos \theta _j;\sin \theta _j \right] ^T\left( \boldsymbol{c}_{j}^{l}-\boldsymbol{c}^g \right),
        \label{proof_l2g}
\end{align}
where $h_j^l$, $h_j^g$ and $h_j^a$ represent the foots of their respective perpendiculars in Fig. \ref{elu_proof}.
Analogously, the derivation of Eq. (\ref{positions}) is provided as follows:
\begin{align}
        &\overrightarrow{c^gp_{i,j}^{s}}\cdot \overrightarrow{c^gh_{j}^{g}}=\overrightarrow{c^gh_{j}^{g}}\cdot \overrightarrow{c^gh_{j}^{g}} \notag\\
        \Rightarrow &\overrightarrow{c^gp_{i,j}^{s}}\cdot \overrightarrow{c^gh_{j}^{g}}=\left\| \overrightarrow{c^gh_{j}^{g}} \right\| \left\| \overrightarrow{c^gh_{j}^{g}} \right\| \notag\\
        \Rightarrow &\frac{\overrightarrow{c^gh_{j}^{g}}}{\left\| \overrightarrow{c^gh_{j}^{g}} \right\|}\cdot \overrightarrow{c^gp_{i,j}^{s}}=\left\| \overrightarrow{c^gh_{j}^{g}} \right\| \notag\\
        \Rightarrow &\left[ \cos \theta _j;\sin \theta _j \right] ^T\left( \boldsymbol{p}_{i,j}^{s}-\boldsymbol{c}^g \right) =r_{j}^{g}\notag\\
        \Rightarrow &x_{i,j}^{s}\cos \theta _j+y_{i,j}^{s}\sin \theta _j=r_{j}^{g}+\left[ \cos \theta _j;\sin \theta _j \right] ^T\boldsymbol{c}^g \notag\\
        \Rightarrow &x_{i,j}^{s}=-y_{i,j}^{s}\tan \theta _j+\frac{r_{j}^{g}+\left[ \cos \theta _j;\sin \theta _j \right] ^T\boldsymbol{c}^g}{\cos \theta _j},
        \label{proof_sample}
\end{align}
where $p_{i,j}^{s}$ represents the $i$-th sampled point of the $j$-th lane anchor, whose coordinate is $\boldsymbol{p}_{i,j}^{s}\equiv(x_{i,j}^s, y_{i,j}^s)$.

\label{appendix_coord}

\section{The Design Principles of the One-to-one classification Head}
Two fundamental prerequisites of the NMS-free framework lie in the label assignment strategies and the head structures.

As for the label assignment strategy, previous work use one-to-many label assignments, which make the detection head make redundant predictions for one ground truth, resulting in the need of NMS post-processing. Thus, some works \cite{detr}\cite{learnNMS} proposed one-to-one label assignment such as Hungarian algorithm. This force the model to predict one positive sample for each lane.

However, directly using one-to-one label assignment damage the learning of the model, and structures such as MLPs and CNNs struggle to assimilate the ``one-to-one'' characteristics, resulting in the decreasing of performance compared to one-to-many label assignments with NMS post-processing\cite{yolov10}\cite{o2o}. Consider a trivial example: Let $\boldsymbol{F}^{roi}_{i}$ denotes the ROI features extracted from the $i$-th anchor, and the model is trained with one-to-one label assignment. Assuming that the $i$-th anchor and the $j$-th anchor are both close to the ground truth and overlap with each other. So the corresponding RoI features are similar, which can be expressed as follows:
\begin{align}
        \boldsymbol{F}_{i}^{roi}\approx \boldsymbol{F}_{j}^{roi}.
\end{align}
Suppose that $\boldsymbol{F}^{roi}_{i}$ is assigned as a positive sample while $\boldsymbol{F}^{roi}_{j}$ as a negative sample,  the ideal outcome should manifest as:
\begin{align}
        f_{cls}\left( \boldsymbol{F}_{i}^{roi} \right) &\rightarrow 1, \notag\\
        f_{cls}\left( \boldsymbol{F}_{j}^{roi} \right) &\rightarrow 0,
\label{sharp fun}
\end{align}
where $f_{cls}$ represents a classification head with an ordinary structure such as MLPs and CNNs. The Eq. (\ref{sharp fun}) implies that the property of $f_{cls}$ need to be ``sharp'' enough to differentiate between two similar features. In other words, the output of $f_{cls}$ changes rapidly over short periods or distances. This ``sharp'' pattern is hard to train for MLPs or CNNs solely. Consequently, additional new heuristic structures like \cite{o3d}\cite{relationnet} need to be developed.

We draw inspiration from Fast NMS \cite{yolact} for the design of the O2O classification subhead. Fast NMS serves as an iteration-free post-processing algorithm based on traditional NMS. Furthermore, we have incorporated a sort-free strategy along with geometric priors into Fast NMS, with the specifics delineated in Algorithm \ref{Graph Fast NMS}.

\begin{algorithm}[t]
        \caption{Fast NMS with Geometric Prior.}
        \begin{algorithmic}[1] 
        \REQUIRE ~~\\ 
            The index of all anchors, $1, 2, \cdots, i, \cdots, K$;\\
            The positive corresponding anchors, $\left\{ \theta _i,r_{i}^{g} \right\} |_{i=1}^{K}$;\\
            The confidence emanating from the O2M classification subhead, $s_i^g$;\\
            The regressions emanating from the O2M regression subhead, denoted as $\left\{ Lane_i \right\} |_{i=1}^{K}$\\
            The predetermined thresholds $\tau^\theta$, $\lambda^g$, $\tau_d$ and $\tau_{o2m}$.
        \ENSURE ~~\\ 
            \STATE Calculate the confidence-prior adjacency matrix $\boldsymbol{A}^{C}\in\mathbb{R}^{K\times K}$, defined as follows:
            \begin{align}
                A_{ij}^{C}=\begin{cases}
                        1,\, \mathrm{if}\,\, s_i>s_j\,\,or\,\,\left( s_i^g=s_j^g\,\,and\,\,i>j \right);\\
                        0,\, \mathrm{others}.\\
                \end{cases}
                \label{confidential matrix}
                \end{align}
            \STATE Calculate the geometric-prior adjacency matrix $\boldsymbol{A}^{G}\in\mathbb{R}^{K\times K}$, which is defined as follows:
            \begin{align}
                A_{ij}^{G}=\begin{cases}
                        1,\, \mathrm{if}\,\, \left| \theta _i-\theta _j \right|<\tau^{\theta}\,\,and\,\,\left| r_{i}^{g}-r_{j}^{g} \right|<\lambda^g;\\
                        0,\, \mathrm{others}.\\
                \end{cases}
                \label{geometric prior matrix}
        \end{align}
            \STATE Calculate the inverse distance matrix $\boldsymbol{D}  \in \mathbb{R} ^{K \times K}$ The element $D_{ij}$ in $\boldsymbol{D}$ is defined as follows: 
                \begin{align}
                        D_{ij}=d^{-1}\left( Lane_i,Lane_j \right),
                \label{al_1-3}
                \end{align}
            where $d\left(\cdot, \cdot \right)$ is some predefined function to quantify the distance between two lane predictions such as IoU.
            \STATE Define the adjacent matrix $\boldsymbol{A} = \boldsymbol{A}^{C} \odot \boldsymbol{A}^{G}$ and the final confidence $\tilde{s}_i^g$ is calculate as following:
            \begin{align}
                \tilde{s}_{i}^{g}=\begin{cases}
                        1,\, \mathrm{if}\,\, \mathrm{Max}\left(\mathcal{D}(:,j)|\boldsymbol{A}(:,j)=1\right)<\left( \tau ^d \right) ^{-1};\\
                        0,\, \mathrm{others},\\
                \end{cases}
                \label{al_1-4}
            \end{align}
                 where $j=1,2,\cdots,K$ and $\mathrm{Max}(\cdot|\boldsymbol{A}(:,j)=1)$ is a max operator along the $j$-th column of adjacency matrix $\boldsymbol{A}$ with the element $A_{:j}=1$.
            \STATE Get the final selection set:
            \begin{align}
            \varOmega_{nms}^{pos}=\left\{ i|\tilde{s}_{j}^{g}=1 \right\} \cap \left\{i|s_{i}^{g}>\tau_{o2m} \right\}.
            \label{al_1-5}
            \end{align}
        
        \RETURN The final selection result $\varOmega_{nms}^{pos}$.
        \end{algorithmic}
        \label{Graph Fast NMS}
\end{algorithm}

The new algorithm possesses a distinctly different format from its predecessor\cite{yolact}. We introduce a geometric-prior adjacency matrix characterized by $\boldsymbol{A}^G$, alleviating the suppression relationship between disparate anchors. It is manifestly to demonstrate that, when all elements within $\boldsymbol{A}^{G}$ are all set as $1$ (\textit{i.e.}, disregarding geometric priors), Algorithm \ref{Graph Fast NMS} is equivalent to Fast NMS. Building upon our newly proposed sort-free Fast NMS with geometric prior, we design the structure of the one-to-one classification head.

The principal limitations of the NMS lie in two steps, namely the definitions of distance stem from geometry (\textit{i.e.}, Eq. (\ref{al_1-3})) and the threshold employed to eliminate redundant predictions (\textit{i.e.}, Eq. (\ref{al_1-4})). For instance, in the scenarios involving double lines, despite the minimal geometric distance between the two lane instances, their semantic divergence is remarkably pronounced. Consequently, we replace the aforementioned two steps with trainable neural networks, allowing them to alleviate the limitation of Fast NMS in a data-driven fashion. The neural network blocks to replace Eq. (\ref{al_1-3}) are Eqs. (\ref{edge_layer_1})-(\ref{edge_layer_3}) in the main text.

In Eq. (\ref{edge_layer_3}), the inverse distance $\boldsymbol{D}_{ij}^{edge}\in\mathbb{R}^{d_n}$ transcends its scalar form, encapsulating the semantic distance between predictions.
We use element-wise max pooling for the tensor, as the repalcement of the max operation applied to scalar, as delineated in Eq. (\ref{maxpooling}) in the main text. Furthermore, the predetermined $\left( \tau ^d \right) ^{-1}$ is no longer utilized as the threshold of the distance. We defined a neural work as the implicit decision plane to formulate the final score $\tilde{s}_{i}^{g}$, as defined in Eq. (\ref{node_layer}), serving as the replacement of Eq. (\ref{al_1-4}).

The score $\tilde{s}_{i}^{g}$ output by the neural network transitions from a binary score to a continuous soft score ranging from 0 to 1. We introduce a new threshold $\tau_{o2o}$ within the updated criteria of Eq. (\ref{al_1-5}):
\begin{align}
        \varOmega_{nms-free}^{pos}=\left\{i|\tilde{s}_{i}^{g}>\tau_{o2o} \right\} \cap \left\{ i|s_{i}^{g}>\tau_{o2m} \right\}.
\end{align}
This criteria is also referred to as the \textit{dual confidence selection} in the main text.
\label{NMS_appendix}

\begin{table*}[htbp]
        \centering
        \caption{Infos and hyperparameters for five datasets. For the CULane dataset, $*$ denotes the actual number of training samples used to train the model. Labels for some validation/test sets are missing and different splits (\textit{i.e.}, validation and test set) are selected for different datasets.}
        \begin{adjustbox}{width=\linewidth}
        \begin{tabular}{l|l|ccccc}
        \toprule
        \multicolumn{2}{c|}{\textbf{Dataset}} & CULane & TUSimple & LLAMAS & DL-Rail & CurveLanes \\
        \midrule
        \multirow{7}*{Dataset Description}
        & Train      &88,880/$55,698^{*}$&3,268 &58,269&5,435&100,000\\
        & Validation &9,675 &358   &20,844&-    &20,000 \\
        & Test       &34,680&2,782 &20,929&1,569&-      \\
        & Resolution &$1640\times590$&$1280\times720$&$1276\times717$&$1920\times1080$&$2560\times1440$, etc\\
        & Lane &$\leqslant4$&$\leqslant5$&$\leqslant4$&$=2$&$\leqslant10$\\
        & Environment &urban and highway & highway&highway&railay&urban and highway\\
        & Distribution &sparse&sparse&sparse&sparse&sparse and dense\\
        \midrule
        \multirow{2}*{Dataset Split}
        & Evaluation &Test&Test&Test&Test&Val\\
        & Visualization &Test&Test&Val&Test&Val\\
        \midrule
        \multirow{1}*{Data Preprocess}
        & Crop Height &270&160&300&560&640, etc\\
        \midrule
        \multirow{6}*{Training Hyperparameter}
        & Epoch Number &32&70&20&90&32\\
        & Batch Size   &40&24&32&40&40\\
        & Warm up iterations &800&200&800&400&800\\
        & $w_{aux}$  &0.2&0  &0.2&0.2&0.2\\
        & $w_{rank}$ &0.7&0.7&0.1&0.7&0  \\
        \midrule
        \multirow{4}*{Evaluation Hyperparameter}
        & $H^{l}\times W^{l}$ &$4\times10$&$4\times10$&$4\times10$&$4\times10$&$6\times13$\\
        & $K$ &20&20&20&12&50\\
        & $d_n$ &5&8&10&5&5\\
        & $\tau_{o2m}$ &0.48&0.40&0.40&0.40&0.45\\
        & $\tau_{o2o}$ &0.46&0.46&0.46&0.46&0.44\\
        \bottomrule
        \end{tabular}
        \end{adjustbox}
        \label{dataset_info}
    \end{table*}

\begin{figure}[t]
        \centering
        \includegraphics[width=\linewidth]{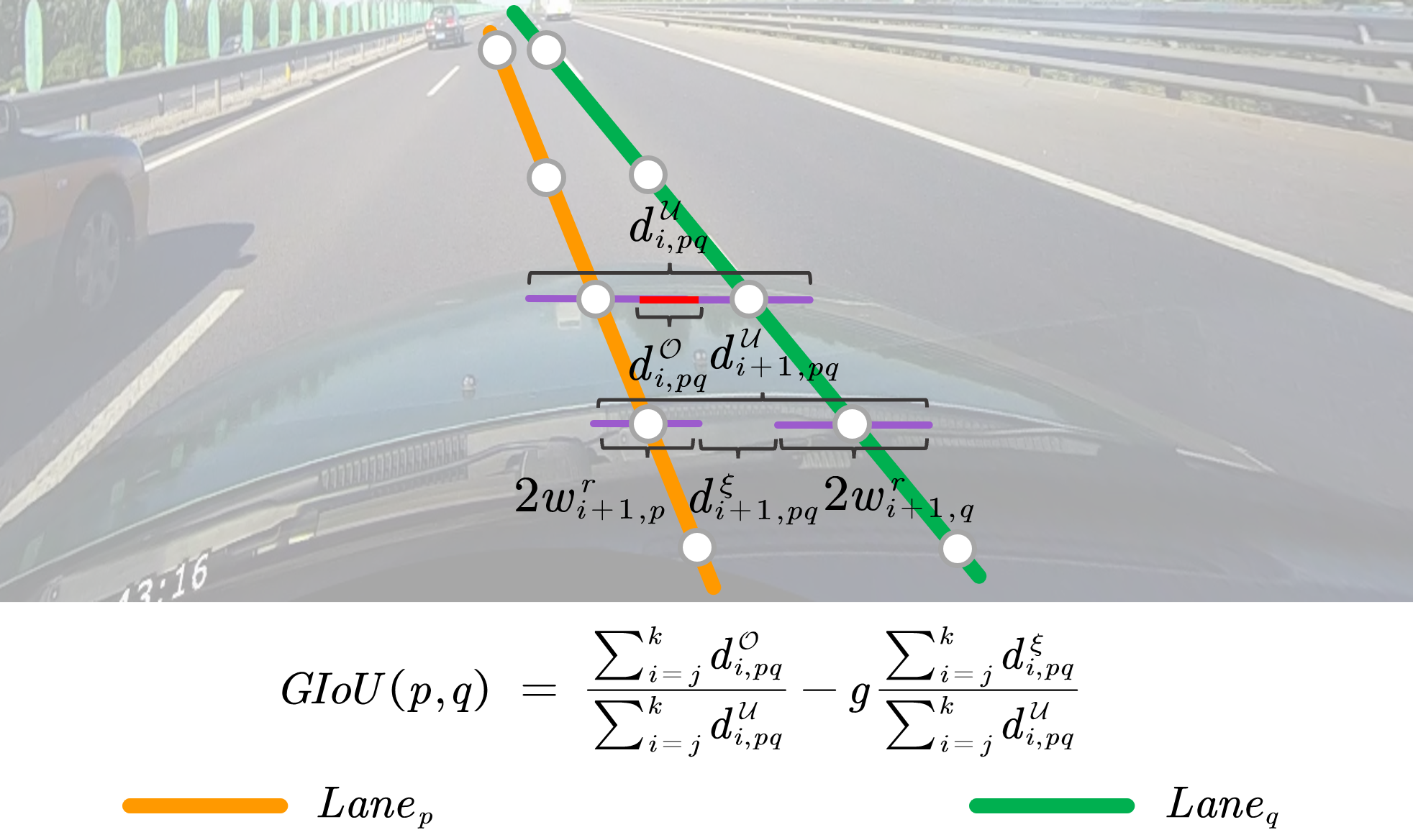} 
        \caption{Illustrations of GLaneIoU redefined in our work.}
        \label{glaneiou}
\end{figure}

\section{Details of Intersection Over Union between Lane Instances}
To ensure the IoU between lane instances aligns with the conventions of general object detection methods \cite{iouloss}\cite{giouloss}, we have redefined the IoU of lane pairs. As depicted in Fig. \ref{glaneiou}, the newly defined IoU for lanes pairs, which we refer to as GLaneIoU, is elaborated as follows:
\begin{align}
        \Delta x_{i,p}^{d}&=x_{i+1,p}^{d}-x_{i-1,p}^{d},\,\, \Delta y_{i,p}^{d}=y_{i+1,p}^{d}-y_{i-1,p}^{d}, \\
        w_{i,p}&=\frac{\sqrt{\left( \Delta x_{i,p}^{d} \right) ^2+\left( \Delta y_{i,p}^{d} \right) ^2}}{\Delta y_{i,p}^{d}}w^b,\\
        b_{i,p}^{l}&=x_{i,p}^{d}-w_{i,p},\,\, b_{i,p}^{r}=x_{i,p}^{d}+w_{i,p},
\end{align}
where $w^{b}$ is the base semi-width parameter and $w_{i,p}$ is the actual semi-width of $p$-th lane instance. The sets $\left\{ b_{i,p}^{l} \right\} _{i=1}^{N}$ and $\left\{ b_{i,p}^{r} \right\} _{i=1}^{N}$ signify the left and right boundaries of the $p$-th lane instance. Subsequently, we defined inter and union between lane instances:
\begin{align}
        d_{i,pq}^{\mathcal{O}}&=\max \left( \min \left( b_{i,p}^{r}, b_{i,q}^{r} \right) -\max \left( b_{i,p}^{l}, b_{i,q}^{l} \right) , 0 \right),\\
        d_{i,pq}^{\xi}&=\max \left( \max \left( b_{i,p}^{l}, b_{i,q}^{l} \right) -\min \left( b_{i,p}^{r}, b_{i,q}^{r} \right) , 0 \right),\\
        d_{i,pq}^{\mathcal{U}}&=\max \left( b_{i,p}^{r}, b_{i,q}^{r} \right) -\min \left( b_{i,p}^{l}, b_{i,q}^{l} \right).
\end{align}
 The defination of $\left\{d_{i,pq}^{\mathcal{O}}\right\}_{i=1}^{N}$, $\left\{d_{i,pq}^{\xi}\right\}_{i=1}^{N}$ and $\left\{d_{i,pq}^{\mathcal{U}}\right\}_{i=1}^{N}$ denote the over distance, gap distance, and union distance, respectively. These definitions closely resemble but slightly differ from those in \cite{clrnet} and \cite{adnet}, modifications to ensure non-negative values. This formulation aims to maintain consistency with the IoU definitions used for bounding boxes. Thus, the overall GLaneIoU between the $p$-th and $q$-th lane instances is expressed as:
\begin{align}
        GIoU\left( p,q \right)=\frac{\sum\nolimits_{i=j}^k{d_{i,pq}^{\mathcal{O}}}}{\sum\nolimits_{i=j}^k{d_{i,pq}^{\mathcal{U}}}}-g\frac{\sum\nolimits_{i=j}^k{d_{i,pq}^{\xi}}}{\sum\nolimits_{i=j}^k{d_{i,pq}^{\mathcal{U}}}},
\end{align}
where j and k are the indices of the start point and the end point, respectively. It's evident that when $g=0$, the $GIoU$ for lane pairs corresponds to that for bounding box, with a value range of $\left[0, 1 \right]$. When $g=1$, the $GIoU$ for lane pairs corresponds to that for bounding box, with a value range of $\left(-1, 1 \right]$. 

\label{giou_appendix}

\section{Details about The Label assignment and Loss function.}

\begin{figure}[t]
        \centering
        \includegraphics[width=\linewidth]{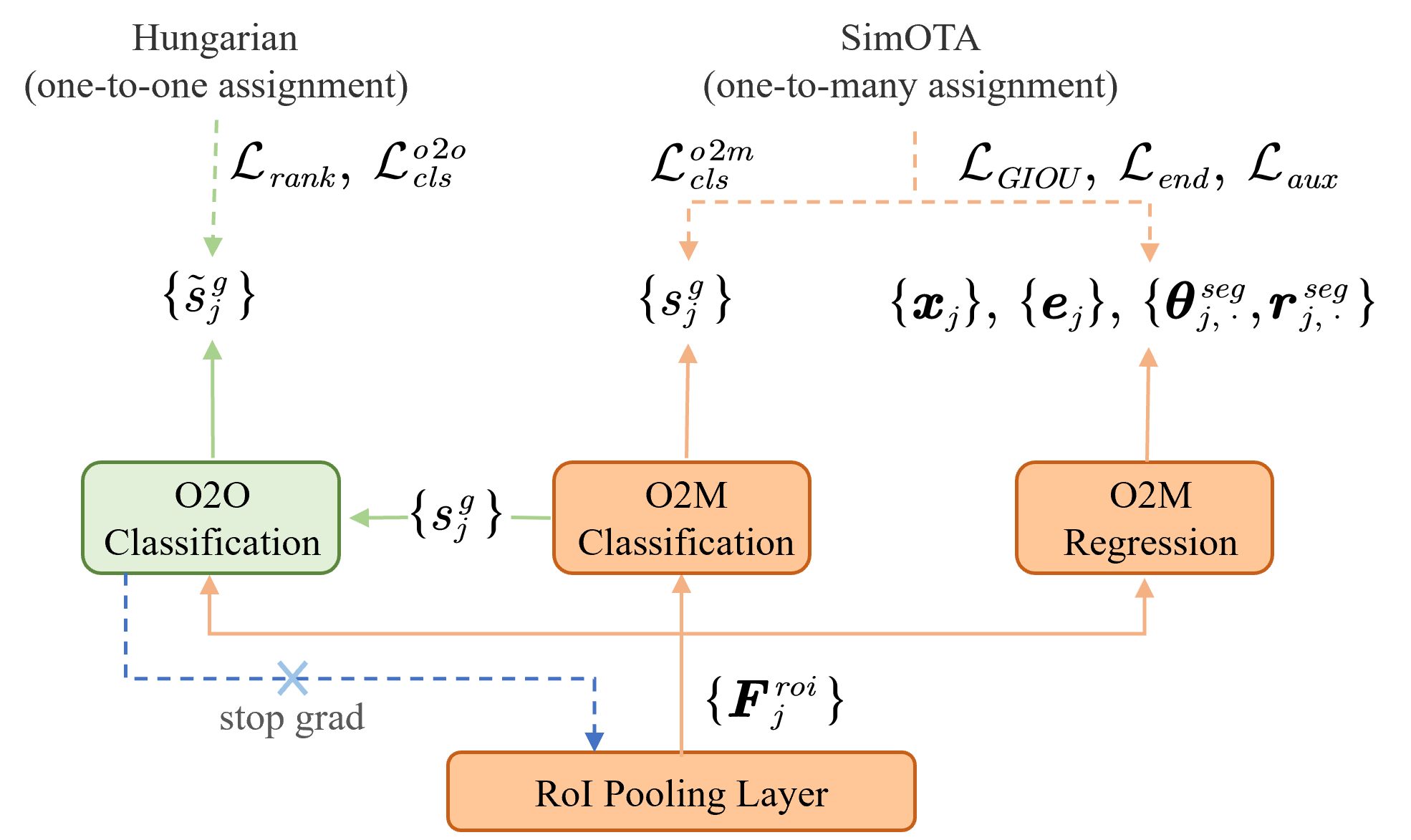}
        \caption{Label assignment and loss function for the triplet head.}
        \label{head_assign}
\end{figure}
Details about cost function and label assignments for the triplet head are furnished here. A dual label assignment strategy \cite{date} is employed for the triplet head, as illustrated in Fig. \ref{head_assign}. Specifically, we implement one-to-many label assignments for both the O2O classification subhead and the O2M regression subhead. This section closely aligns with previous work \cite{clrernet}. To endow our model with NMS-free paradigm, we additionally incorporate the O2O classification subhead and apply a one-to-one label assignment to it. 

The cost metrics for both one-to-one and one-to-many label assignments are articulated as follows:
\begin{align}
        \mathcal{C} _{p,q}^{o2o}=\tilde{s}_{p}^{g}\times \left( GIoU\left( p,q \right) \right) ^{\beta} \label{o2o_cost},\\
        \mathcal{C} _{p,q}^{o2m}=s_{p}^{g}\times \left( GIoU\left( p,q \right) \right) ^{\beta}, \label{o2m_cost}
\end{align}
where $\mathcal{C} _{pq}^{o2o}$ and $\mathcal{C} _{pq}^{o2m}$ denote the cost metric between $p$-th prediction and $q$-th ground truth and $g$ in $GIoU$ are set to $0$ to ensure it maintains non-negative. These metrics imply that both the confidence score and geometric distance contribute to the cost metrics.

Suppose that there exist $K$ predictions and $G$ ground truth. Let $\pi$ denotes the one-to-one label assignment strategy and $\pi(q)$ represent that the $\pi(q)$-th prediction is assigned to the $q$-th anchor. Additionally, $\mathscr{S}_{K, G}$ denotes the set of all possible one-to-one assignment strategies for K predictions and G ground truth. It's straightforward to demonstrate that the total number of one-to-one assignment strategies $\left| \mathscr{S} _{K,G} \right|$ is $\frac{K!}{\left( K-G \right)!}$. The final optimal assignment $\hat{\pi}$ is determined as follows:
\begin{align}
        \hat{\pi}=\underset{\pi \in \mathscr{S}_{K,G}}{arg\max}\sum_{q=1}^G{\mathcal{C} _{\pi \left( q \right) ,q}^{o2o}}.
\end{align}
This assignment problem can be solved by Hungarian algorithm \cite{detr}. Finally, $G$ predictions are assigned as positive samples and $K-G$ predictions are assigned as negative samples.
\begin{figure*}[ht]
        \centering
        \def\pagewidth{0.49\textwidth}
        \def\subwidth{0.47\linewidth}
        \def\imgwidth{\linewidth}
        \def\imgheight{0.5625\linewidth}
        \def\dashheight{0.8\linewidth}

        \begin{subfigure}{\pagewidth}
            \rotatebox{90}{\small{GT}}
            \begin{minipage}{\subwidth}
                \includegraphics[width=\imgwidth, height=\imgheight]{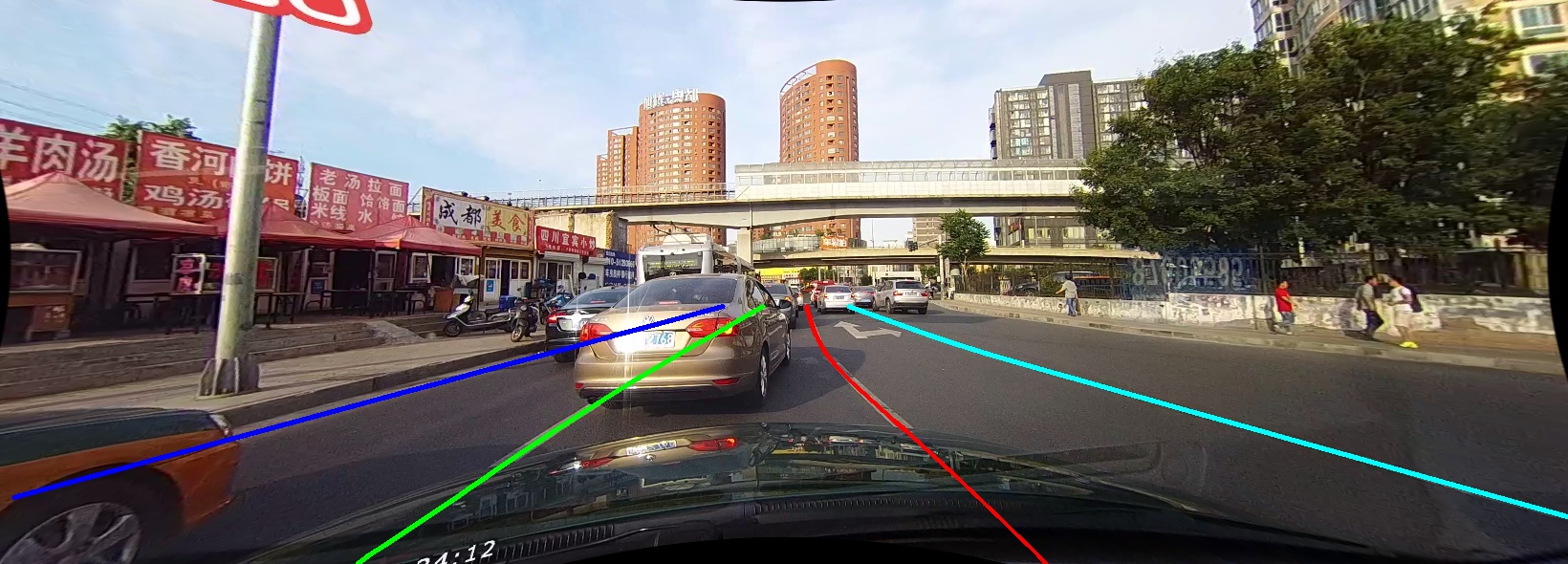}
            \end{minipage}
            \begin{minipage}{\subwidth}
                \includegraphics[width=\imgwidth, height=\imgheight]{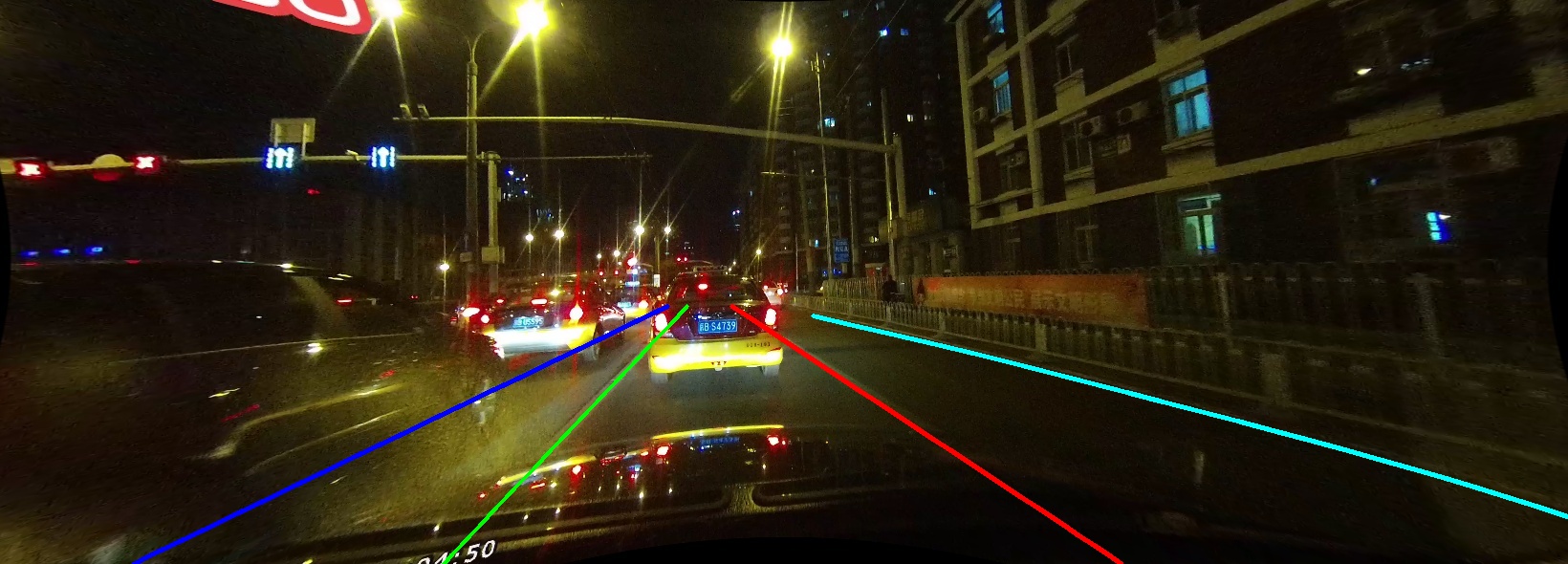}
            \end{minipage}
        \end{subfigure}
        \begin{subfigure}{\pagewidth}
            \begin{minipage}{\subwidth}
                \includegraphics[width=\imgwidth, height=\imgheight]{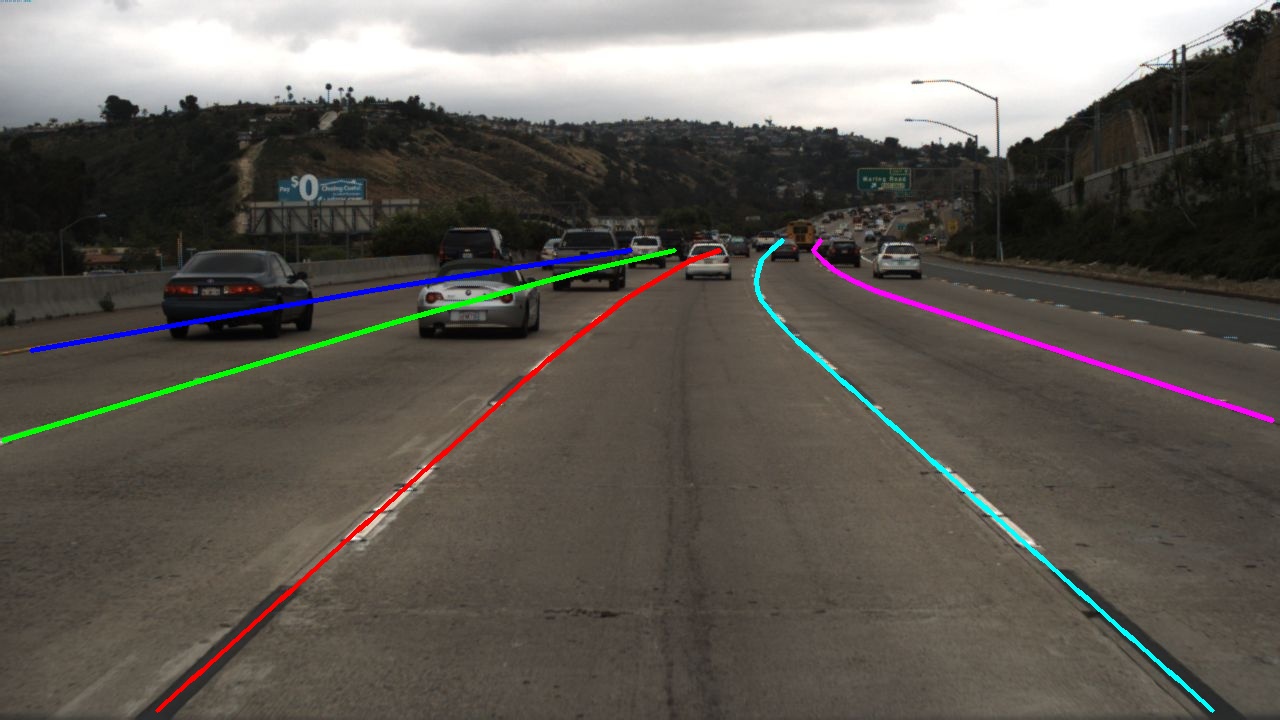}
            \end{minipage}
            \begin{minipage}{\subwidth}
                \includegraphics[width=\imgwidth, height=\imgheight]{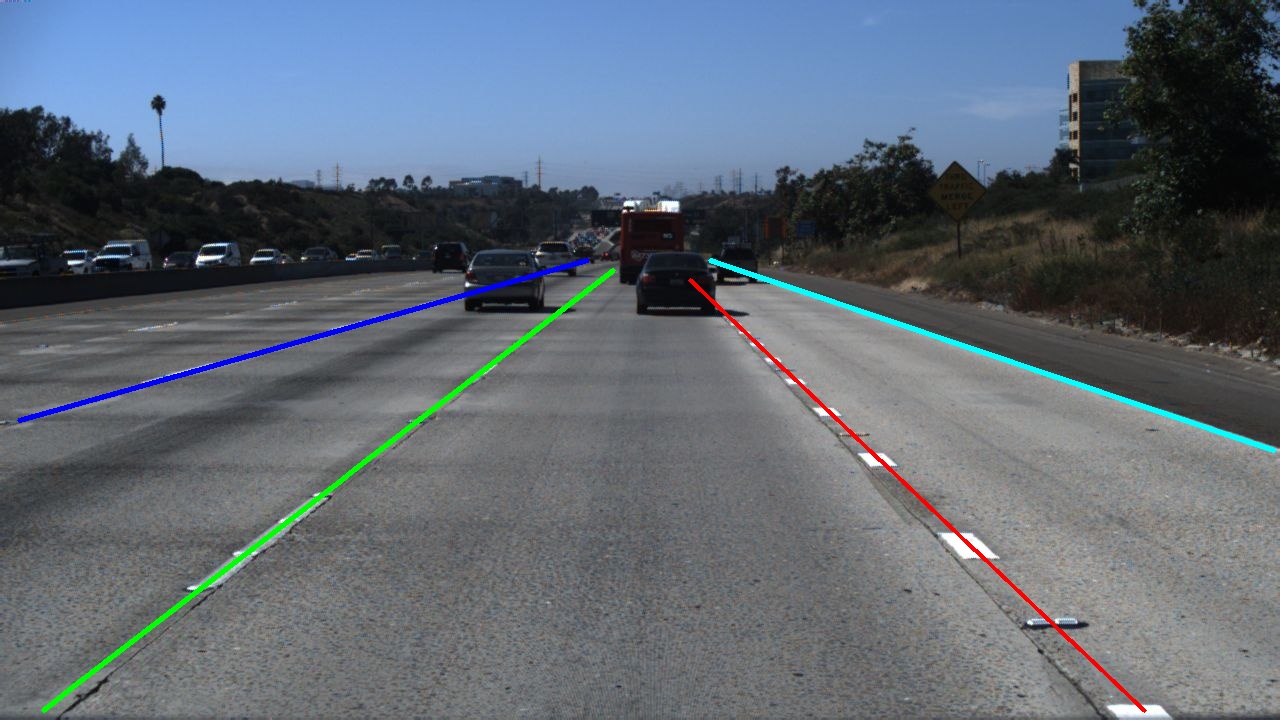}
            \end{minipage}
        \end{subfigure}
        \vspace{0.5em}

        \begin{subfigure}{\pagewidth}
            \raisebox{-1.5em}{\rotatebox{90}{\small{Anchors}}}
            \begin{minipage}{\subwidth}
                \includegraphics[width=\imgwidth, height=\imgheight]{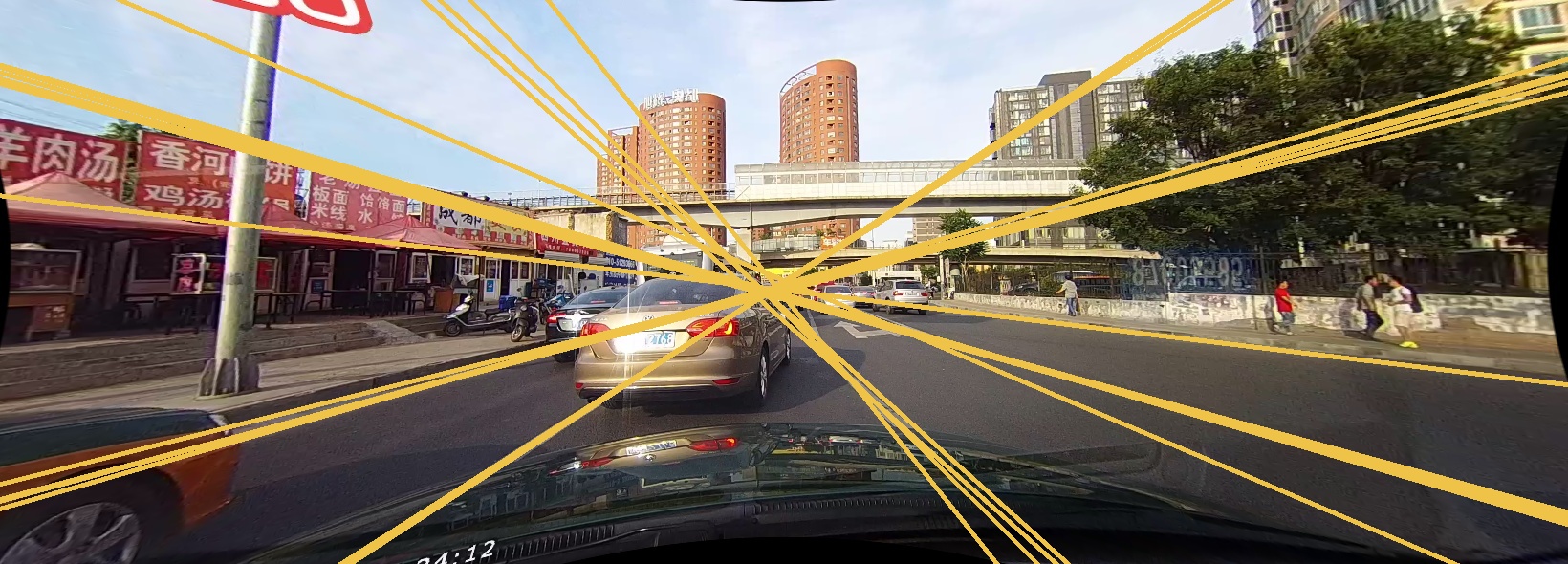}
            \end{minipage}
            \begin{minipage}{\subwidth}
                \includegraphics[width=\imgwidth, height=\imgheight]{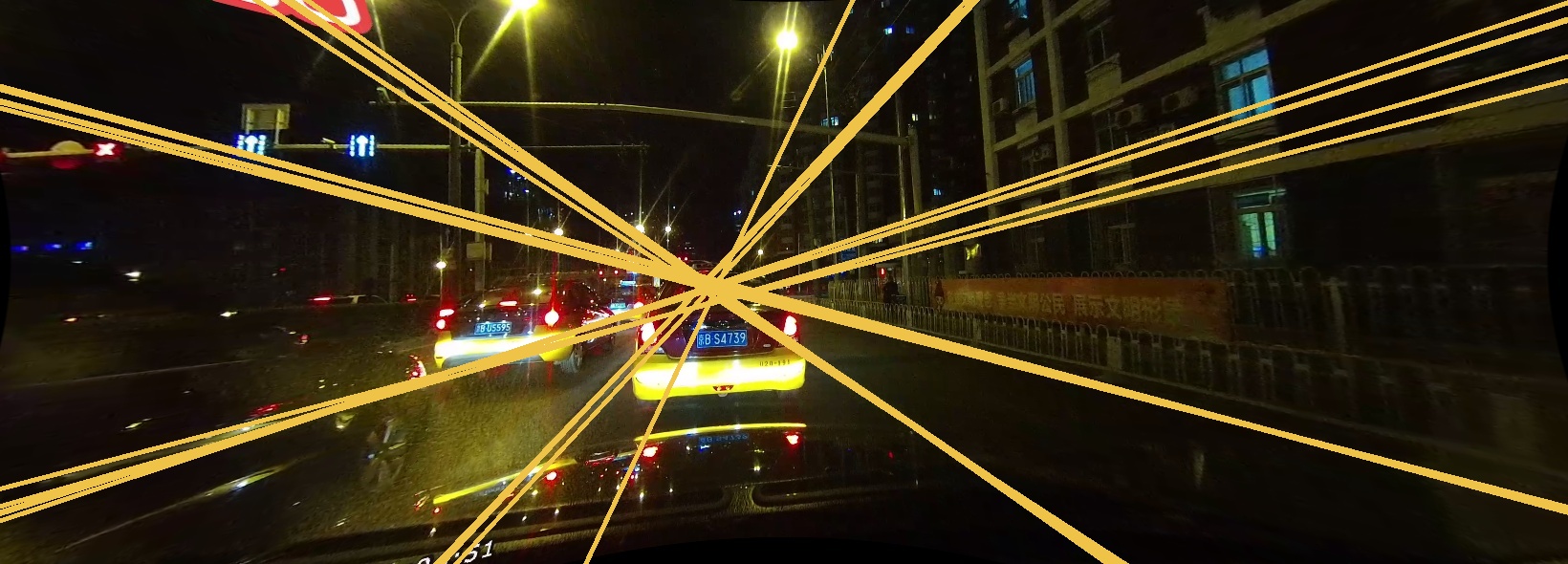}
            \end{minipage}
        \end{subfigure}
        \begin{subfigure}{\pagewidth}
            \begin{minipage}{\subwidth}
                \includegraphics[width=\imgwidth, height=\imgheight]{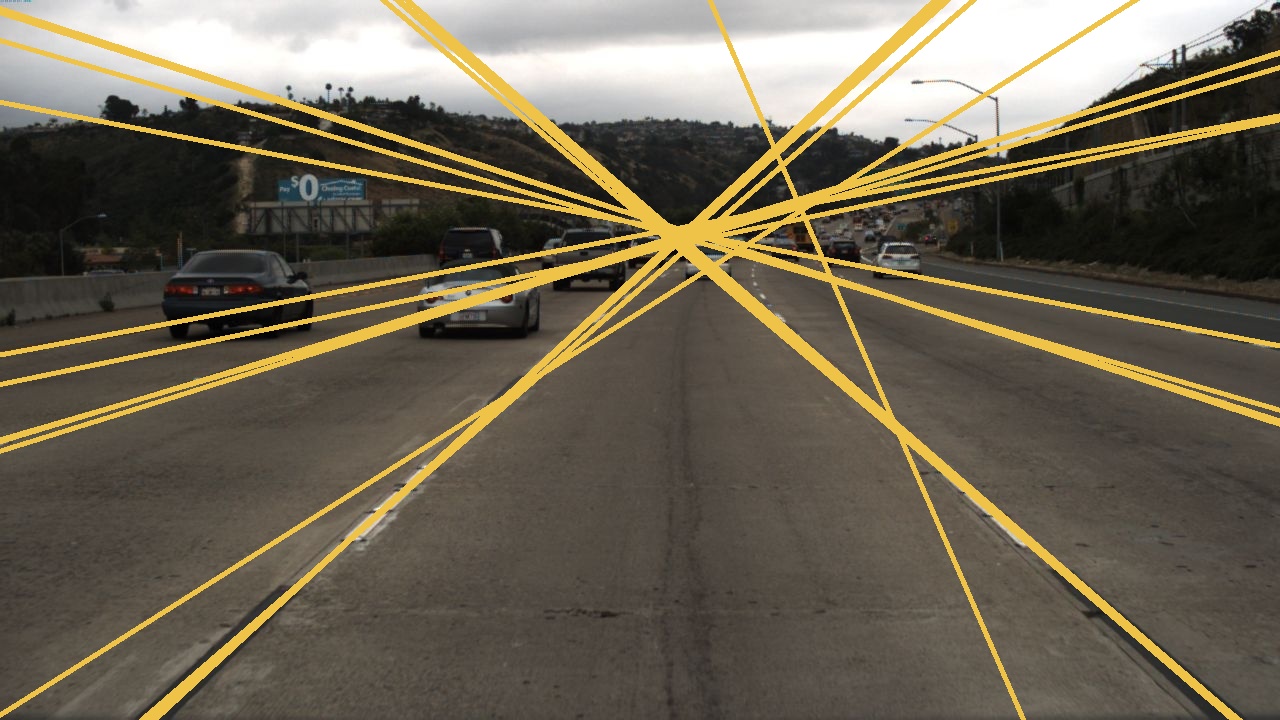}
            \end{minipage}
            \begin{minipage}{\subwidth}
                \includegraphics[width=\imgwidth, height=\imgheight]{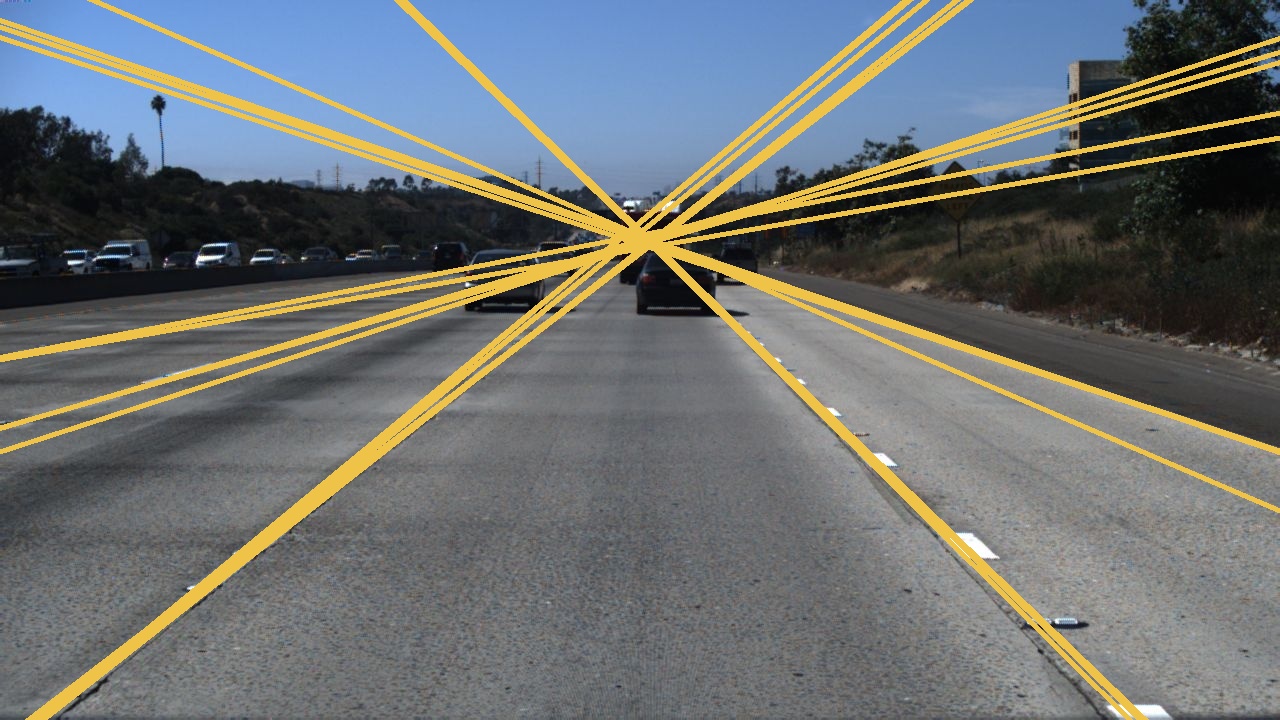}
            \end{minipage}
        \end{subfigure}
        \vspace{0.5em}
        
        \begin{subfigure}{\pagewidth}
            \raisebox{-2em}{\rotatebox{90}{\small{Predictions}}}
            \begin{minipage}{\subwidth}
                \includegraphics[width=\imgwidth, height=\imgheight]{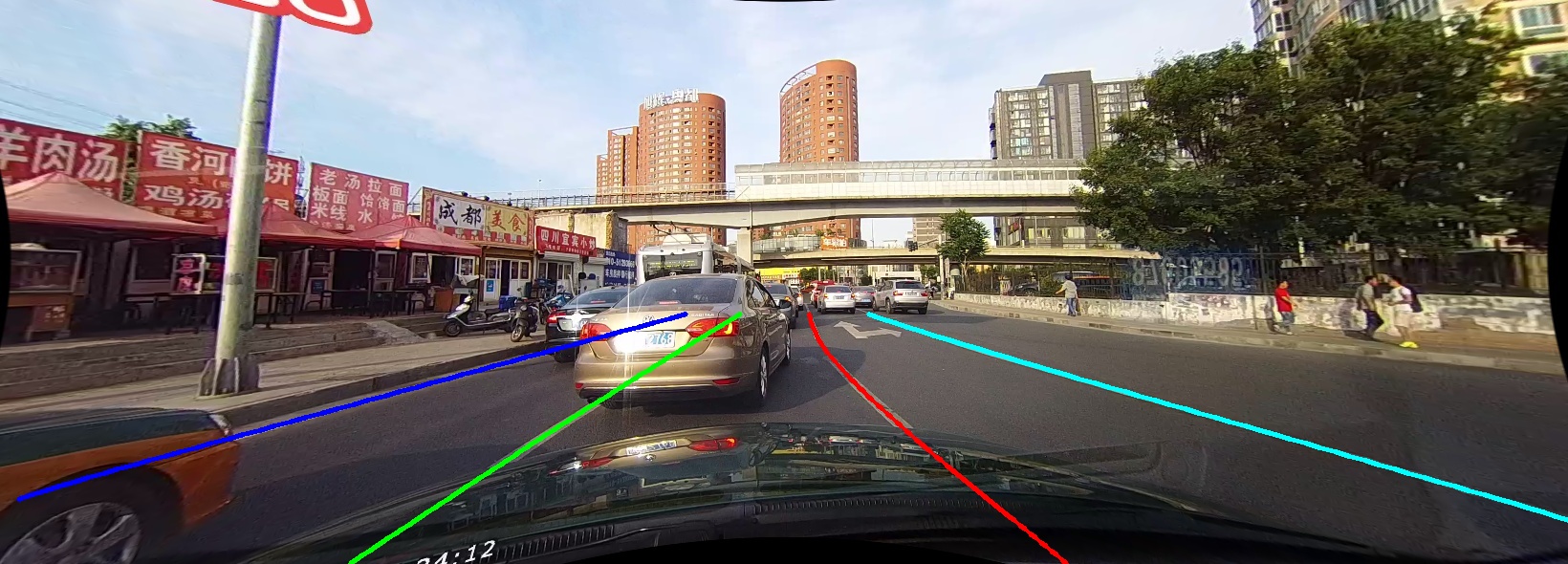}
            \end{minipage}
            \begin{minipage}{\subwidth}
                \includegraphics[width=\imgwidth, height=\imgheight]{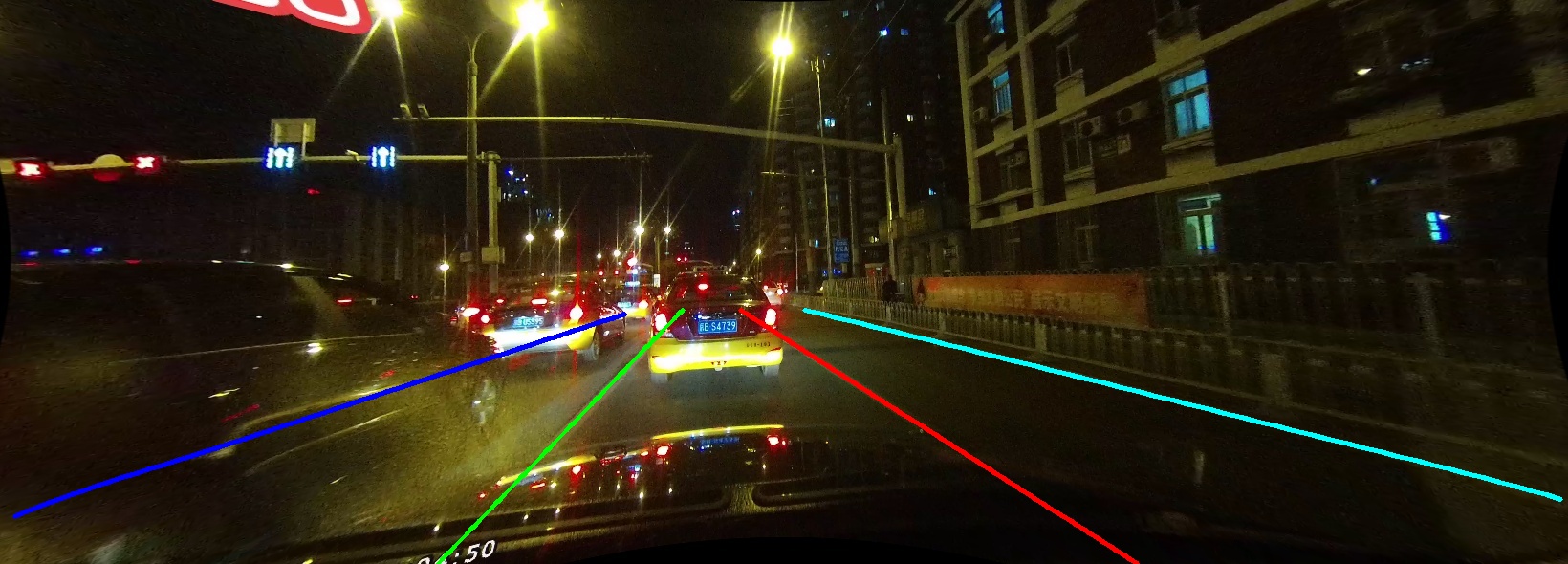}
            \end{minipage}
            \caption{CULane}
        \end{subfigure}
        \begin{subfigure}{\pagewidth}
            \begin{minipage}{\subwidth}
                \includegraphics[width=\imgwidth, height=\imgheight]{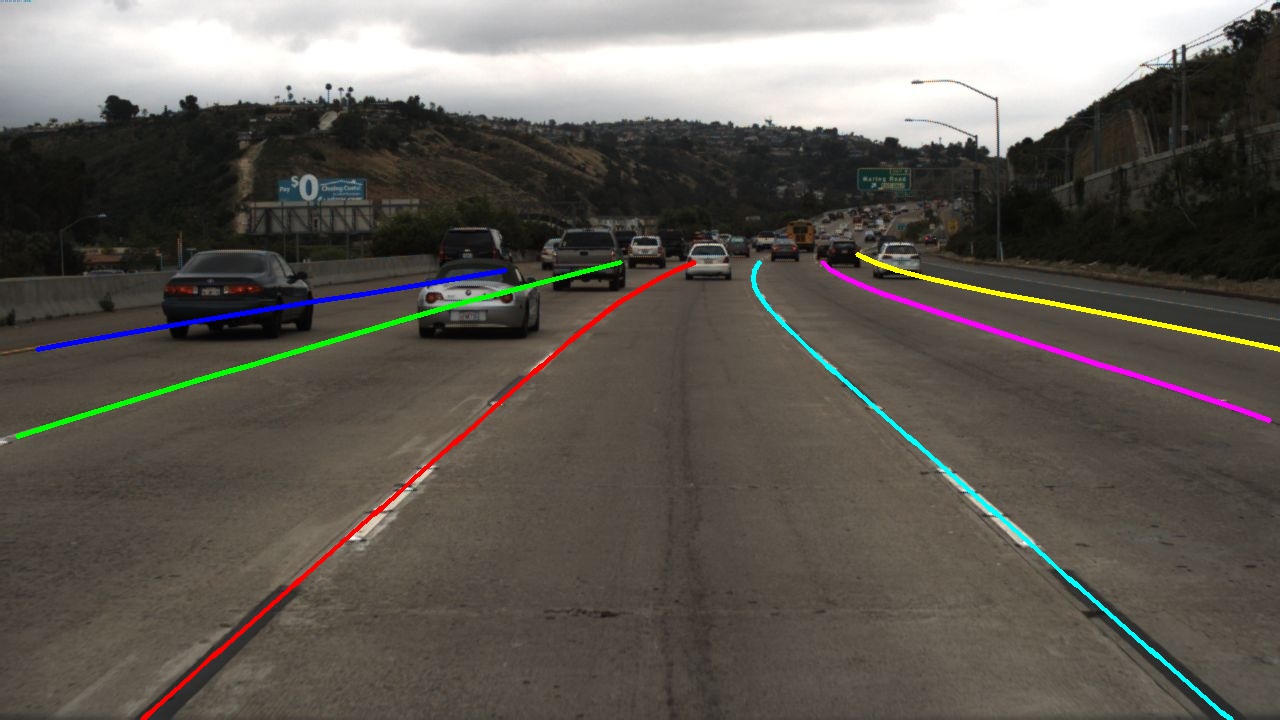}
            \end{minipage}
            \begin{minipage}{\subwidth}
                \includegraphics[width=\imgwidth, height=\imgheight]{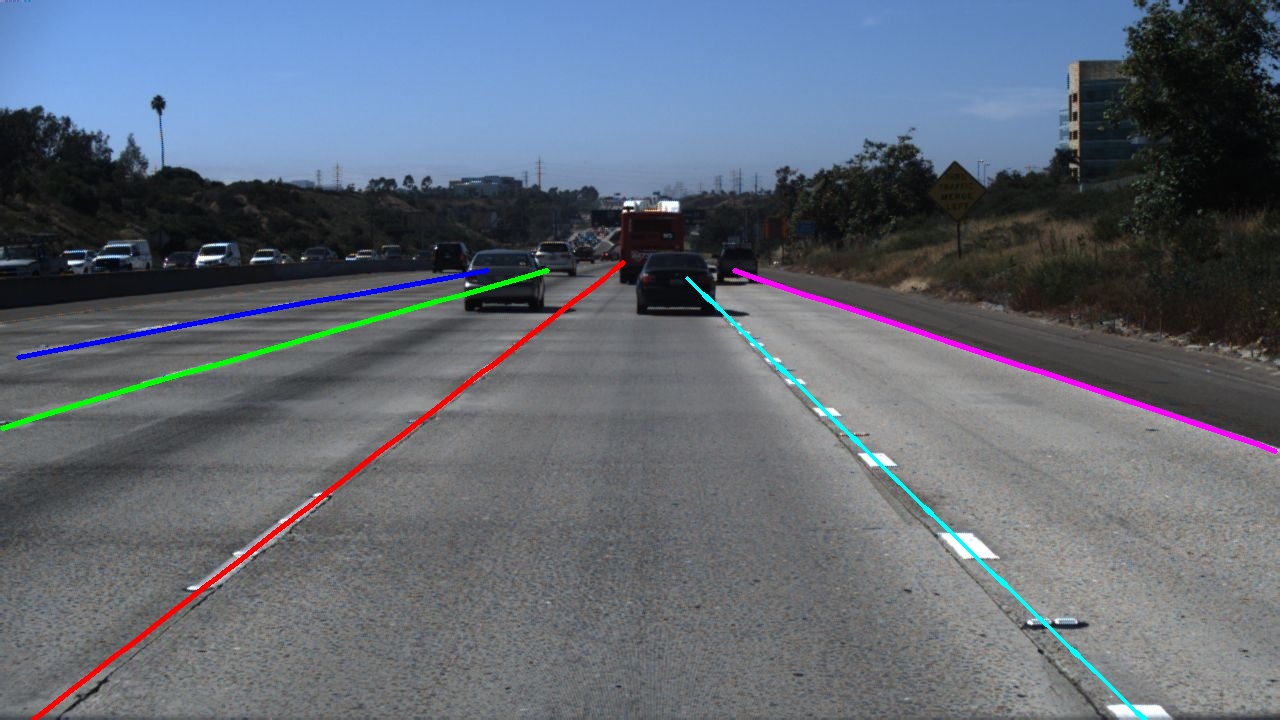}
            \end{minipage}
            \caption{TuSimple}
        \end{subfigure}
        \vspace{0.5em}

        
        \begin{subfigure}{\pagewidth}
            \rotatebox{90}{\small{GT}}
            \begin{minipage}{\subwidth}
                    \includegraphics[width=\imgwidth, height=\imgheight]{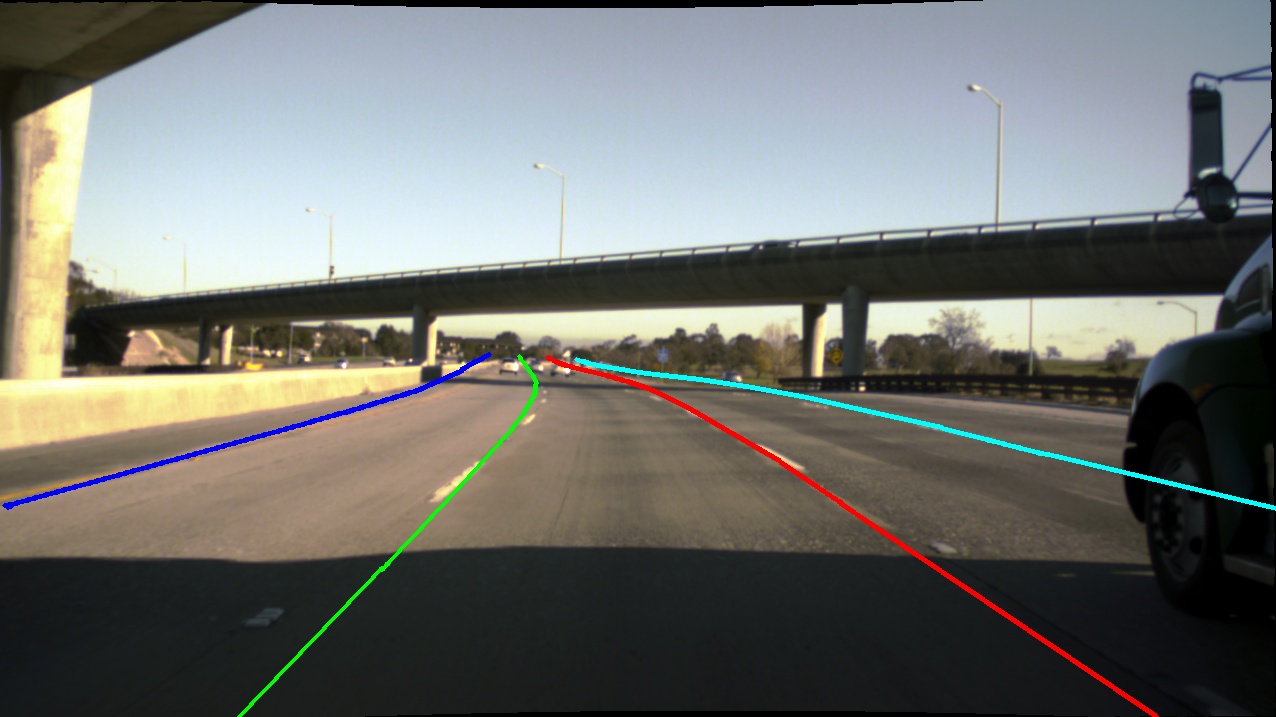}
            \end{minipage}
            \begin{minipage}{\subwidth}
                    \includegraphics[width=\imgwidth, height=\imgheight]{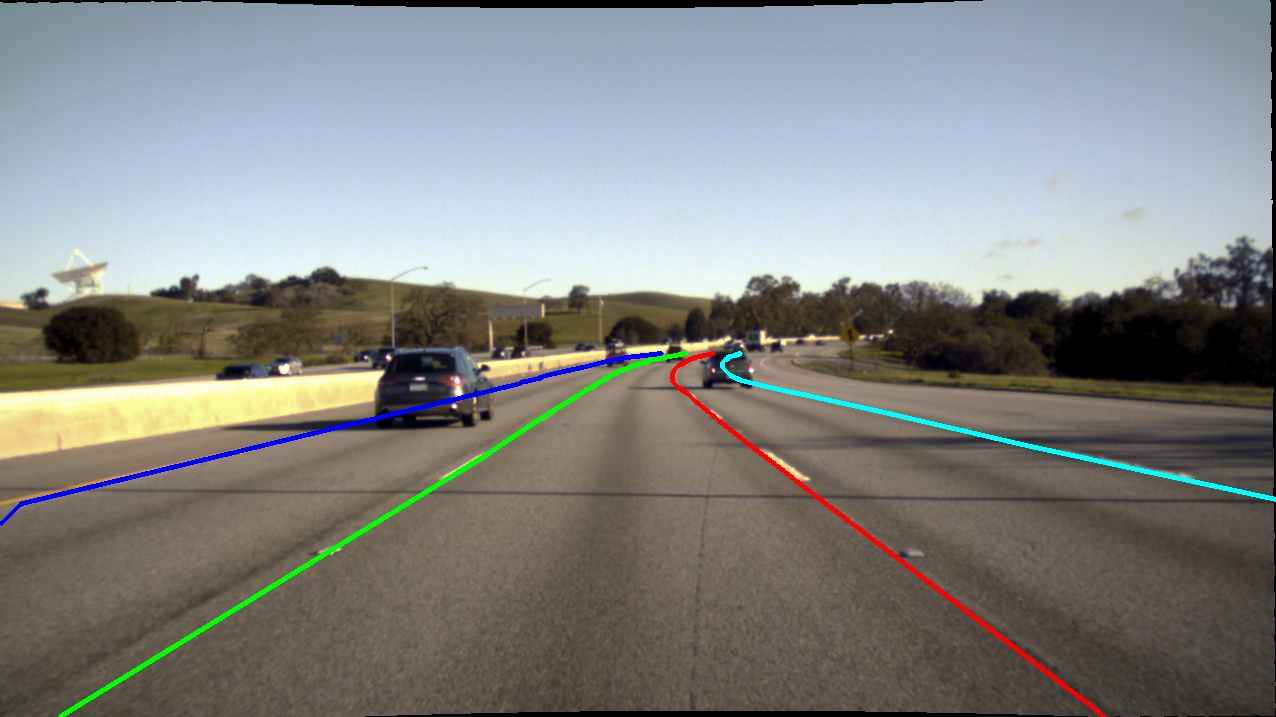}
            \end{minipage}
        \end{subfigure}
        \begin{subfigure}{\pagewidth}
            \begin{minipage}{\subwidth}
                    \includegraphics[width=\imgwidth, height=\imgheight]{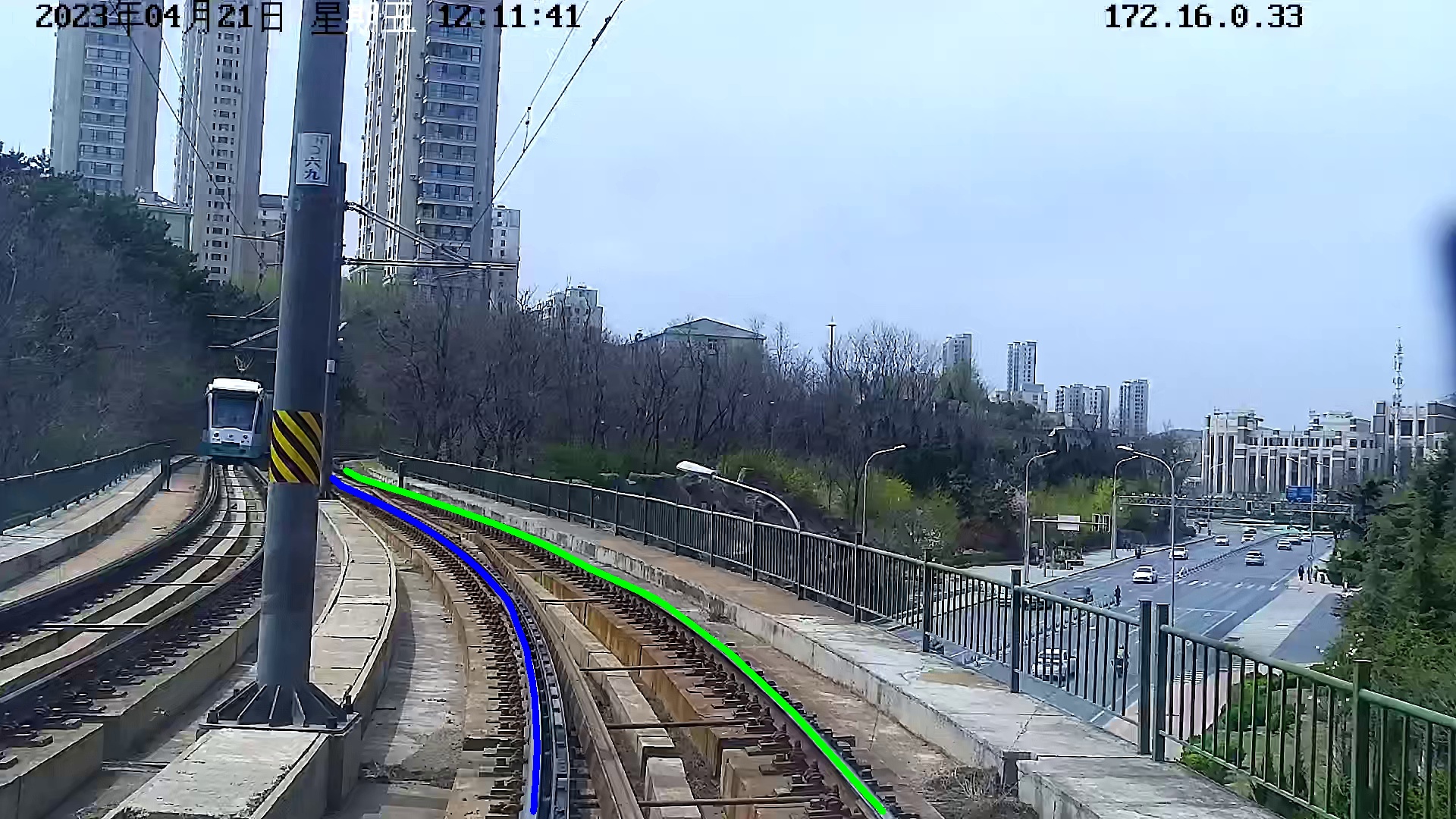}
            \end{minipage}
            \begin{minipage}{\subwidth}
                    \includegraphics[width=\imgwidth, height=\imgheight]{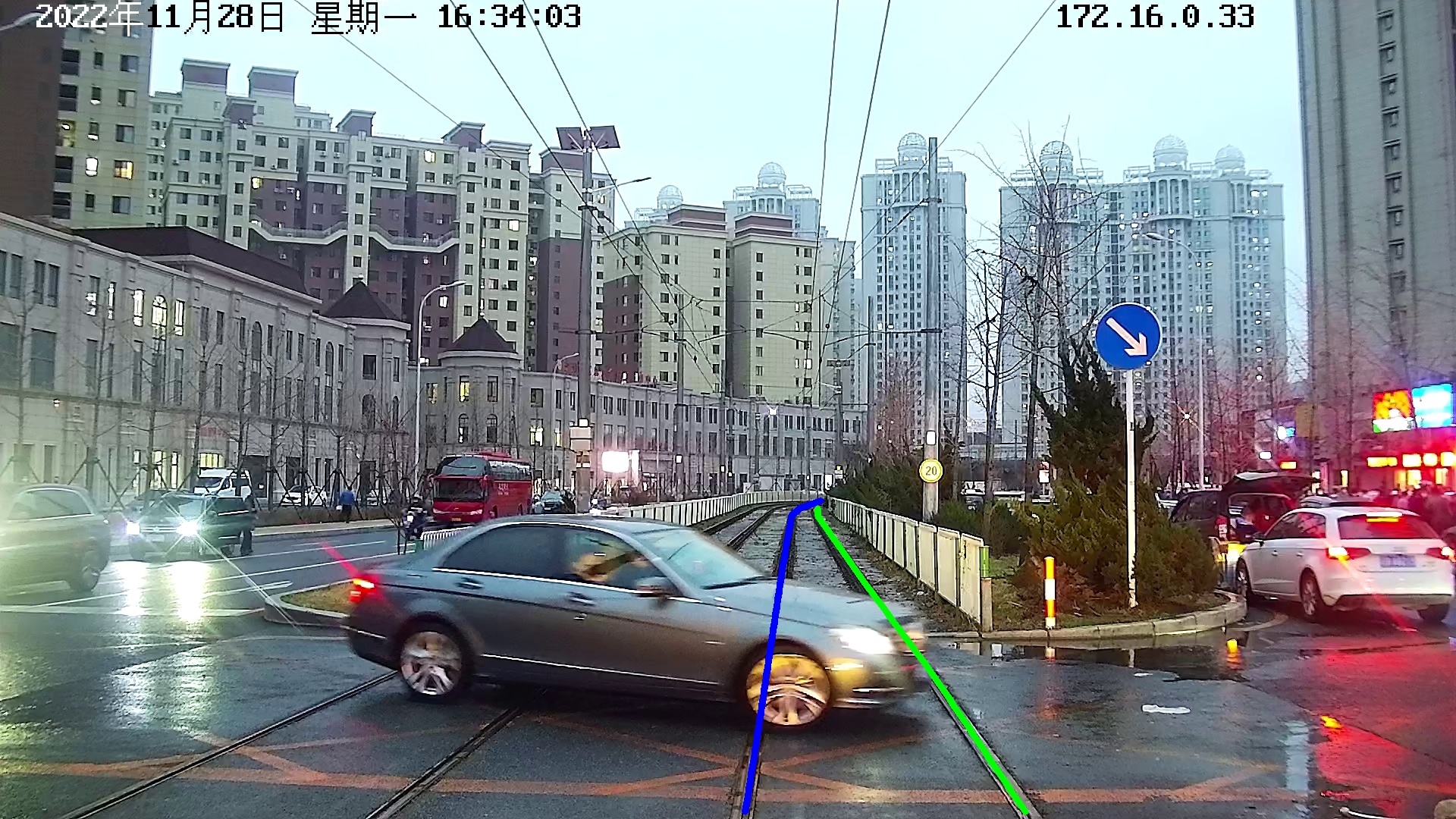}
            \end{minipage}
        \end{subfigure}
        \vspace{0.5em}
        
        \begin{subfigure}{\pagewidth}
            \raisebox{-1.5em}{\rotatebox{90}{\small{Anchors}}}
            \begin{minipage}{\subwidth}
                    \includegraphics[width=\imgwidth, height=\imgheight]{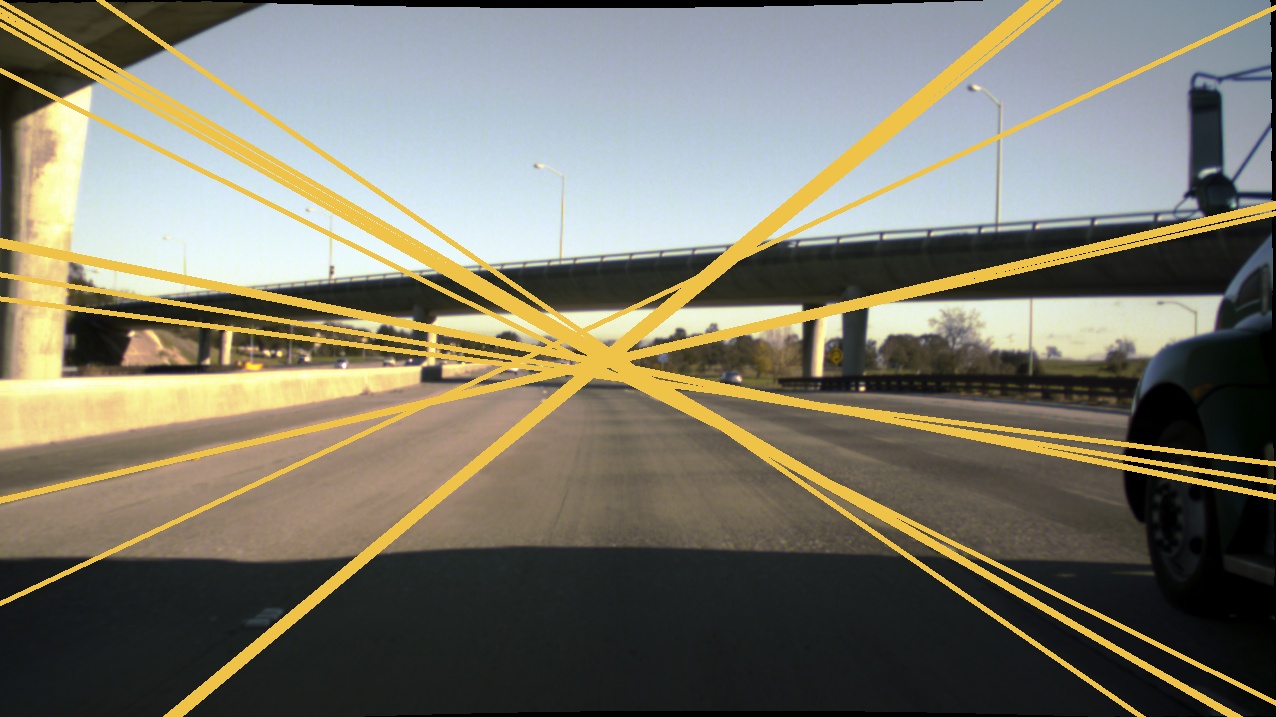}
            \end{minipage}
            \begin{minipage}{\subwidth}
                    \includegraphics[width=\imgwidth, height=\imgheight]{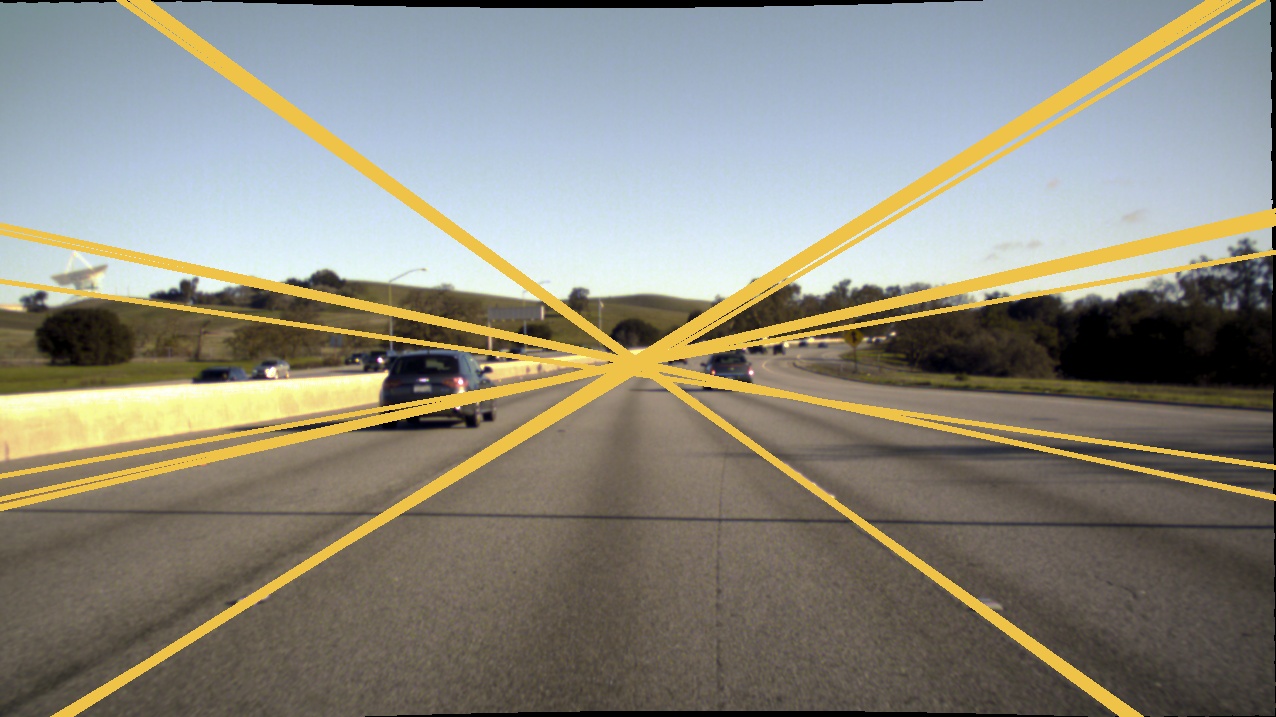}
            \end{minipage}
        \end{subfigure}
        \begin{subfigure}{\pagewidth}
            \begin{minipage}{\subwidth}
                    \includegraphics[width=\imgwidth, height=\imgheight]{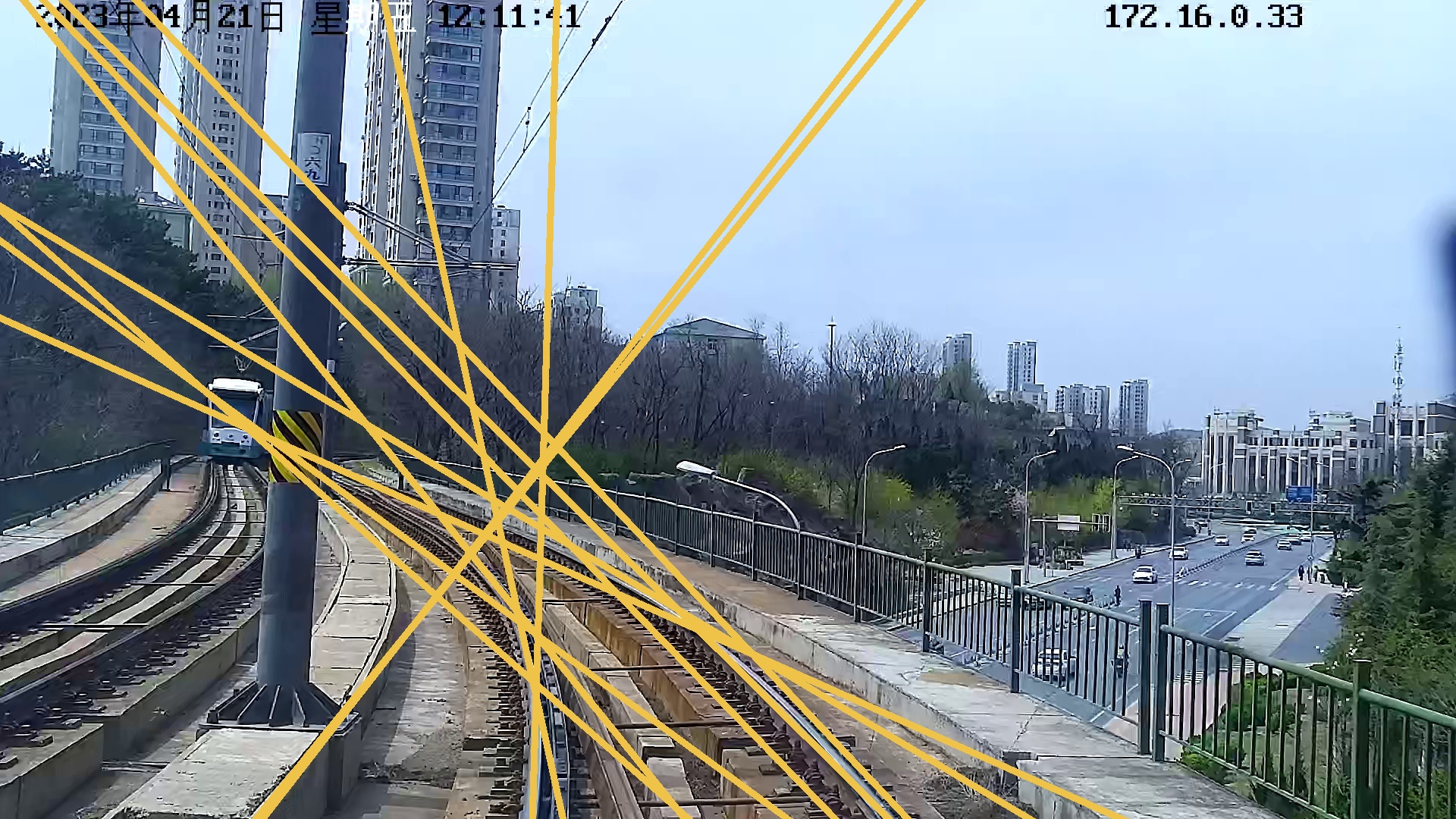}
            \end{minipage}
            \begin{minipage}{\subwidth}
                    \includegraphics[width=\imgwidth, height=\imgheight]{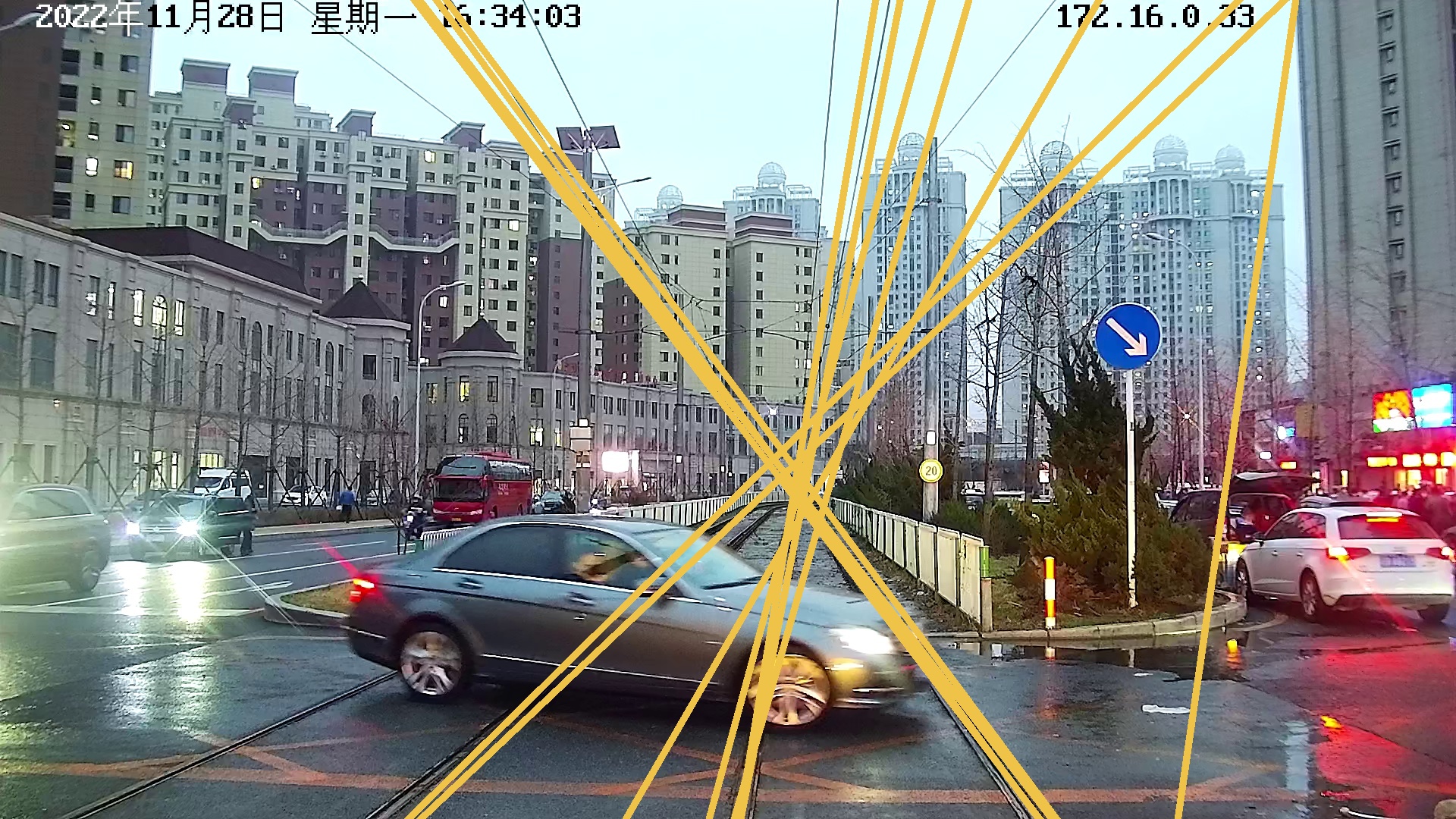}
            \end{minipage}
        \end{subfigure}
        \vspace{0.5em}

        \begin{subfigure}{\pagewidth}
            \raisebox{-2em}{\rotatebox{90}{\small{Predictions}}}
            \begin{minipage}{\subwidth}
                    \includegraphics[width=\imgwidth, height=\imgheight]{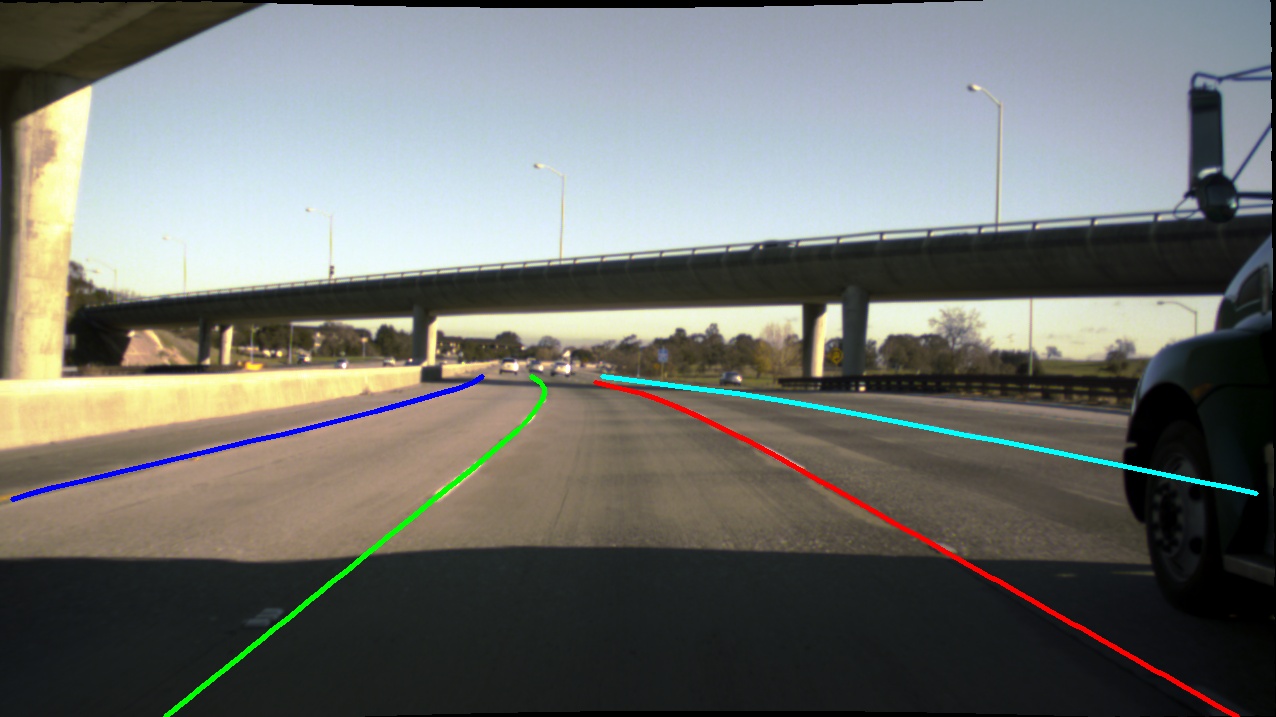}
            \end{minipage}
            \begin{minipage}{\subwidth}
                    \includegraphics[width=\imgwidth, height=\imgheight]{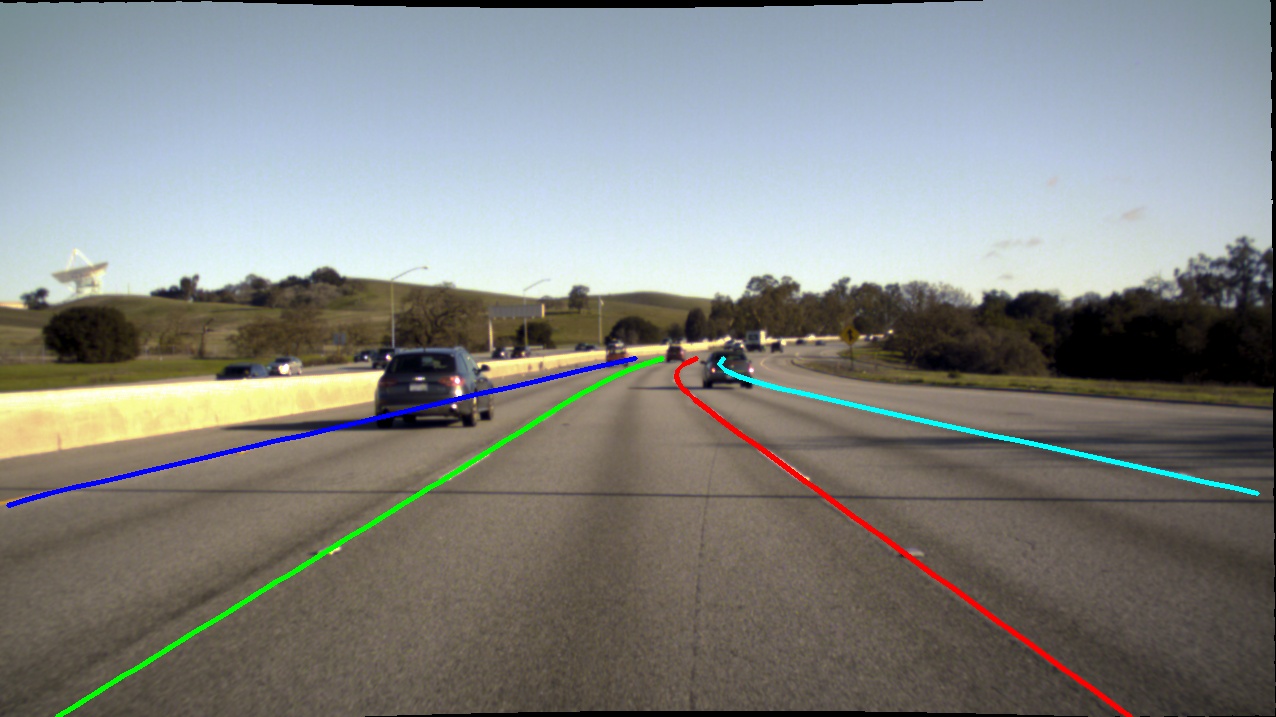}
            \end{minipage}
            \caption{LLAMAS}
        \end{subfigure}
        \begin{subfigure}{\pagewidth}
            \begin{minipage}{\subwidth}
                    \includegraphics[width=\imgwidth, height=\imgheight]{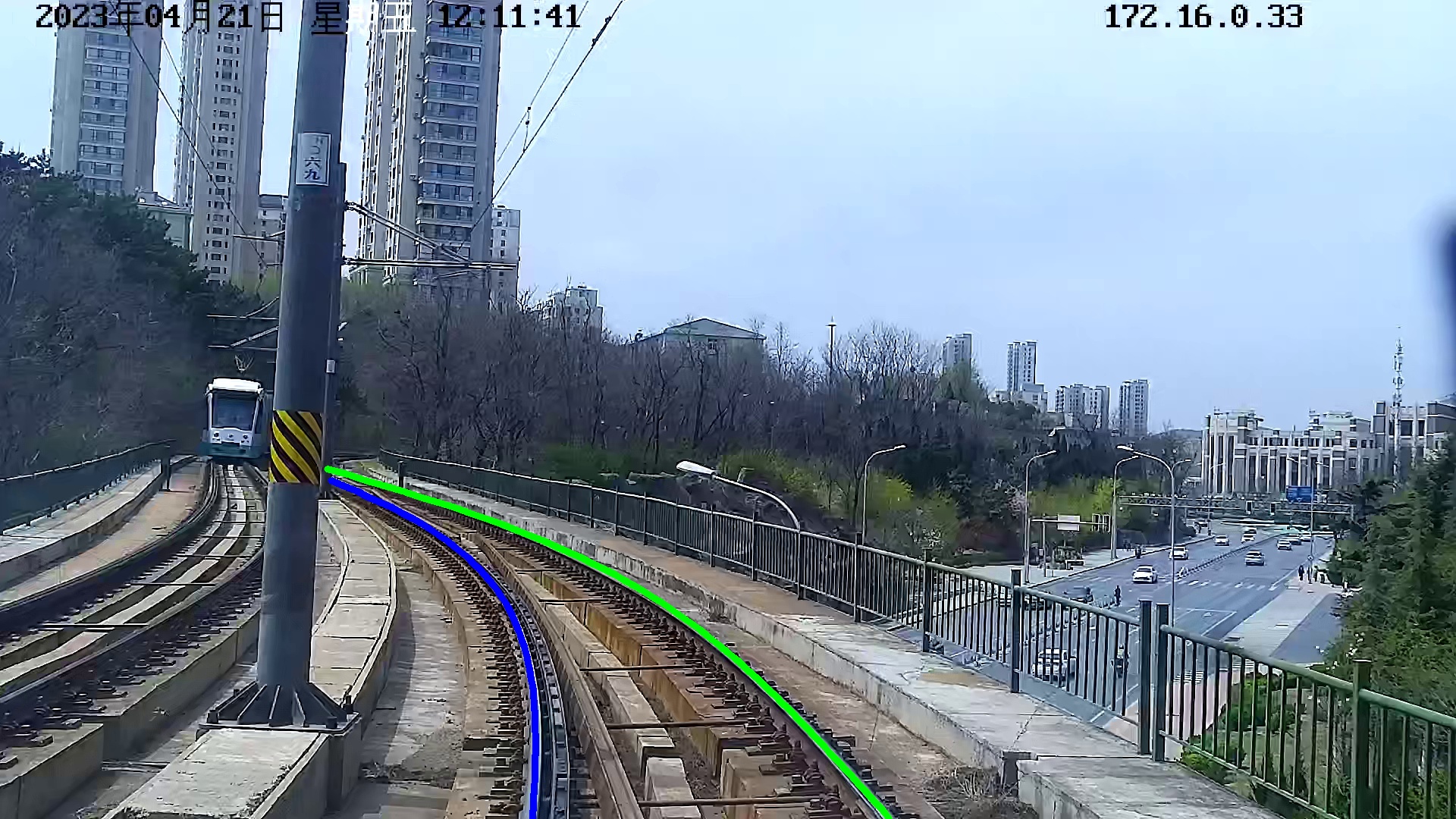}
            \end{minipage}
            \begin{minipage}{\subwidth}
                    \includegraphics[width=\imgwidth, height=\imgheight]{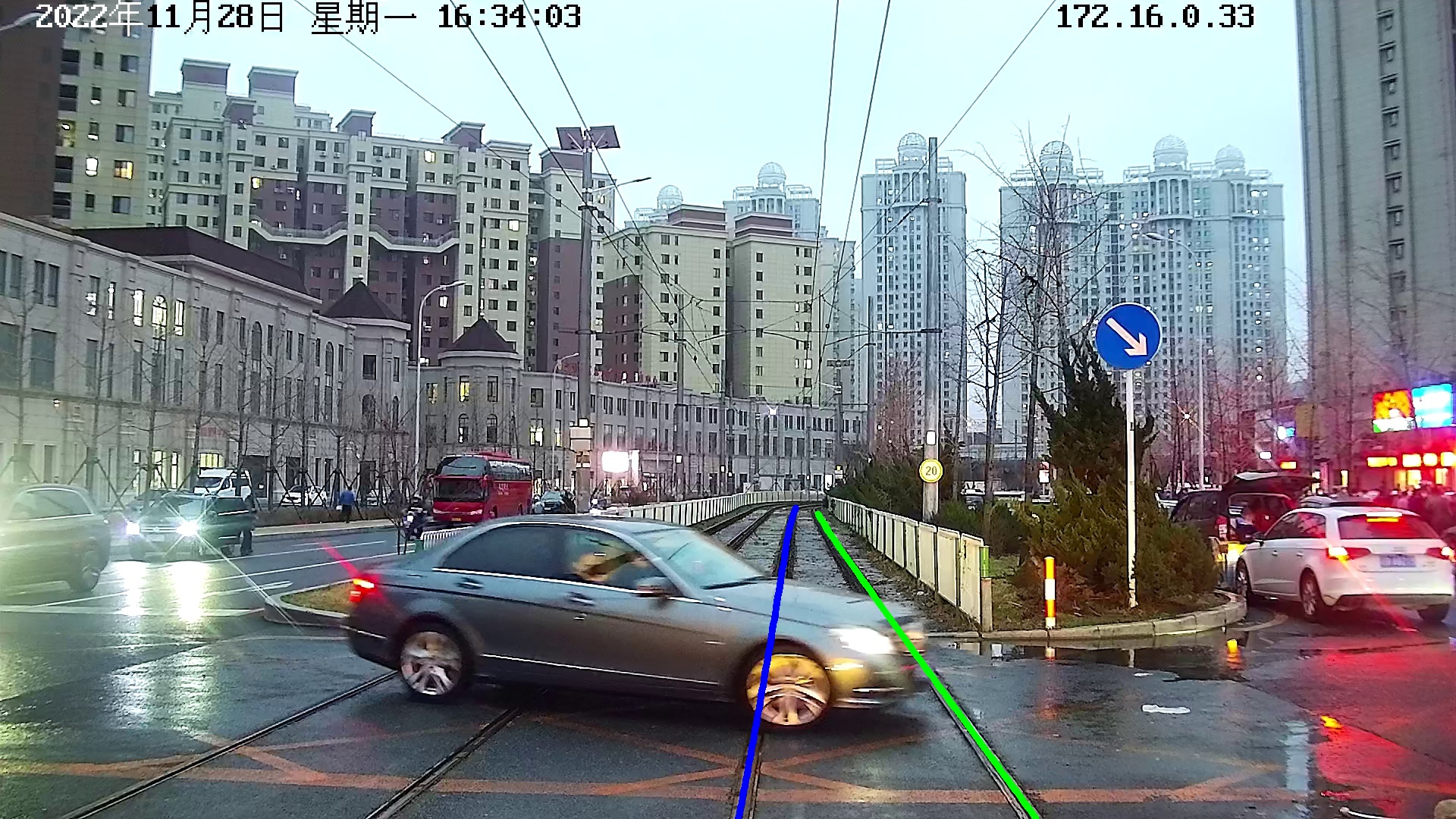}
            \end{minipage}
            \caption{DL-Rail}
        \end{subfigure}
        \vspace{0.5em}
        
        \caption{Visualization of detection outcomes in sparse scenarios of four datasets.}
        \label{vis_sparse}
\end{figure*}

\begin{figure*}[t]
        \centering
        \def\subwidth{0.24\textwidth}
        \def\imgwidth{\linewidth}
        \def\imgheight{0.5625\linewidth}

        \begin{subfigure}{\subwidth}
                \includegraphics[width=\imgwidth, height=\imgheight]{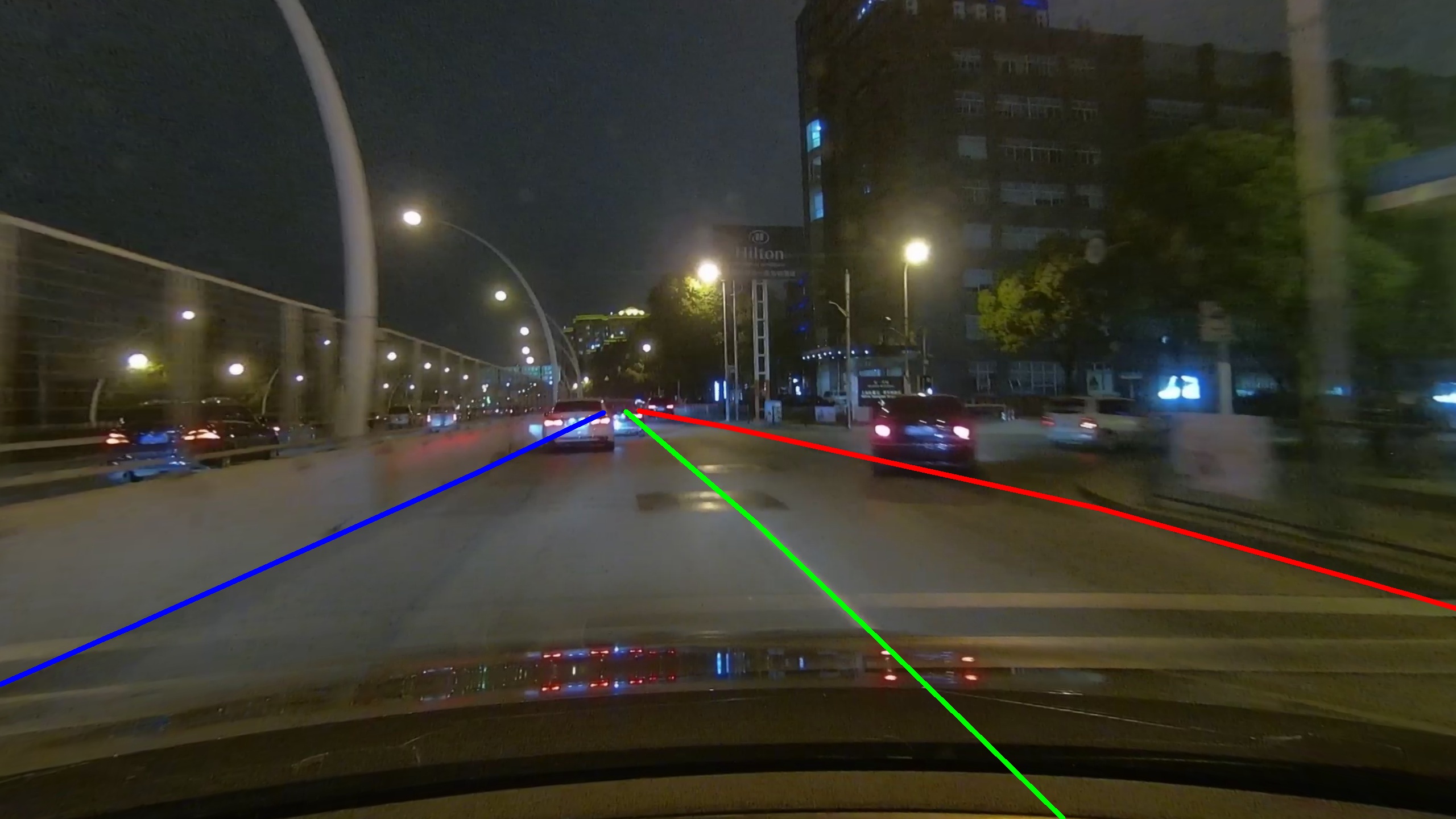}
        \end{subfigure}
        \begin{subfigure}{\subwidth}
                \includegraphics[width=\imgwidth, height=\imgheight]{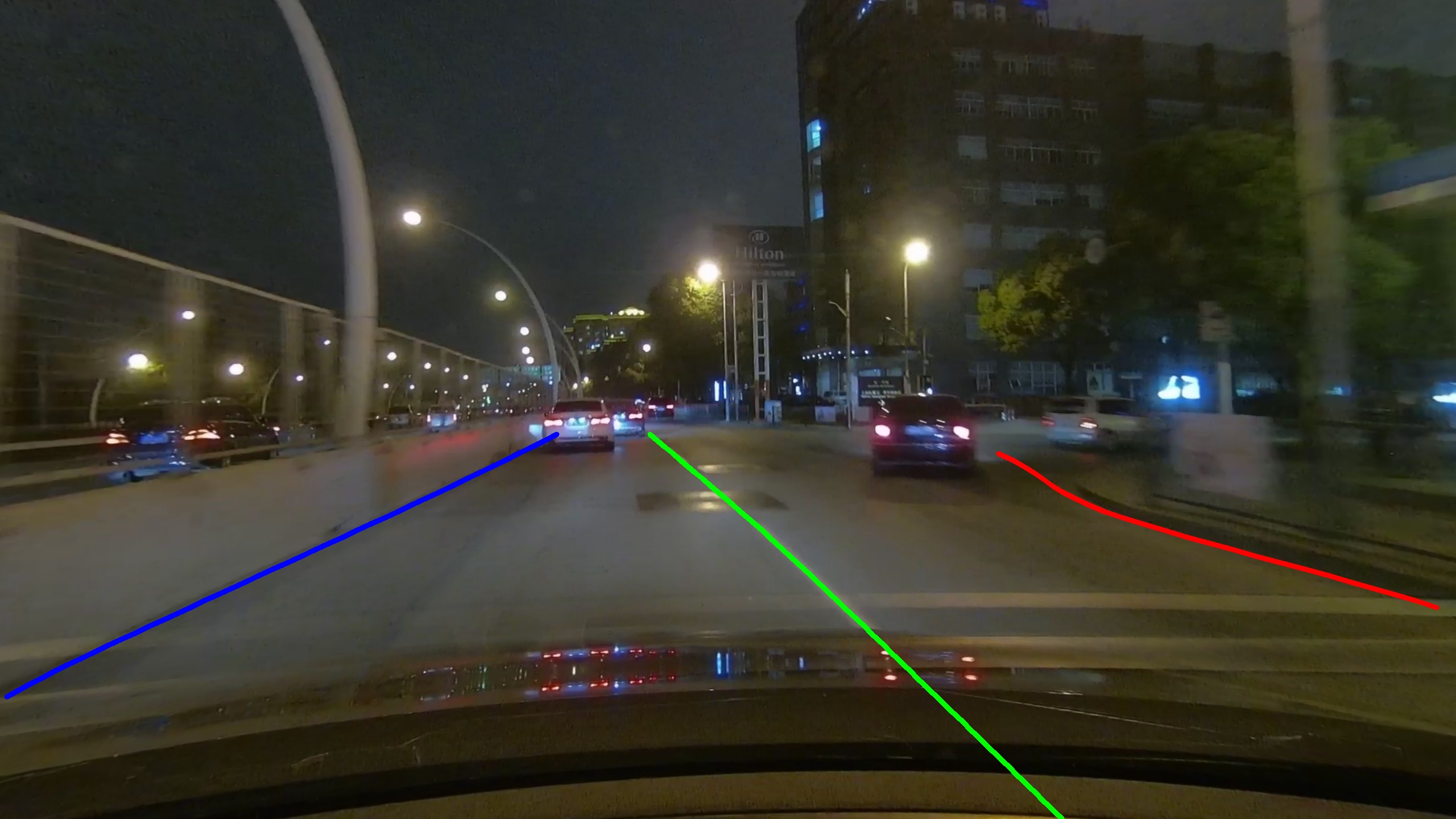}
        \end{subfigure}
        \begin{subfigure}{\subwidth}
                \includegraphics[width=\imgwidth, height=\imgheight]{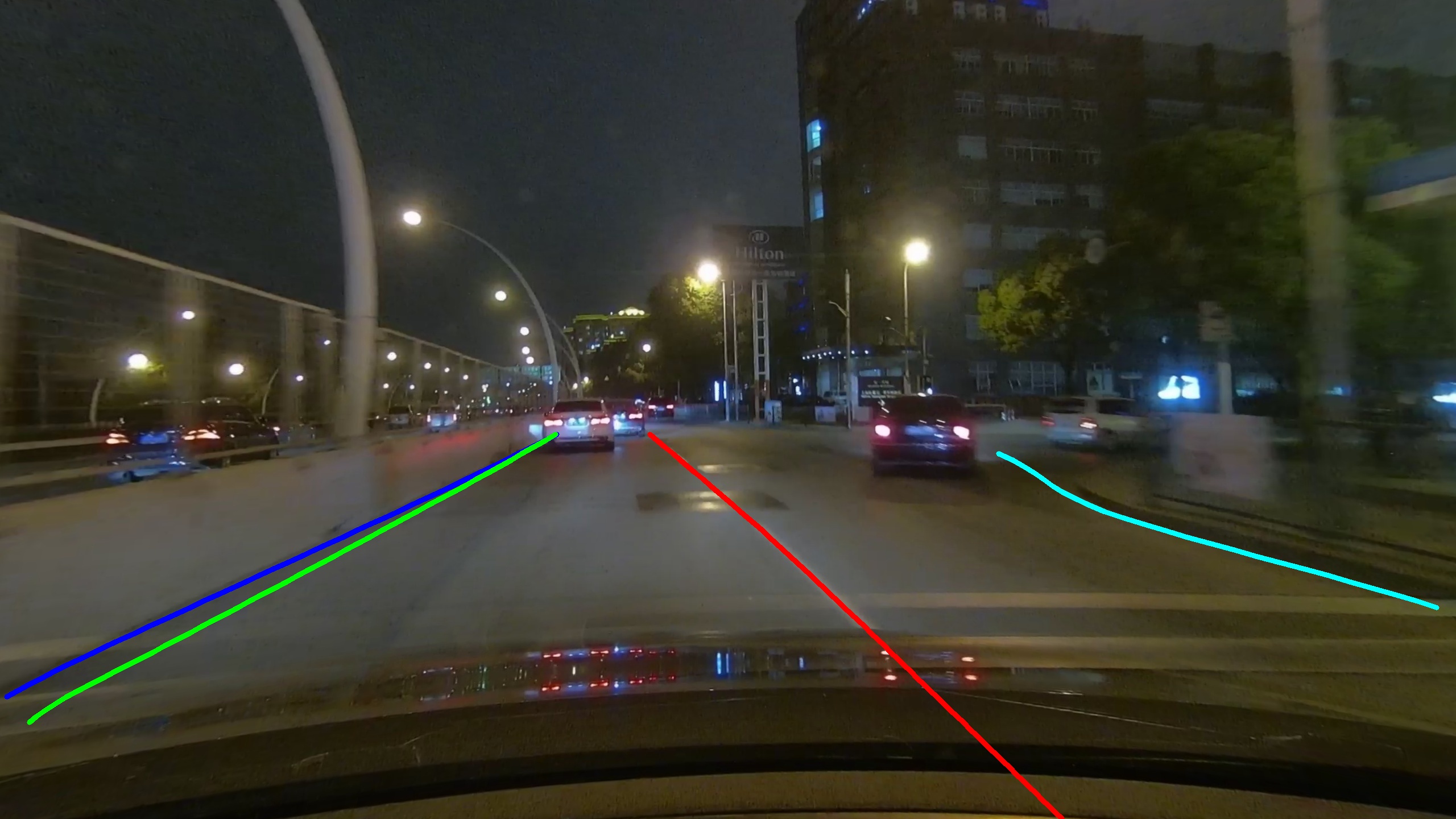}
        \end{subfigure}
        \begin{subfigure}{\subwidth}
                \includegraphics[width=\imgwidth, height=\imgheight]{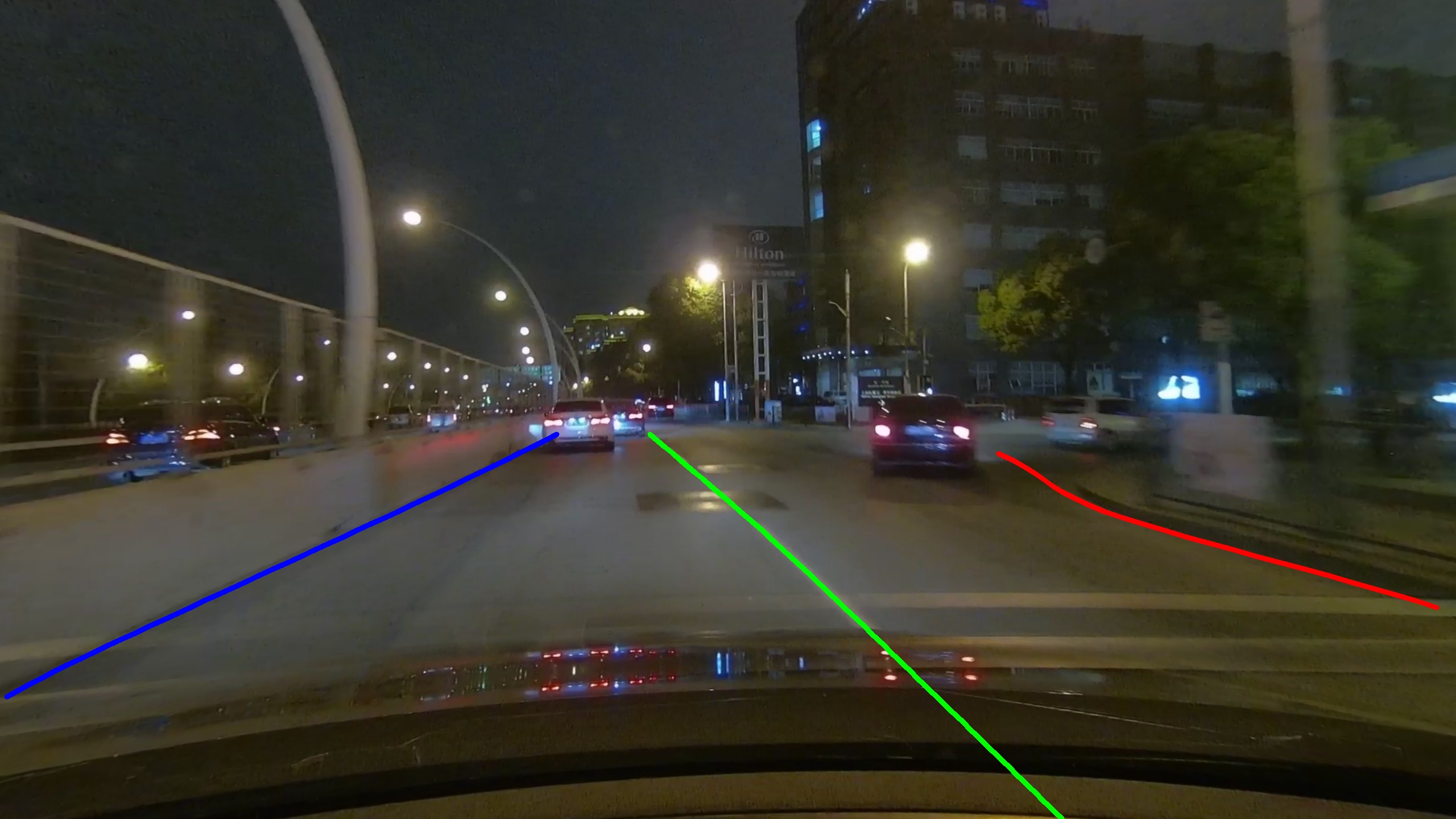}
        \end{subfigure}
        \vspace{0.5em}

        \begin{subfigure}{\subwidth}
                \includegraphics[width=\imgwidth, height=\imgheight]{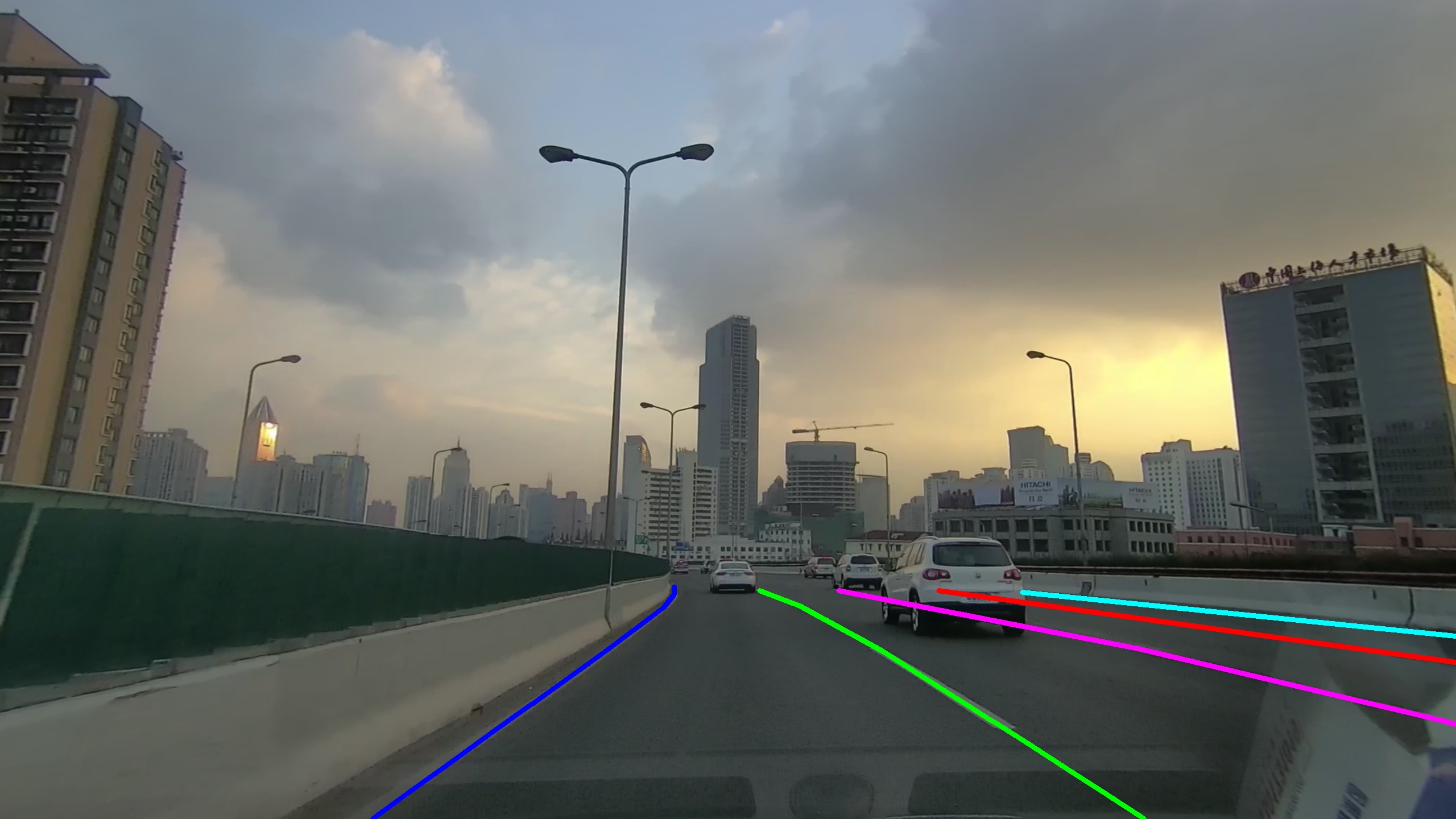}
        \end{subfigure}
        \begin{subfigure}{\subwidth}
                \includegraphics[width=\imgwidth, height=\imgheight]{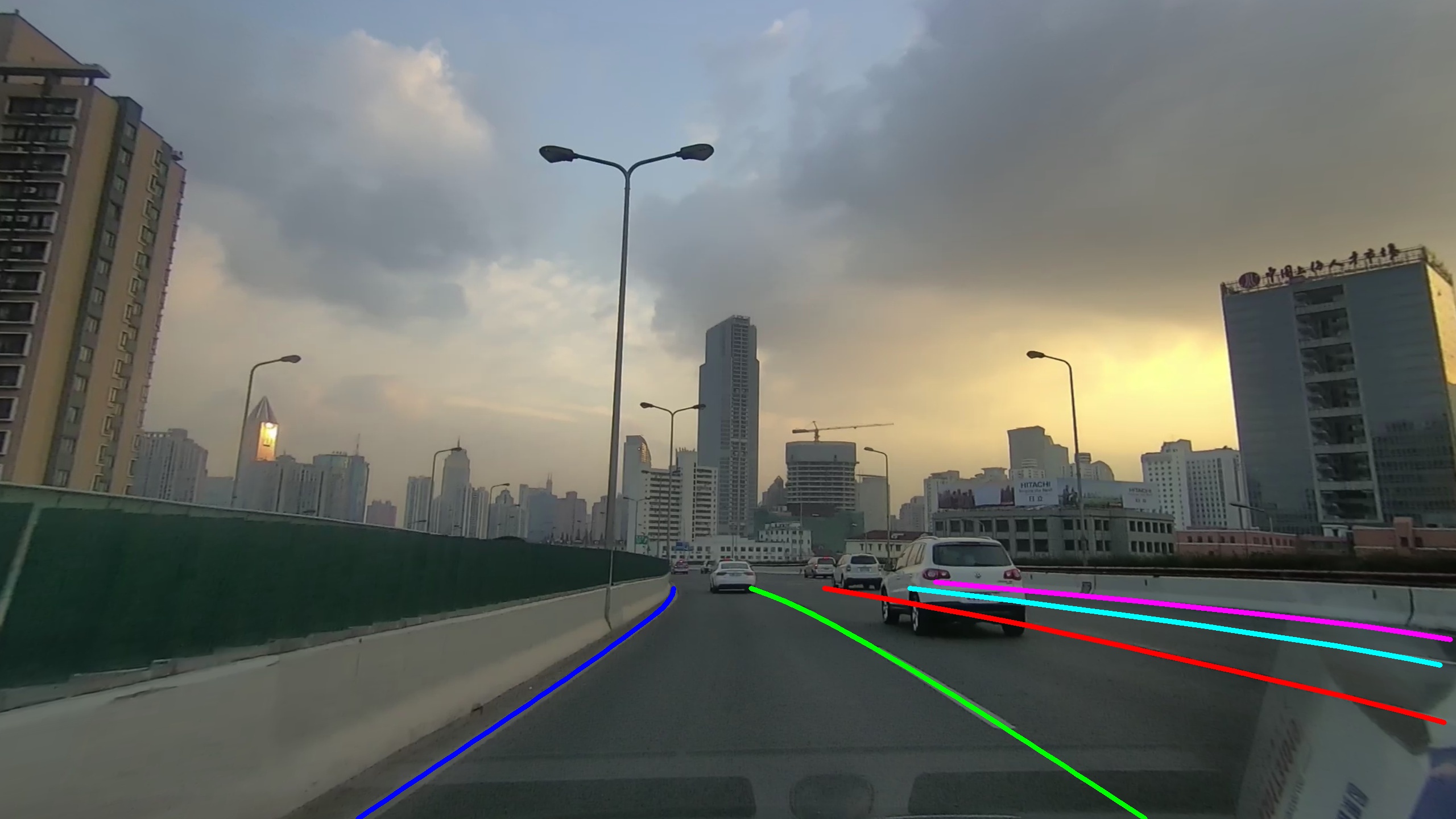}
        \end{subfigure}
        \begin{subfigure}{\subwidth}
                \includegraphics[width=\imgwidth, height=\imgheight]{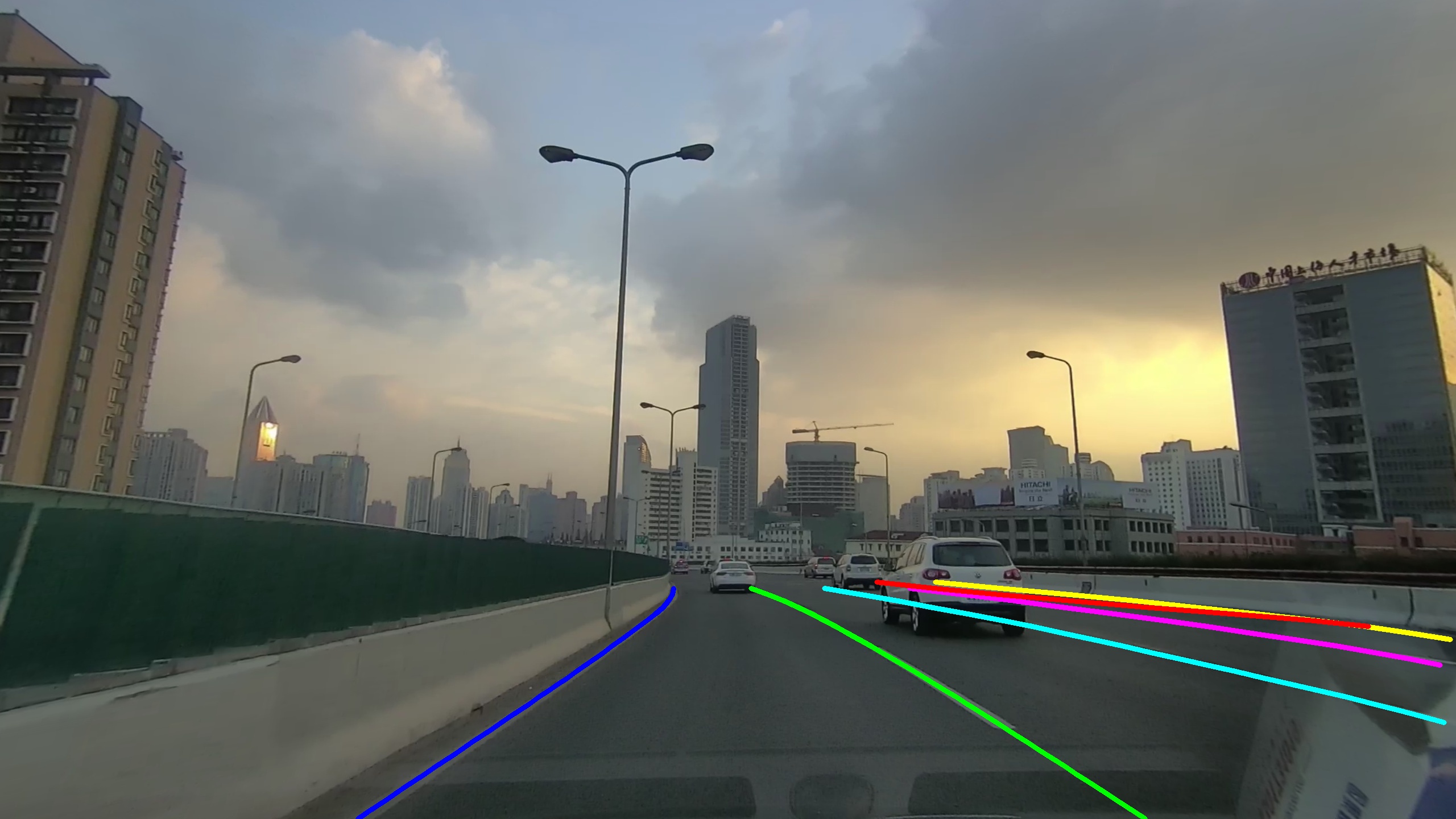}
        \end{subfigure}
        \begin{subfigure}{\subwidth}
                \includegraphics[width=\imgwidth, height=\imgheight]{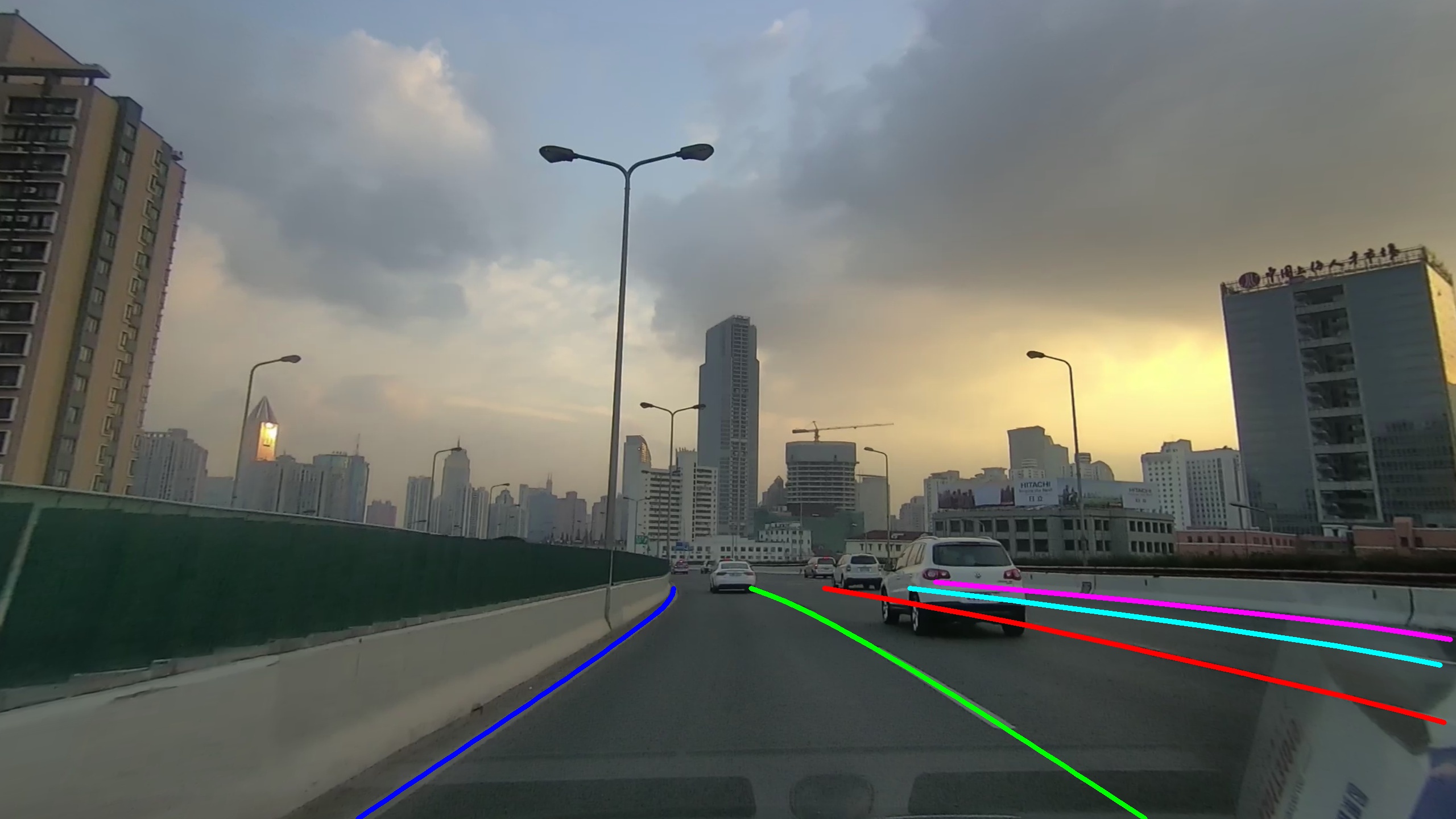}
        \end{subfigure}
        \vspace{0.5em}

        \begin{subfigure}{\subwidth}
                \includegraphics[width=\imgwidth, height=\imgheight]{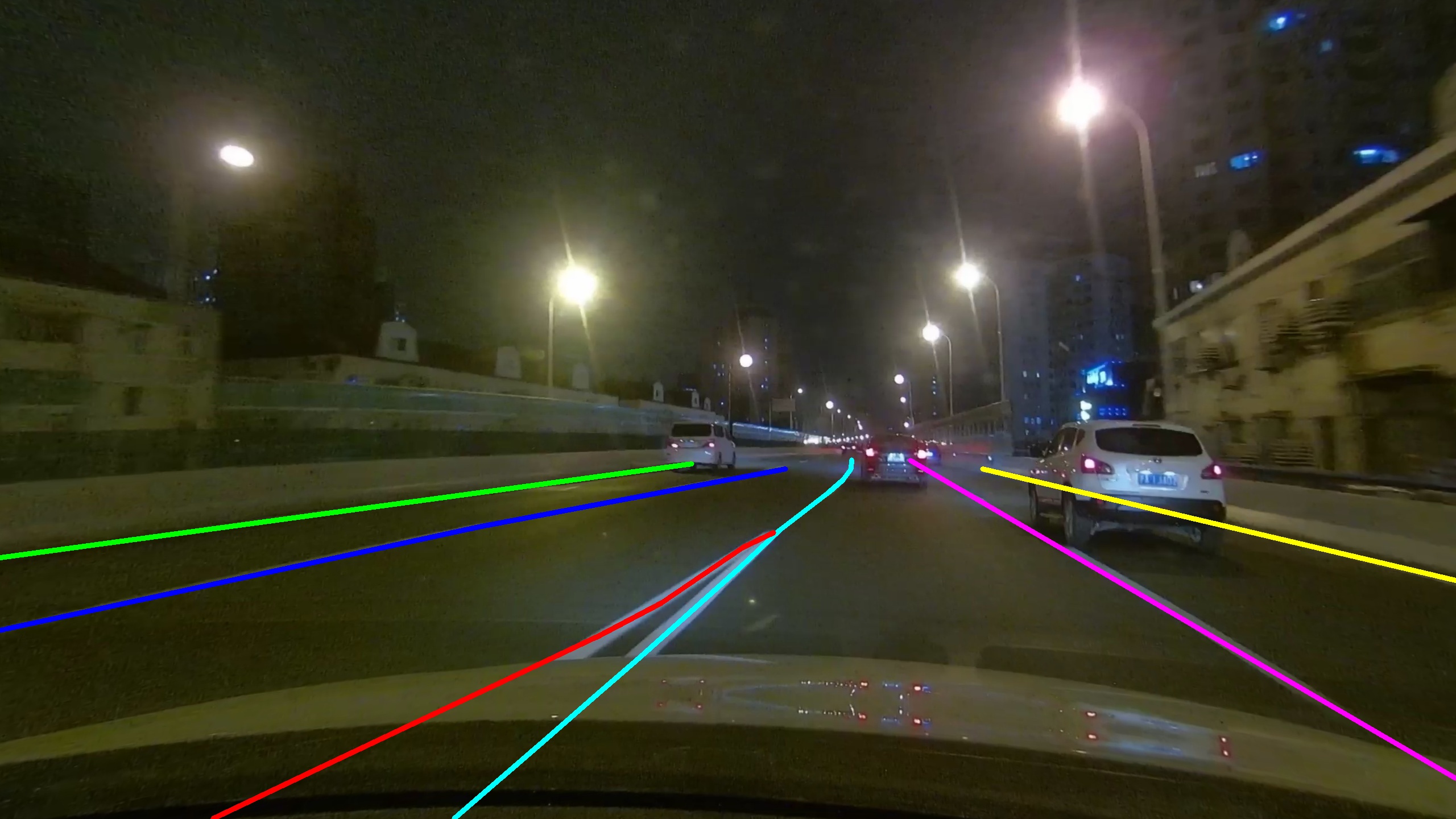}
        \end{subfigure}
        \begin{subfigure}{\subwidth}
                \includegraphics[width=\imgwidth, height=\imgheight]{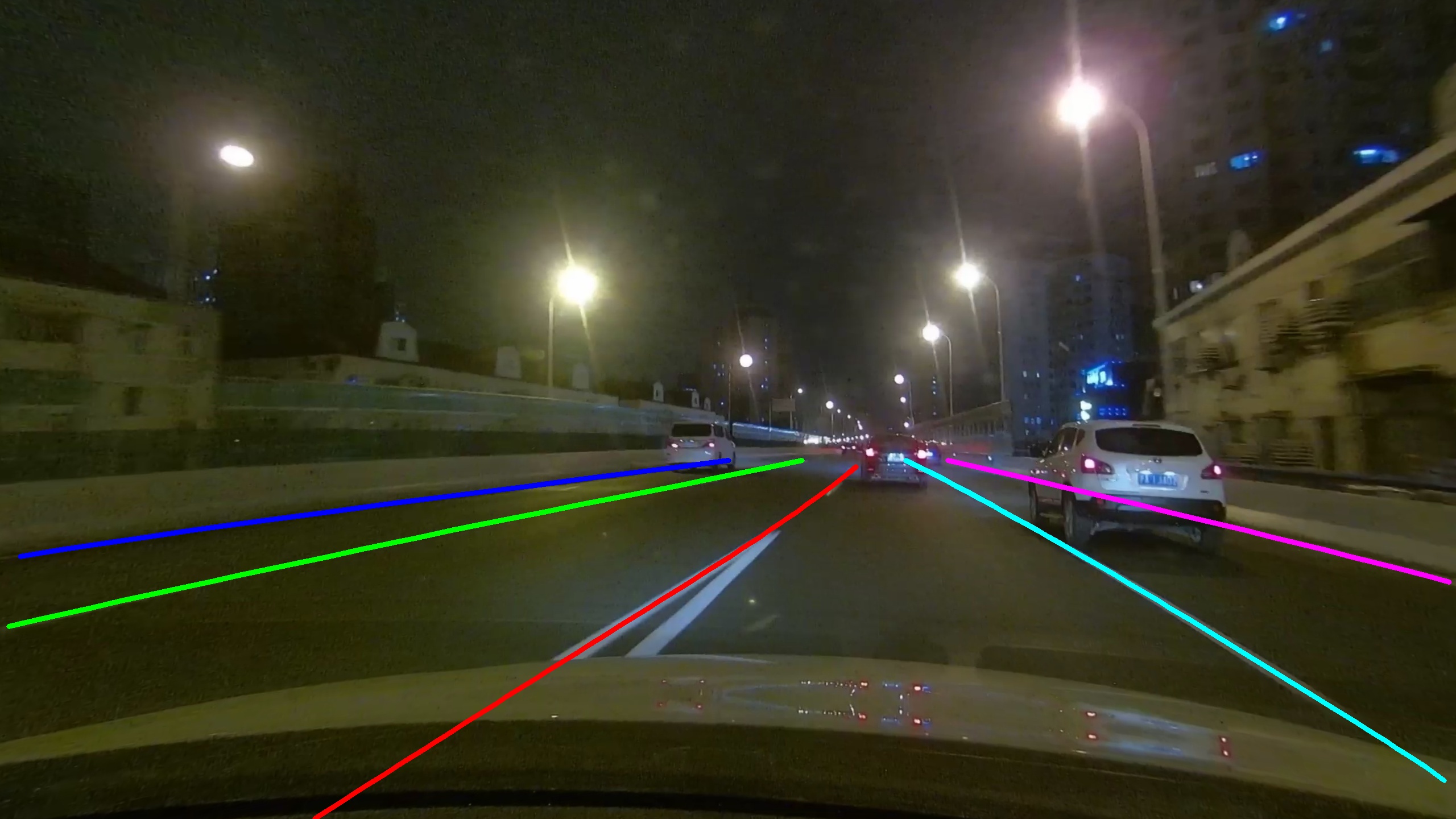}
        \end{subfigure}
        \begin{subfigure}{\subwidth}
                \includegraphics[width=\imgwidth, height=\imgheight]{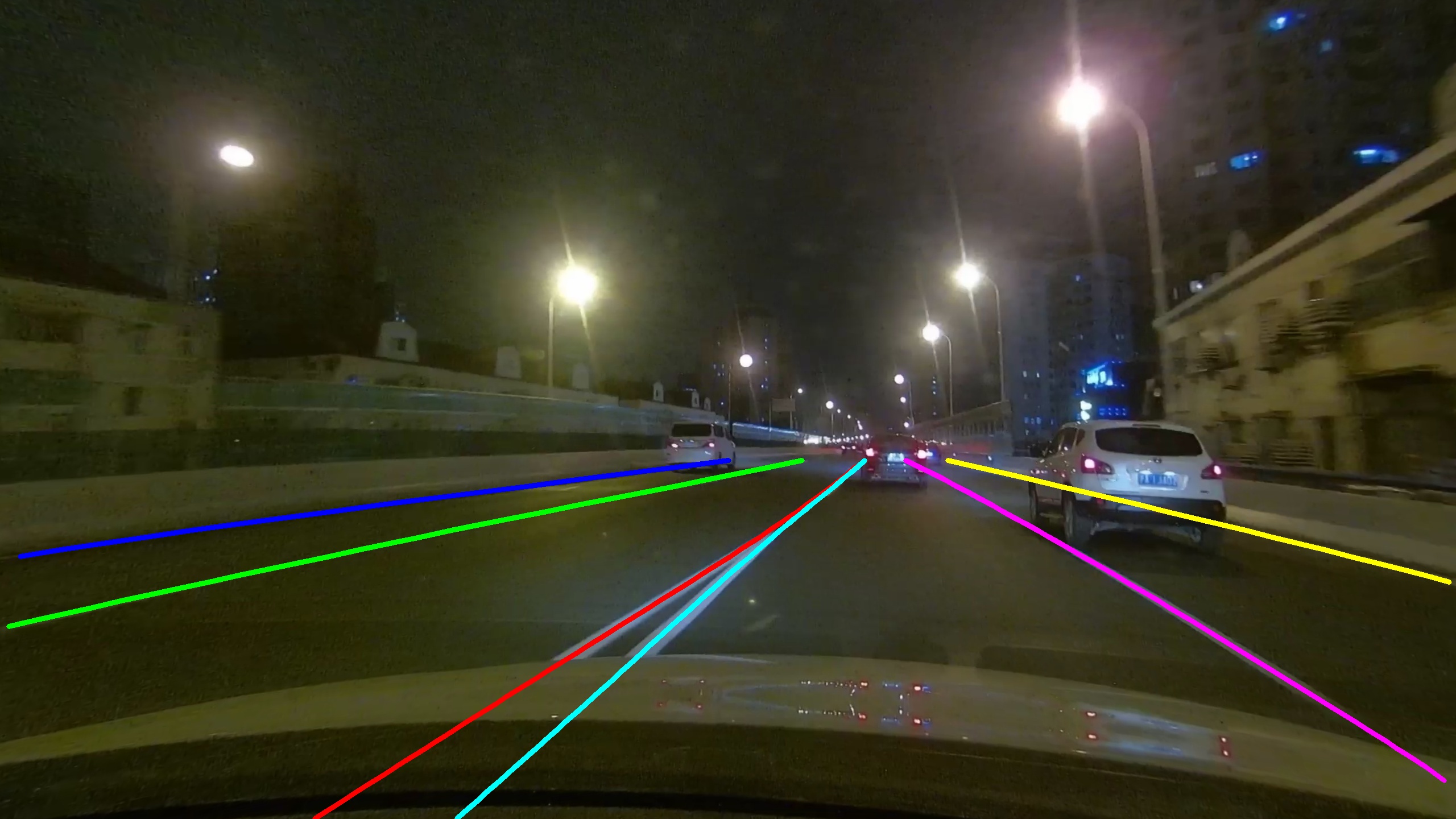}
        \end{subfigure}
        \begin{subfigure}{\subwidth}
                \includegraphics[width=\imgwidth, height=\imgheight]{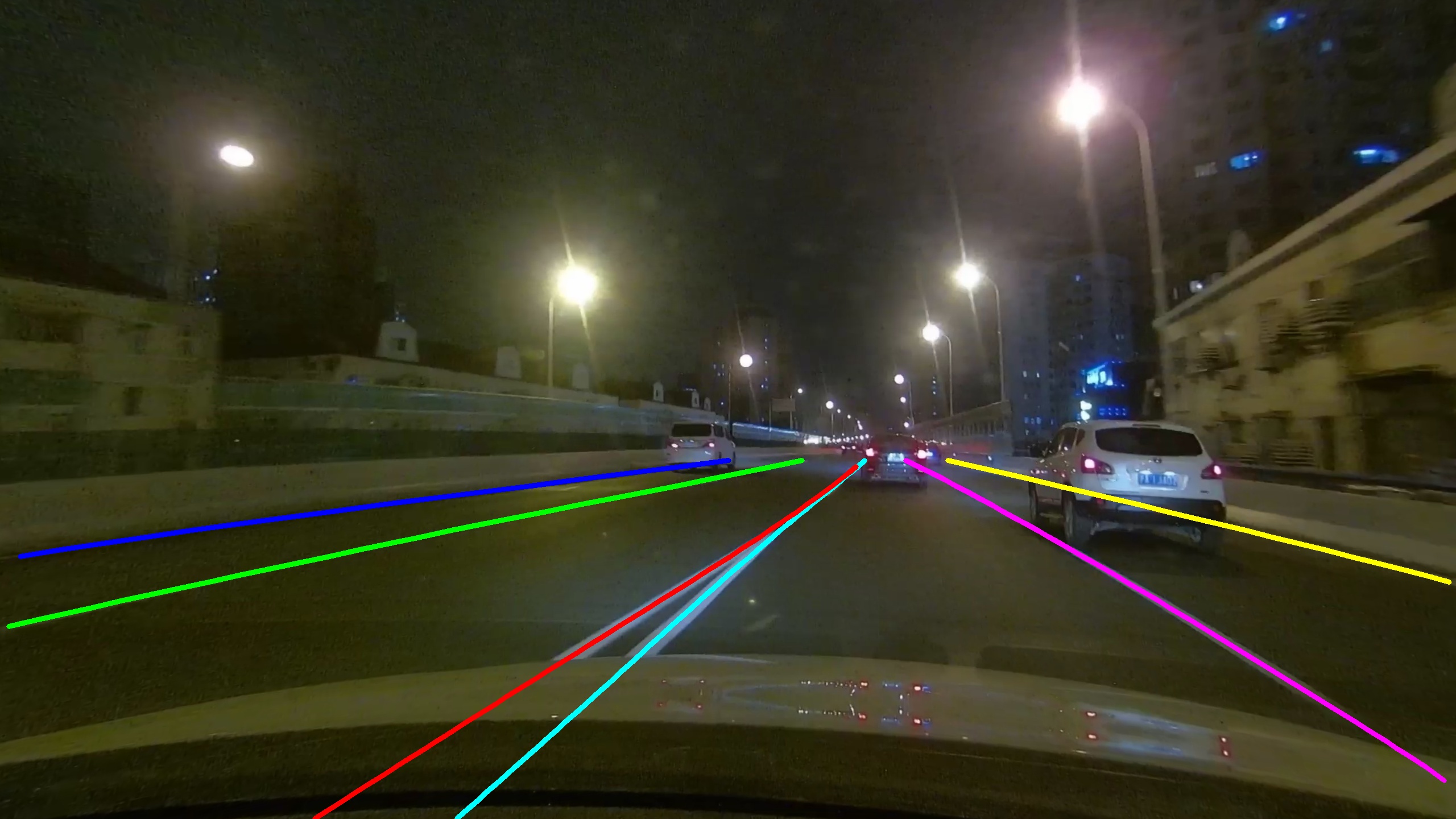}
        \end{subfigure}
        \vspace{0.5em}

        \begin{subfigure}{\subwidth}
                \includegraphics[width=\imgwidth, height=\imgheight]{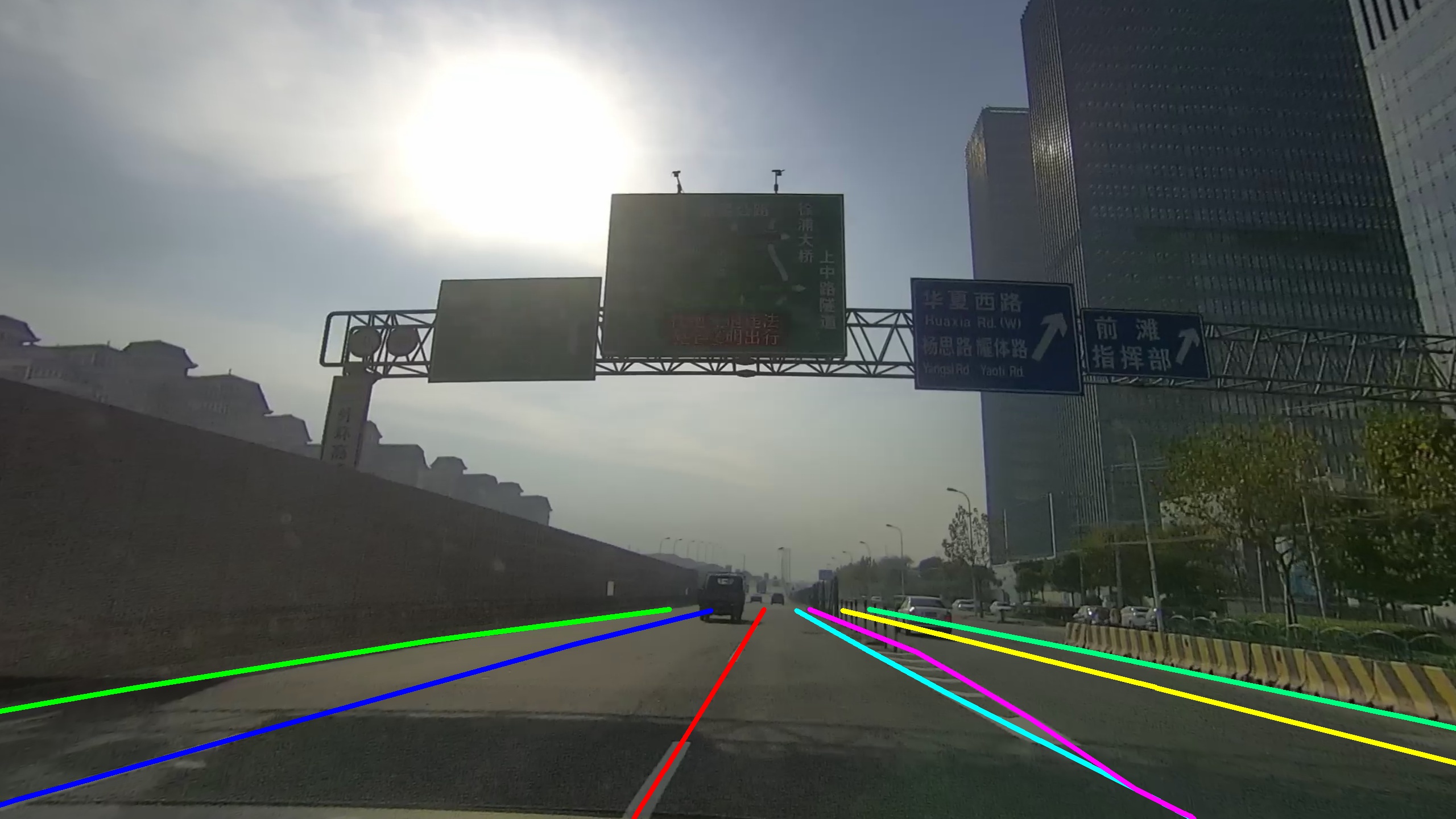}
        \end{subfigure}
        \begin{subfigure}{\subwidth}
                \includegraphics[width=\imgwidth, height=\imgheight]{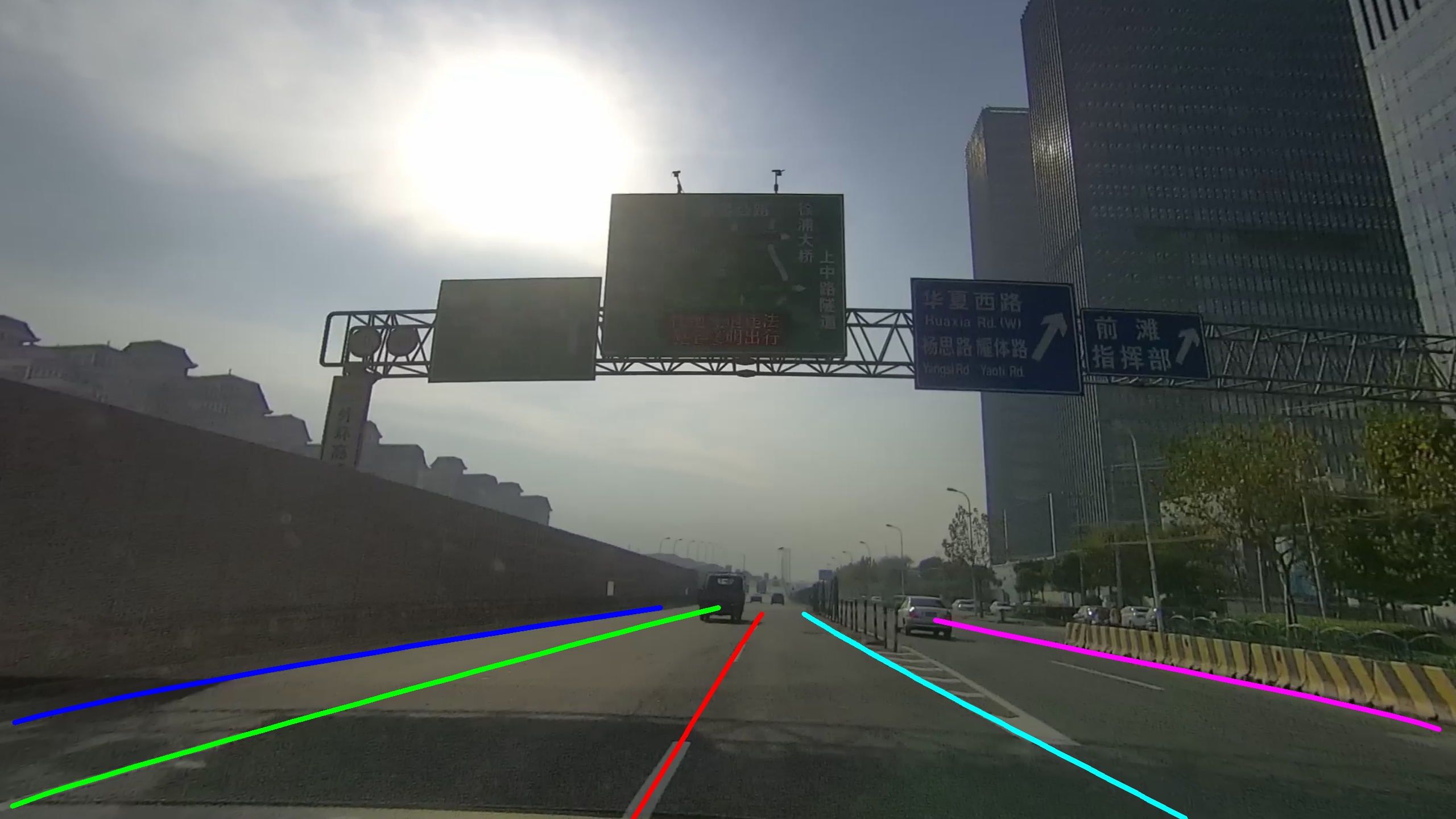}
        \end{subfigure}
        \begin{subfigure}{\subwidth}
                \includegraphics[width=\imgwidth, height=\imgheight]{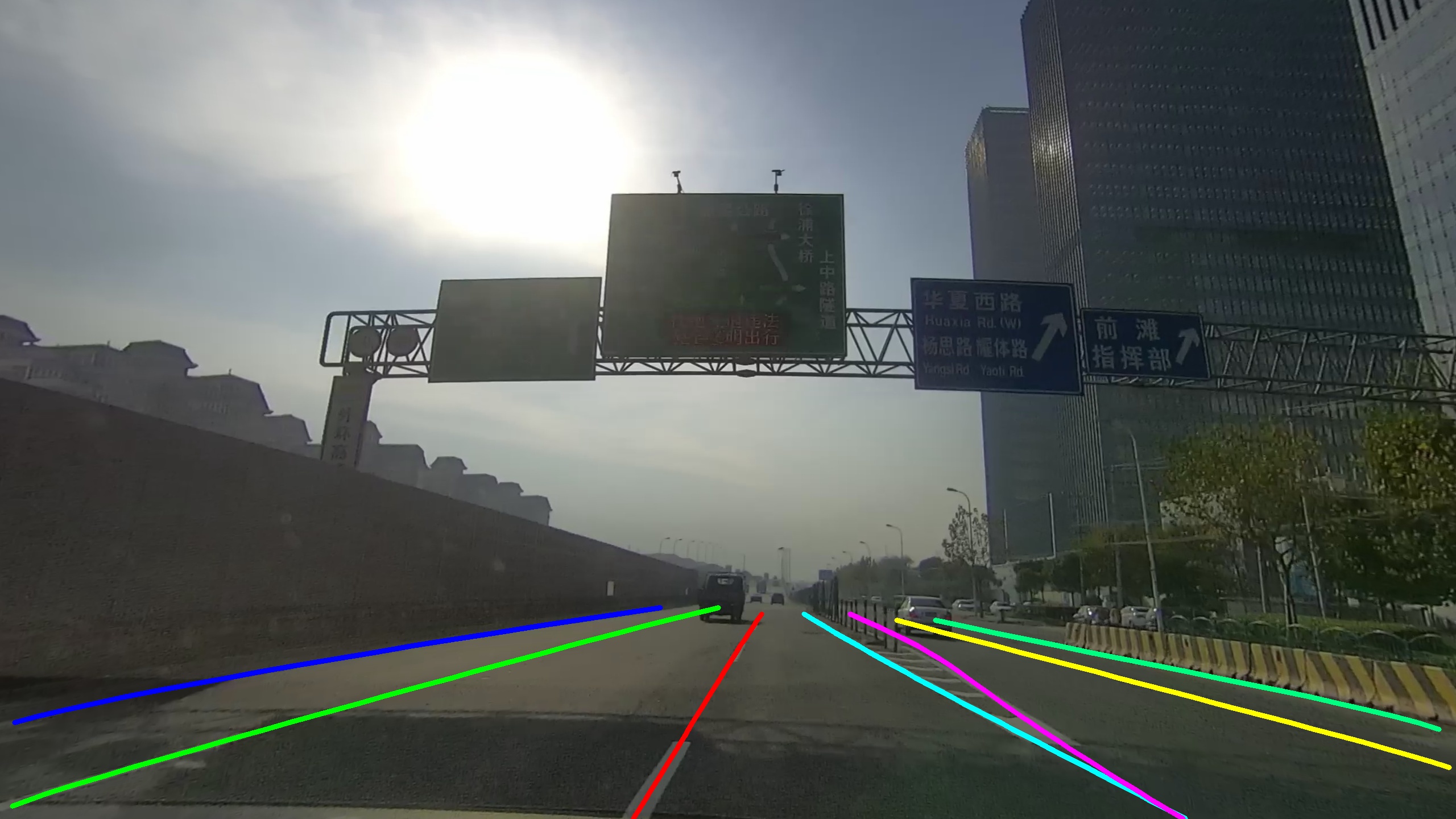}
        \end{subfigure}
        \begin{subfigure}{\subwidth}
                \includegraphics[width=\imgwidth, height=\imgheight]{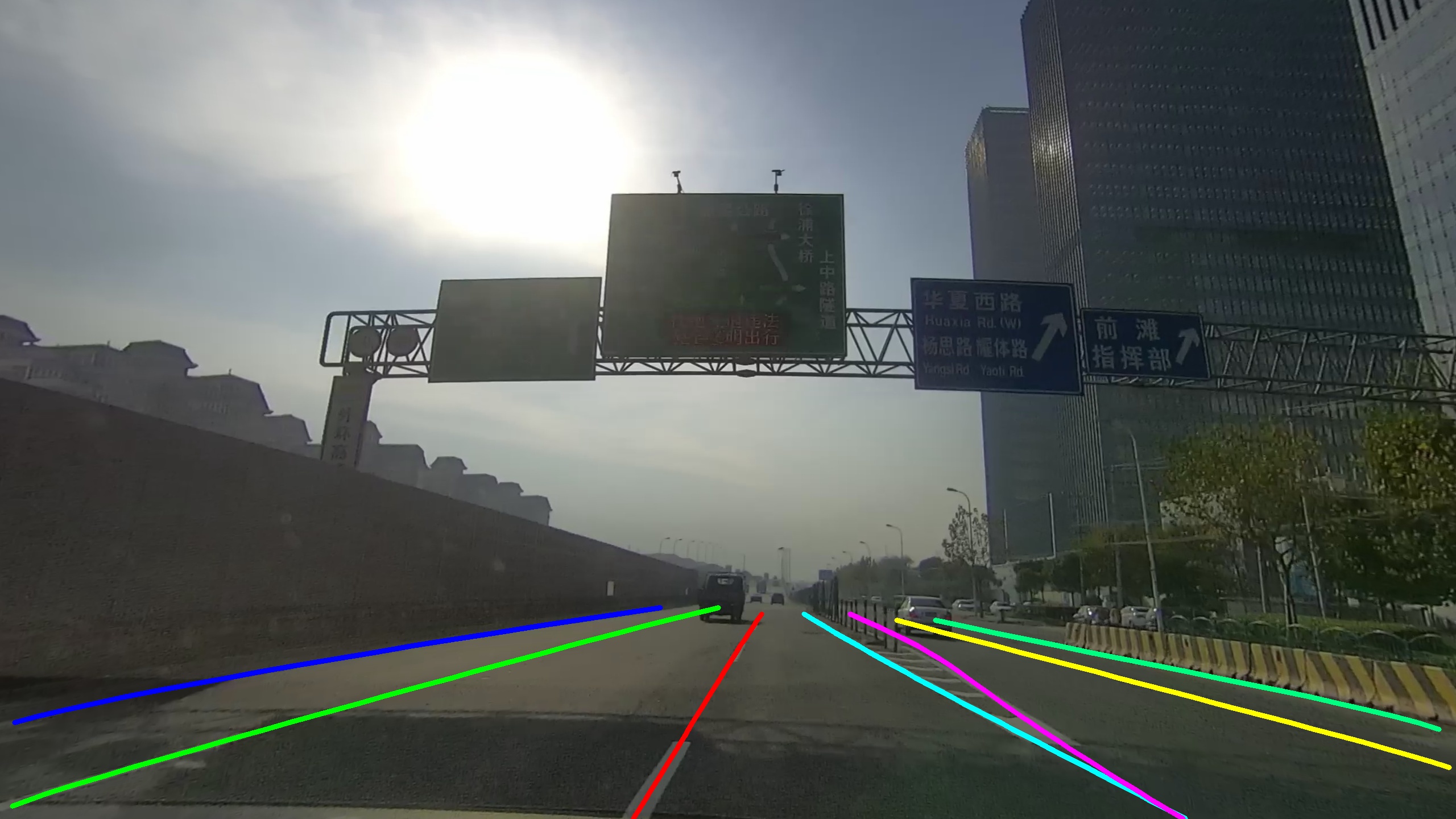}
        \end{subfigure}
        \vspace{0.5em}

        \begin{subfigure}{\subwidth}
                \includegraphics[width=\imgwidth, height=\imgheight]{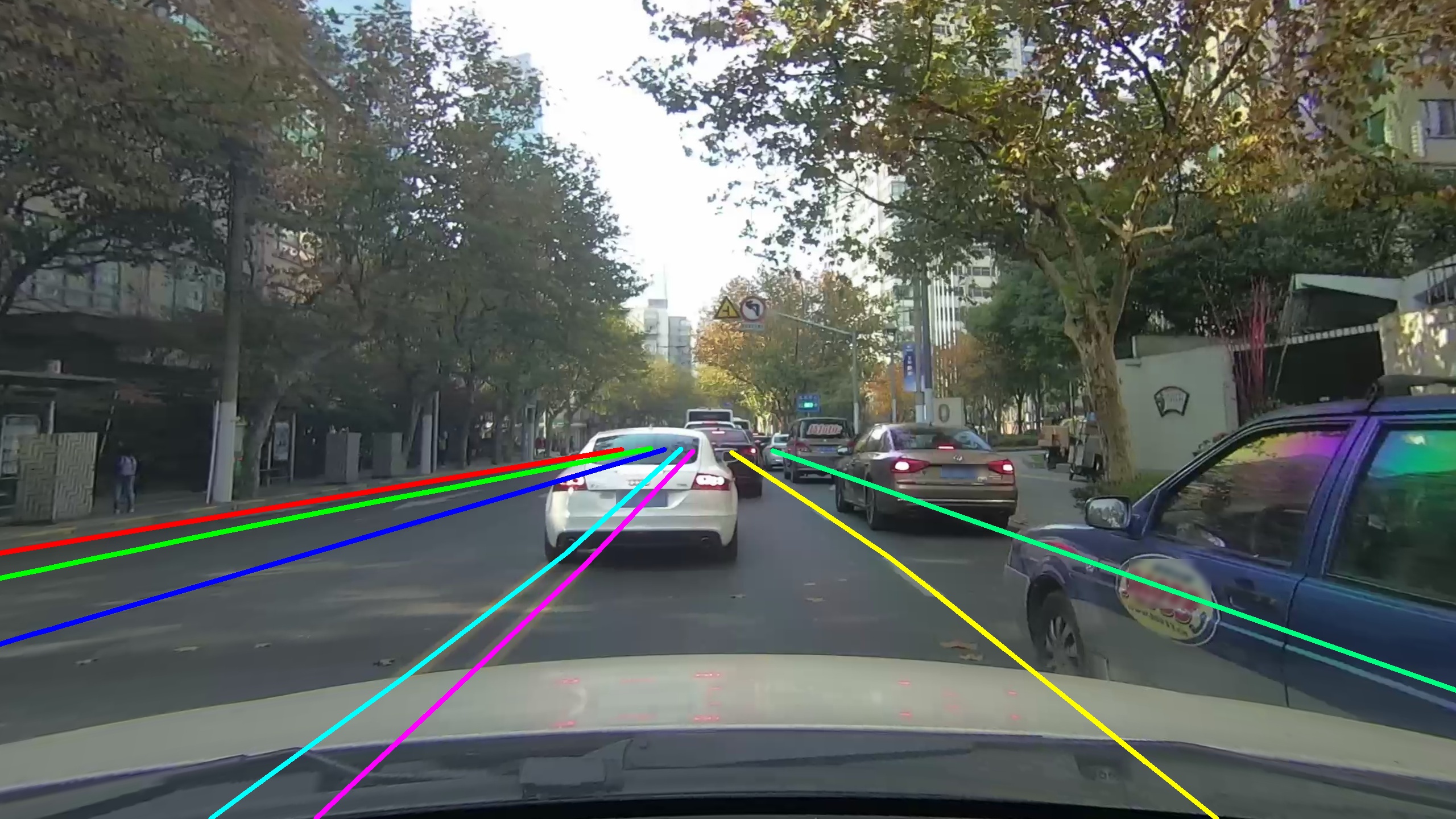}
        \end{subfigure}
        \begin{subfigure}{\subwidth}
                \includegraphics[width=\imgwidth, height=\imgheight]{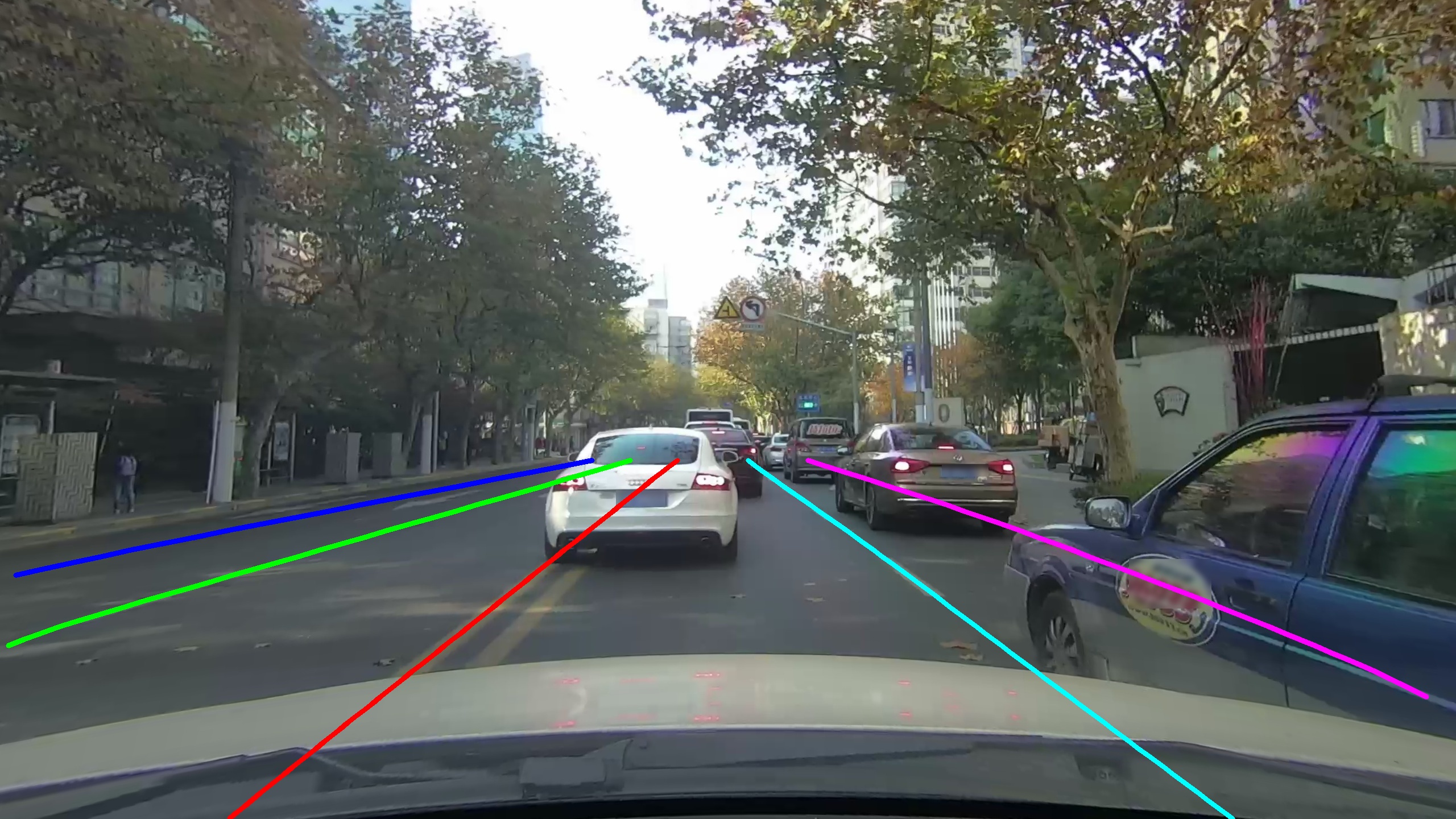}
        \end{subfigure}
        \begin{subfigure}{\subwidth}
                \includegraphics[width=\imgwidth, height=\imgheight]{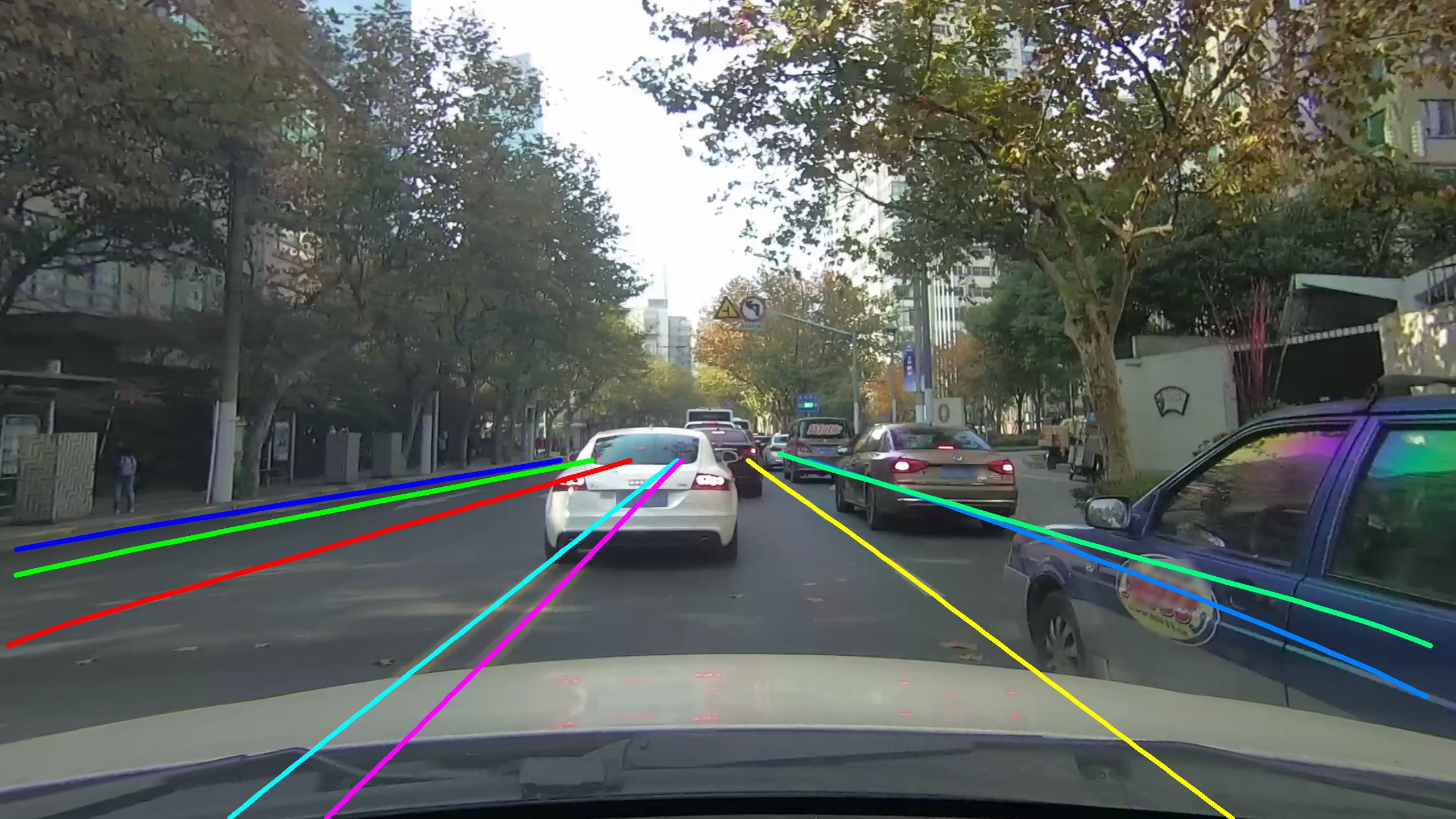}
        \end{subfigure}
        \begin{subfigure}{\subwidth}
                \includegraphics[width=\imgwidth, height=\imgheight]{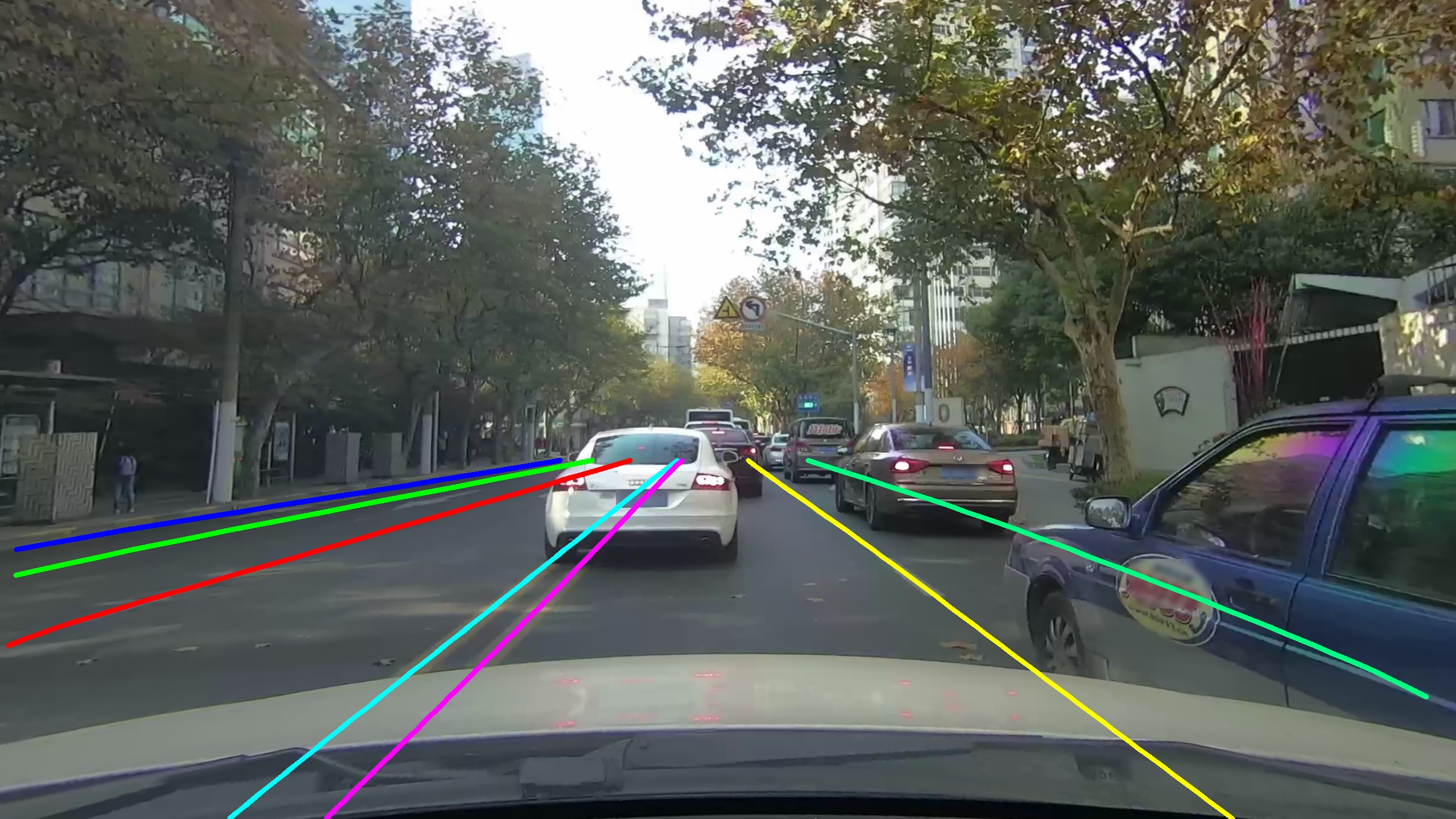}
        \end{subfigure}
        \vspace{0.5em}

        \begin{subfigure}{\subwidth}
                \includegraphics[width=\imgwidth, height=\imgheight]{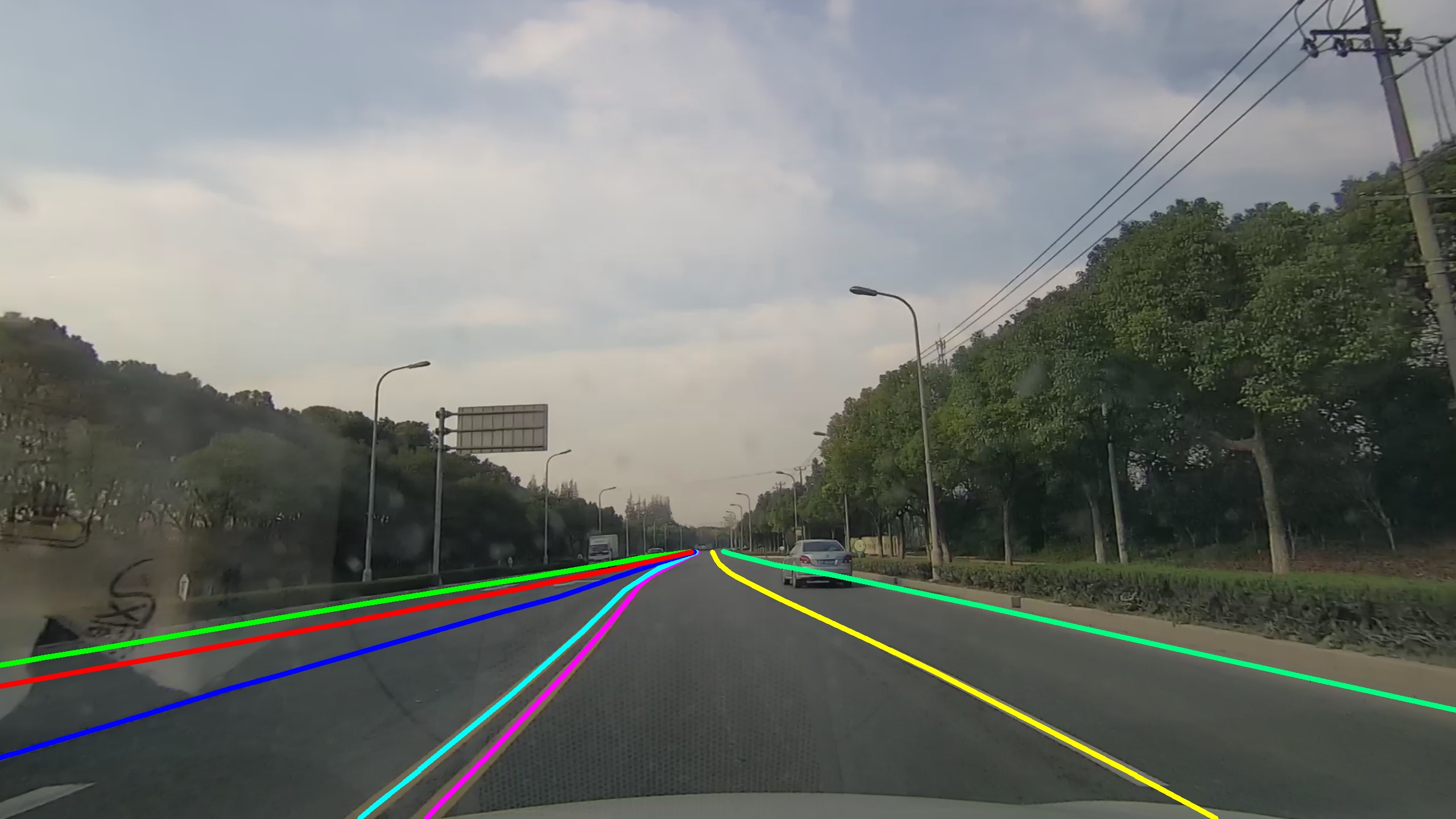}
                \caption{GT}
        \end{subfigure}
        \begin{subfigure}{\subwidth}
                \includegraphics[width=\imgwidth, height=\imgheight]{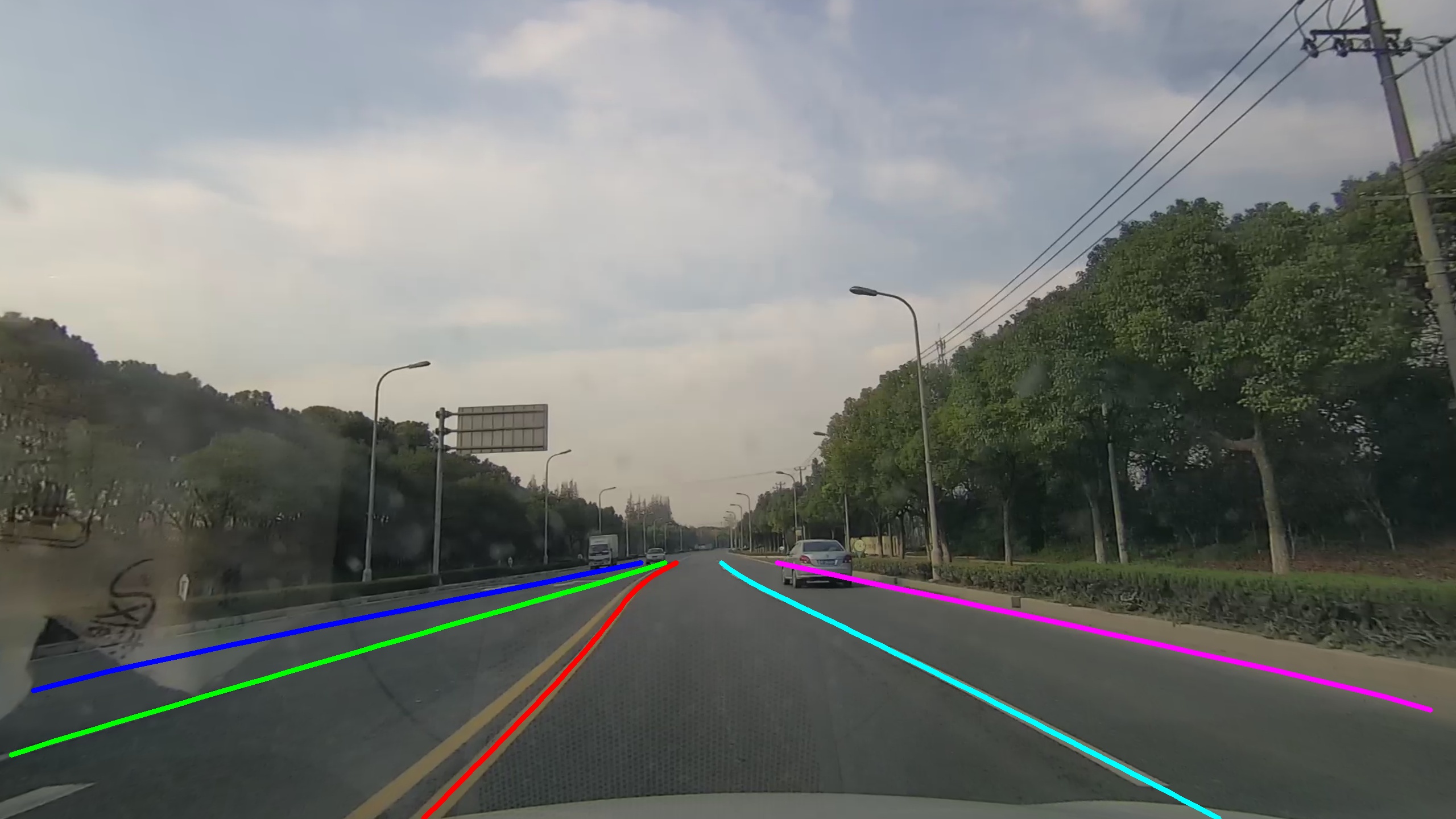}
                \caption{NMS@50}
        \end{subfigure}
        \begin{subfigure}{\subwidth}
                \includegraphics[width=\imgwidth, height=\imgheight]{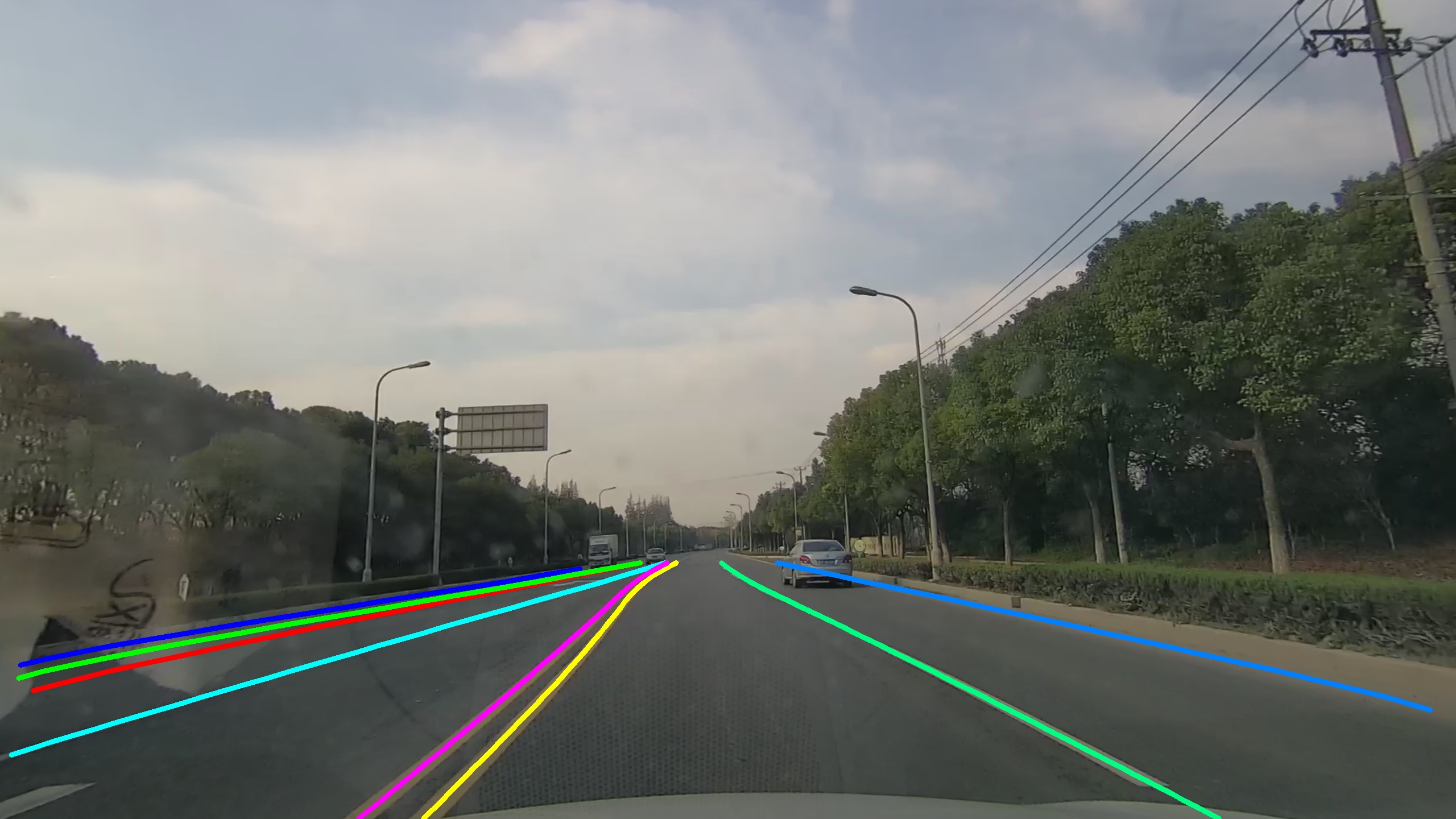}
                \caption{NMS@15}
        \end{subfigure}
        \begin{subfigure}{\subwidth}
                \includegraphics[width=\imgwidth, height=\imgheight]{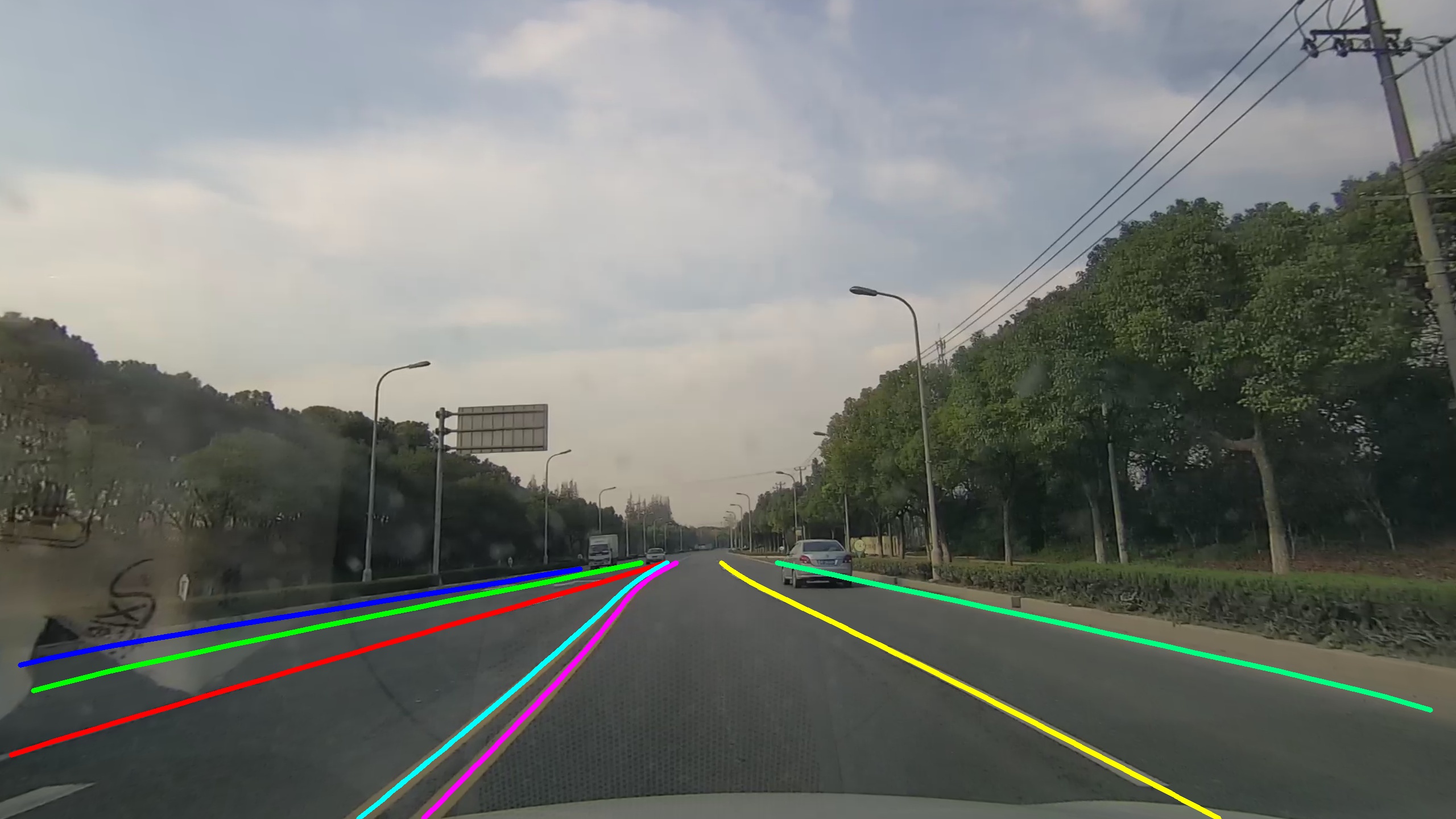}
                \caption{NMSFree}
        \end{subfigure}
        \vspace{0.5em}

        \caption{Visualization of the detection outcomes in sparse and dense scenarios on the CurveLanes dataset.}
        \label{vis_dense}
\end{figure*}
In the one-to-many label assignment, we simply use SimOTA \cite{yolox}, which aligns with previous works \cite{clrernet}. Omitting the detailed process of SimOTA, we only introduce the inputs to it, namely the cost matrix $\boldsymbol{M}^C\in \mathbb{R}^{G\times K}$ and the IoU matrix $\boldsymbol{M}^{IoU}\in \mathbb{R}^{G\times K}$. The elements in the two matrices are defined as $M^C_{qp}=\mathcal{C} _{p,q}^{o2m}$ and $M^{IoU}_{qp}= GIoU\left( p,q \right)$ (with $g=0$), respectively. The number of assigned predictions for each ground truth is variable but does not exceed an upper bound $k_{dynamic}$, which is set to $4$ in our experiment. Finally, there are $K_{pos}$ positive samples and $K-K_{pos}$ negative samples, where $K_{pos}$ ranges from $0$ to $Gk_{dynamic}$.

Given the ground truth label generated by the label assignment strategy for each prediction, we can conduct the loss function during phase. As illustrated in Fig. \ref{head_assign}, $\mathcal{L}_{cls}^{o2o}$ and $\mathcal{L}_{rank}$ are for the O2O classification subhead, $\mathcal{L}_{cls}^{o2m}$ is for the O2M classification subhead whereas $\mathcal{L}_{GIOU}$ (with $g=1$), $\mathcal{L}_{end}$ and $\mathcal{L}_{aux}$ for the O2M regression subhead.
\label{assign_appendix}

\section{The Supplement of Implement Detail and Visualization Results.}
Some important implement details for each dataset are shown in Table \ref{dataset_info}. It includes the dataset information we employed to conduct experiments and visualizations, the parameters for data processing as well as hyperparameters of Polar R-CNN. 

Fig. \ref{vis_sparse} illustrates the visualization outcomes in sparse scenarios spanning four datasets. The top row depicts the ground truth, while the middle row shows the proposed lane anchors and the bottom row exhibits the predictions generated by Polar R-CNN with NMS-free paradigm. In the top and bottom row, different colors aim to distinguish different lane instances, which do not correspond across the images. From images of the middle row, we can see that LPH of Polar R-CNN effectively proposes anchors that are clustered around the ground truth, providing a robust prior for GPH to achieve the final lane predictions. Moreover, the number of anchors has significantly decreased compared to previous works, making our method faster than other anchor-based methods in theory. 

Fig. \ref{vis_dense} shows the visualization outcomes in dense scenarios. The first column displays the ground truth, while the second and the third columns reveal the detection results with NMS paradigm of large (\textit{i.e.}, the default threshold NMS@50 with 50 pixels) and small (\textit{i.e.}, the optimal threshold NMS@15 with 15 pixels) NMS thresholds, respectively. The final column shows the detection results with NMS-free paradigm. We observe that NMS@50 mistakenly removes some predictions, leading to false negatives, while NMS@15 fails to eliminate some redundant predictions, leading to false positives. This underscores that the trade-off struggles between large and small NMS thresholds. The visualization distinctly demonstrates that distance becomes less effective in dense scenarios. Only the proposed O2O classification subhead, driven by data, can address this issue by capturing semantic distance beyond geometric distance. As shown in the last column of Fig. \ref{vis_dense}, the O2O classification subhead successfully eliminates redundant predictions while preserving dense predictions, despite their minimal geometric distances.
\label{vis_appendix}
\end{appendices}
\end{document}